\documentclass[english, PhD, oneside]{sapthesis}%remove "english" for a thesis written in Italian
\pdfoutput=1
\usepackage[utf8]{inputenx}
\usepackage{indentfirst}
\usepackage{microtype}
\usepackage{multirow}
\usepackage{xspace} 
\usepackage{tabu}
\usepackage{enumitem}
\usepackage{subfloat}
\usepackage{float}
\usepackage{lettrine}
\usepackage[nottoc, notlof, notlot]{tocbibind}
\usepackage{amsmath} % assumes amsmath package installed
\usepackage{amssymb}  % assumes amsmath package installed
\usepackage{algorithm,algcompatible}
\usepackage{bm}

\usepackage{graphicx}
\usepackage{subcaption}
\usepackage{adjustbox}
\usepackage{titlesec}

\usepackage[hyperfootnotes=true,			
			bookmarks=true]{hyperref}
\newcommand{\myTitle}{%Learning to Generalize across Domains and Modalities \\
Learning to see across Domains and Modalities}
\newcommand{\set}[1]{\ensuremath{\mathcal{#1}}}
\newcommand{\con}[1]{#1} %\ensuremath{\mathsf{#1}}}

\newcommand{\argmax}{\operatornamewithlimits{\arg\,\max}}

\def\etal{\textit{et al.}~}

\newcommand{\DAL}{DA-layer{} }
\newcommand{\DALs}{DA-layers{} }
\newcommand{\DIAL}{AutoDIAL\xspace}
\newcommand{\Alex}{AlexNet\xspace}
\newcommand{\Inception}{Inception-BN\xspace}

\newcommand{\VGGf}{VGGf}
\newcommand{\DIALAlex}{\DIAL{} -- \Alex}
\newcommand{\DIALInception}{\DIAL{} -- \Inception}

\newcommand{\DIALVGGf}{\DIAL{} -- \VGGf}

\algnewcommand\INPUT{\item[\textbf{Input:}]}%
\algnewcommand\OUTPUT{\item[\textbf{Output:}]}%
\newcommand{\DECO}[0] {(DE)\textsuperscript{2}CO}

\newcommand{\MDA}{\textit{MDA}\xspace}
\newcommand{\DG}{\textit{DG}\xspace}
\newcommand*{\eg}{e.g.\@\xspace}
\newcommand*{\ie}{i.e.\@\xspace}
\newcommand*{\adage}{ADAGE\@\xspace}

\hypersetup{
			colorlinks=true,
			linkcolor=red,
                        linktoc=page,
			anchorcolor=black,
			citecolor=red,
			urlcolor=blue,
			pdftitle={\myTitle},
			pdfauthor={Fabio Maria Carlucci},
			pdfkeywords={thesis, sapienza, roma, university, domain adaptation, computer vision, 
            			recognition, object classification, deep learning, transfer learning, depth, rgb}
 }

\title{\myTitle}
\author{Fabio Maria Carlucci}
\IDnumber{1550340}
\course[]{Ingegneria Informatica}
\courseorganizer{Department of Computer, Control and Management Engineering ”Antonio
Ruberti“, Sapienza - University of Rome}
\submitdate{2018/2019}
\copyyear{2018}
\advisor{Prof. Barbara Caputo}
\coadvisor{Dr. Tatiana Tommasi}
\authoremail{fabiom.carlucci@gmail.com}
\examdate{22\textbackslash02\textbackslash2019}
\cycle{XXXI}
\examiner{Prof. Riccardo Torlone} \examiner{Prof. Alessandro Farinelli} \examiner{Prof. Paolo Prinetto} 

\begin{document}

\frontmatter
\maketitle
\section*{Acknowledgements}
This section has always been the hardest one for me to write; not because I don't know who to thank, but because they deserve a real good acknowledgement and I feel the pressure of giving them what's due.

Firstly, I would like to give a huge thank you to my supervisor, Prof. Barbara Caputo. In a very short time she created, almost like a magician, a strong research group out of thin air.
Not only that, but her constant support and guidance has been crucial in these years; she's taught me a lot, on all related subjects, but one idea which has stuck is that being able to present well a good idea is as important as the idea itself - researchers need to be good at communicating. I \textit{know} that she has really shaped me into the researcher I am today.

Heartfelt thanks are reserved for Prof. Tinne Tuytelaars and Prof. Bastian Leibe, who have taken the time to accurately read my thesis and provide useful and constructive feedback.

The VANDAL lab has been an integral part of my journey and my thanks go to each of its members. Special thanks are reserved for Tatiana and Arjan, who I have personally pestered with more questions than should be allowed. As senior members, they have guided and pushed us all in the right direction. Tatiana, as my frequent co-author, earns extra thanks for adding her special touch to our papers... I know that without you my publication list would have been shorter! Thanks to Paolo for being my most prolific co-author and good friend. Valentina, Antonio, Nizar, Massimilino and Ilja I thank you all for being awesome friends and colleagues.

How can I not mention Giuseppe? Either when acting best man or simply when having a chat at a pub, you are the friend everybody would want, but not everybody has. Thanks to Sara and Elisa for \textit{adopting} me; getting to know you has been one of the best side-effects of meeting Angela.

Without my family I would not be here today: their constant support and love have been my safe haven in the years. Saying "thank you" to my parents feels almost reductive, but it is all I can do in this section. My brothers, grandparents, aunts, uncles and cousins have all been part of this journey and I feel lucky to have them: thank you for being there.

My family gave me all the tools to get here, but the one person that gave me the (very needed) push to achieve this PhD is Angela, my wife. You have always been by my side and managed to let me be the best possible version of myself; for that I love you and will always be grateful.

\begin{abstract}
Deep learning has recently raised hopes and expectations as a general solution for many applications (computer vision, natural language processing, speech recognition, etc.); indeed it has proven effective, but it also showed a strong dependence on large quantities of data. Generally speaking, deep learning models are especially susceptible to overfitting, due to their large number of internal parameters.
Luckily, it has also been shown that, even when data is scarce, a successful model can be trained by reusing prior knowledge. Thus, developing techniques for \textit{transfer learning} (as this process is known), in its broadest definition, is a crucial element towards the deployment of effective and accurate intelligent systems into the real world.
This thesis will focus on a family of transfer learning methods applied to the task of \textit{visual object recognition}, specifically \textit{image classification}. The visual recognition problem is central to computer vision research: many desired applications, from robotics to information retrieval, demand the ability to correctly identify categories, places, and objects. 
Transfer learning is a general term, and specific settings have been given specific names: when the learner has access to only unlabeled data from the \textit{target} domain (where the model should perform) and labeled data from a different domain (the \textit{source}), the problem is called \textit{unsupervised domain adaptation} (DA). The first part of this thesis will focus on three methods for this setting.
% One of the main goals of Computer Vision (CV) research is . Not just for a few clear-cut and well studied applications (\ie pedestrian detection, face recognition, ... ) but for a multitude of tasks, ranging from home robotics to assistance systems for the visually impaired. The dream should be the development of a family of algorithms which can be easily applied to different data without too much tinkering or too many restrictions. Deep Learning has recently raised hopes and expectations as a general solution to many CV problems; indeed, it is effective, but it requires lots of data to do so.
%This thesis deals with the question "how can I train an effective model when I do not have enough data?". It does so by multiple means: in the first part we will focuses on three methods for unsupervised domain adaption, the classical approach to this problem, while in the second part we will investigate two orthogonal solutions.
The three presented techniques for domain adaptation are fully distinct: the first one proposes the use of Domain Alignment layers to structurally align the distributions of the source and target domains in feature space. While the general idea of aligning feature distribution is not novel,
we distinguish our method by being one of the very few that do so
%this is the first work that does so 
without adding losses. The second method is based on GANs: we propose a bidirectional architecture that jointly learns how to map the source images into the target visual style and vice-versa, thus alleviating the domain shift at the pixel level. The third method features an adversarial learning process that transforms both the images and the features of both domains in order to map them to a common, agnostic, space.

While the first part of the thesis presented general purpose DA methods, the second part will focus on the real life issues of robotic perception, specifically RGB-D recognition. 
Robotic platforms are usually not limited to color perception; very often they also carry a Depth camera. 
%Domain adaptations techniques are effective, but have binding requirements: the key one being that source and target must share the same classes. 
%The second part of this thesis proposes two alternative solutions which ignore this requirement. In the context of RGB-D recognition, 
Unfortunately, the depth modality is rarely used for visual recognition due to the lack of pretrained models from which to transfer and little data to train one on from scratch.
We will first explore the use of synthetic data as proxy for real images by training a Convolutional Neural Network (CNN) on virtual depth maps, rendered from $3D$ CAD models, and then testing it on real robotic datasets. The second approach leverages the existence of RGB pretrained models, by learning how to map the depth data into the most discriminative RGB representation and then using existing models for recognition. This second technique is actually a pretty generic Transfer Learning method which can be applied to share knowledge across modalities.

\paragraph{Keywords:} deep learning, domain adaptation, transfer learning, rgb-d, depth, generative, recognition
\end{abstract}

\tableofcontents
% \listoffigures
% \listoftables

\mainmatter
\chapter{Introduction}
We are living the \textit{artificial intelligence} revolution. Almost every day an article in the news pops up, telling us how cars will soon be fully autonomous and robots will finally help us to do our house chores. New AI startups are constantly being launched and big companies are expanding their research teams and creating new labs.
For this new wave of optimism, we must thank the renaissance of the neural network~\cite{rumelhart1985learning,Backpropagation} (today known as \textit{deep learning}) paradigm, which was sparked in 2012 by the spectacular success of Krizhevsky's AlexNet \cite{krizhevsky2012imagenet} on the ImageNet \cite{deng2009imagenet} challenge. 

Since 2010, the annual ImageNet Large Scale Visual Recognition Challenge\cite{russakovsky2015imagenet} (ILSVRC) is a competition where research teams evaluate their algorithms on the given data set, and compete to achieve higher accuracy on several visual recognition tasks. AlexNet, a convolutional neural network (CNN), managed to outperform all other competing, non deep learning (we call them \textit{shallow} today) methods by a large margin on the object classification (over $1000$) categories challenge. This caused a huge paradigm shift in the Computer Vision field (soon extended to many other research fields) which prompted large improvements across many tasks and promoted the widespread optimism in the capabilities of AI we are currently seeing today.
Indeed, deep learning has changed the research landscape in visual object recognition over the last few years. Convolutional neural networks have become the new off the shelf state of the art in visual classification and thanks to these improvements we are actually starting to see real life applications, such as effective face and landmark recognition on our smartphones. 
The robot vision community has also attempted to take advantage of the deep learning trend, as the ability of robots to understand what they see reliably is critical for their deployment in the wild.

Unfortunately, as good as convolutional neural networks may be, they have a strong limitation: training a deep model from scratch requires lots of labeled data (as reference, the ILSVRC recognition dataset contains almost $1.5M$ images). Due to the large number of internal parameters of these deep networks, learning on small sets of data will often lead to overfitting and poor generalization. The good news is that, if we first train on a suitably large and diverse dataset, we obtain a model which, with little supervision, will perform well on a different, but related, task. This process of learning transfer, known as \textit{finetuning} \cite{zheng2016good,yosinski2014transferable,Transfer2,Transfer3}, is what makes it possible to exploit the deep learning potential on smaller sets of data.
Finetuning a model still requires \textit{some} labeled data belonging to our target domain; this may, in today's age of big data, initially appear as a non-problem but it is not so for a variety of reasons.

\begin{figure}[ht!]
    \centering
    \includegraphics[trim={5.5cm 19cm 5.5cm 4cm},clip]{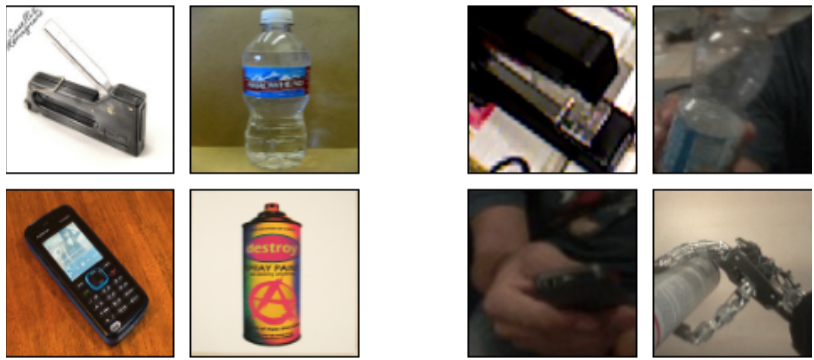}
    \caption{Sample images of certain classes (stapler, water bottle, cellphone, spray can) as seen in a wed dataset, Imagenet  \cite{russakovsky2015imagenet}, on the left, and in a real-life like dataset (JHUIT-50  \cite{jhuit}, HelloiCubWorld  \cite{fanello2013weakly}), on the right. Note that while they should be representing the same things, the images have very little in common}
\label{fig:intro_web_robo_vision}
\end{figure}

The web, with its easy access to large quantities of pictures (Google Image Search\footnote{https://images.google.com/}, Instagram\footnote{https://www.instagram.com/}, Pinterest\footnote{https://www.pinterest.com/}, etc..) seems the ideal and cheap source of labeled images, but present two significant challenges: data annotation and data bias.
Once you download images from the web, you must assign them labels, which is costly, in order to be able to train on them. Most web sources can provide labels, but they usually are noisy \cite{niu2018learning} and this will affect the final performances. For the scope of this thesis we will assume to be working with hand annotated data and consider our labels noise-free.

\begin{figure}
    \centering
    \includegraphics[width=0.8\textwidth]{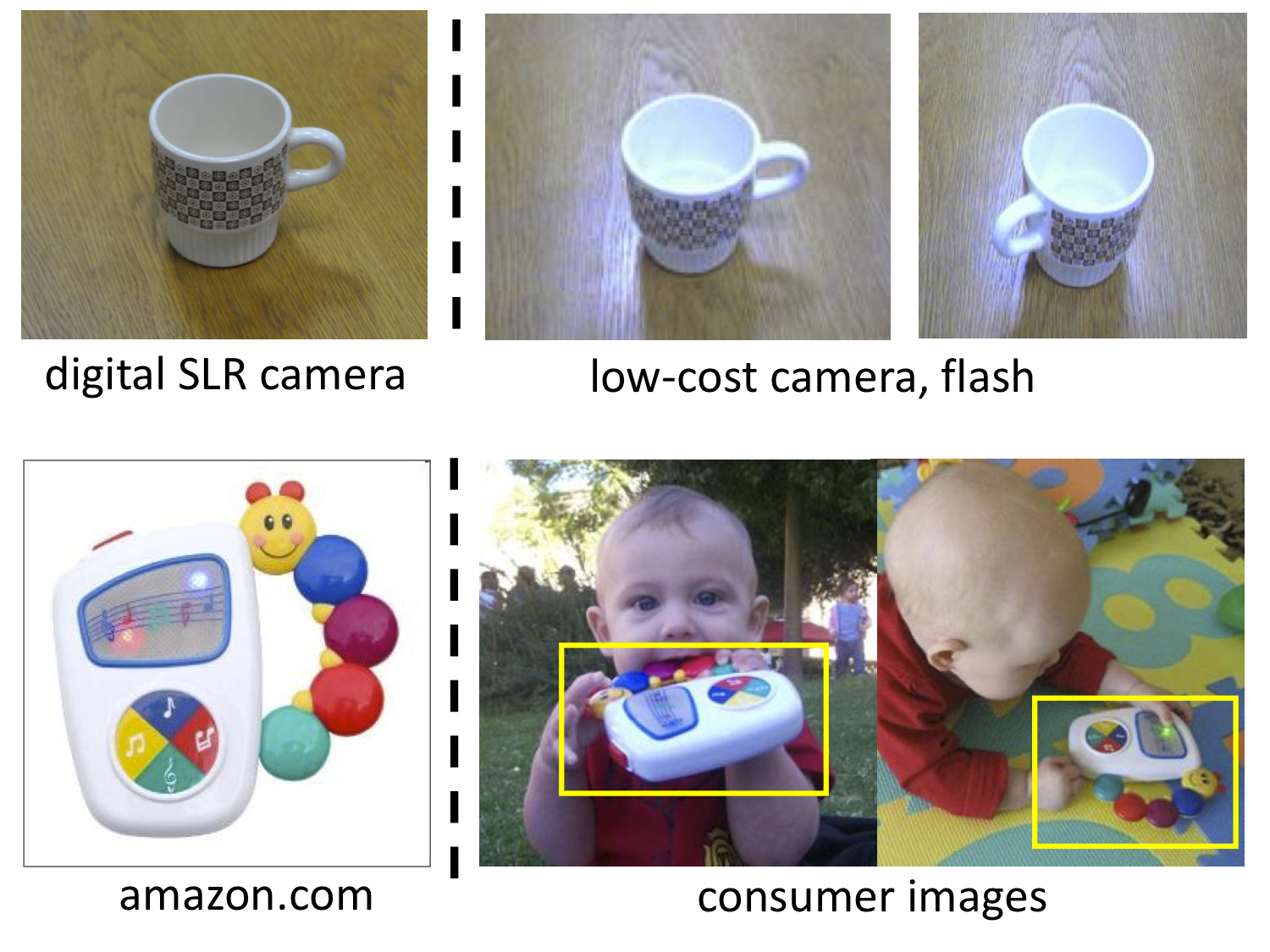}
    \caption{
    %Left: re-identification examples from the Viper dataset \cite{gray2007evaluating}. Right: a 
    A graphical intuition behind the Domain Adaptation problem (image from \href{https://people.eecs.berkeley.edu/~jhoffman/domainadapt}{https://people.eecs.berkeley.edu/~jhoffman/domainadapt}.)}
    \label{fig:domain_adapt_intuition}
\end{figure}

Even with perfect annotations, we still have to deal with the issue of data bias\cite{torralba2011unbiased}. A simple intuition of this problem can be had looking at Fig. \ref{fig:intro_web_robo_vision}. The images on the left belong to a dataset which was mined from the web, the images on the right were captured by a robot: the framing, the lightning conditions, the resolution, the background clutter are all different. If our model is trained with the images on the left it is easy to understand why it will perform poorly in the real world (another example of data bias can be seen in Fig. \ref{fig:domain_adapt_intuition}).
This is a pretty typical setup: we wanted to perform recognition on a set of classes, we used the web to download some training data (we will call it the \textit{source}) and found out that the model did not work well on real world data. During the deployment of our system we gathered some unlabeled data (our \textit{target}) for free.
We know that the labeled source and unlabeled target share the same classes and we would like for our model to perform well on both, ignoring their specific biases. This problem is formally known as that of \textit{unsupervised domain adaptation} (we will define it more rigorously in section \ref{def:DA}).

The first half of this thesis will deal with this issue, by investigating multiple Domain Adaptation (DA) approaches. Domain Adaptation is at its core the quest for principled algorithms enabling the generalization of visual recognition methods. Given at least a source domain for training, the goal is to achieve recognition results as good as those achievable on source test data on any other target domain, in principle belonging to a different probability distribution, without having prior access to labeled images. Solving this problem will represent a major step towards one of the key goals of computer vision, i.e. having machines able to answer the question ‘what do you see?’ in the wild; hence, its increased popularity in the community over the last years (see section \ref{sec:domain_adap_related} for a review of recent work)

%is our most readily available source of data, but presents a number of challenges. Firstly, it has been shown that there exists a large domain gap between Web Vision and Robot Vision \cite{d2017bridging} (in other words, the images we download from the web are pretty different from those that a robot will acquire in the wild, see Fig. \ref{fig:intro_web_robo_vision}). 

\begin{figure}
    \centering
    \includegraphics[height=4cm]{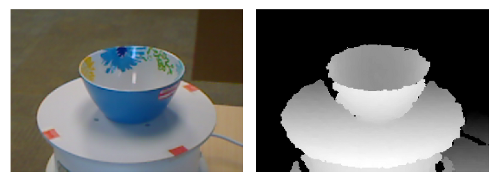}
    \caption{A sample RGB-D image from the Washington dataset  \cite{washington}. Note how the Depth modality provides geometric intuition cues}
    \label{fig:washington_example}
\end{figure}

Of course, DA is not the solution to all CV problems: RGB-D recognition, for example, encounters a different set of challenges.
RGB-D sensors (cameras capable of producing color images and depth maps at the same time) are extremely widespread on robotic platforms, but while depth cues could greatly help the recognition process by providing $3D$ intuition (see Fig. \ref{fig:washington_example}), it is often ignored.
%but while robots can process RGB images and perform effective recognition by using a pretrained model, the \textit{depth} modality is often ignored.
Why is this so? RGB-D datasets tend to be too small to train a CNN from scratch and the large difference in appearance between RGB and depth (see Fig. \ref{fig:washington_example}) severely limits the applicability of current finetuning and DA methods. Furthermore, we do have labels for our target depth dataset, so unsupervised (or semi-supervised) domain adaptation methods are not really suited. 

The second part of this thesis investigates specifically how different ways of transferring knowledge can benefit RGB-D recognition.

 %More in general, in spite of the progress brought by deep learning, the ability to generalize across different visual domains is still out of reach. The assumption that training (source) and test (target) data are independently and identically drawn from the same distribution does not hold in many real world applications. Indeed, it has been shown that, even with powerful deep learning models, the domain shift problem can be alleviated but not removed \cite{Transfer2}
% (domain shift can be visually seen in Fig. \ref{fig:intro_web_robo_vision} and \ref{fig:domain_adapt_intuition}).
 
% The second part of this thesis will deal with settings where classical Domain Adaptation methods are not applicabl. As a practical case study we will focus on non RGB modalities (specifically Depth sensors) and show that it is possible to train a strong classifier even when the available data is not particularly large. % and more broadly on how to best transfer what was previously learned.

%On a personal note, I would like to add that while Deep Learning methods require lots of data to perform well, and data is not always available, this is actually a good thing. It this were not true, there would be no need to come up with interesting solutions and the Computer Vision field would be terribly boring.

\section{Contributions}
Working in the context of visual image recognition, the main contributions of this work are two-fold: on one side, we contribute to the field of unsupervised domain adaptation with three novel and distinct techniques, and on the other we explore alternative approaches for those settings in which classical DA methods cannot be applied.
We then perform a qualitative and quantitative experimental evaluation of the proposed solutions, to quantify their effectiveness and robustness.
%show that employing the proposed solutions assures better performances than simply training on the available data. 

Specifically we present:

\paragraph{a Domain Alignment Layer for domain adaptation  \cite{carlucci2017auto,carlucci2017just}} which aligns multiple source domains between themselves and to the target at a feature level. Contrary to most previous methods, this solutions does not require a new loss term, as the domain aligning effect is implicit in the architecture. Concretely, what this layer does, is project the input features from each domain into a reference distribution, similarly to what a \textit{Batch Normalization} \cite{ioffe2015batch} layer would do, but it does so separately for each domain and learns how much statistic sharing should occur.

\paragraph{a generative approach to apply the style of the target domain to the source and vice versa  \cite{russo17sbadagan}} by exploiting the power of a bidirectional GAN \cite{Goodfellow:GAN:NIPS2014}. This technique allows us to bridge the domain gap at the image level by producing source images (of which we have labels) which look exactly as those from the target; clearly a classifier trained on these images will also perform well on the target. In this work we also explore the opposite transformation: making the target look as the source so that, when evaluated on a classifier trained on the source itself, no domain shift would present itself. Since this technique is applied at the image level, it lends itself well to integration with other feature based methods.

\paragraph{a novel approach which learns an agnostic representation for multiple domains  \cite{carlucci2018agnostic}} by transforming both the images themselves \textbf{and} and the  feature representation. The intuition here it that sometimes it is easier to reduce the domain shift by working on the images (\ie different background) and sometimes it is easier in feature space. There is no way of knowing this beforehand, so the best approach is to tweak both representations at the same time. Concretely this happens thanks to a complex network based on image generators and two \textit{Reverse Grad} \cite{Ganin:DANN:JMLR16} branches.

\paragraph{a case study on building a synthetic depth dataset for object recognition \cite{carlucci2016deep},} with the goal of training a model which can then be used on real robotic images. RGB-D sensors are commonly used in robotics, but the Depth modality is often times used only for navigation or segmentation. This is mainly due to the fact that there are no pretrained models for Depth recognition: this work tries to solve the issue by using 3D CAD models farmed from the Web.

\paragraph{a Transfer Learning method useful when source and target exist in different modalities \cite{carlucci2018text}}. Most transfer learning and domain adaptation methods require some assumptions to be true; here we present a general approach which allows knowledge sharing across modalities, with no prerequisites on the data\footnote{Clearly, the greater the similarity the greater the potential benefits}. The idea is that if we have a strong pretrained network on the some data (\ie ImageNet  \cite{russakovsky2015imagenet}) we can map our target data, from a different modality, to the source modality by learning a mapping which maximizes recognition performance.

\section{Outline}

\textbf{Chapter \ref{chap:related}} will provide a formal definition of the considered problems, present an overview of relevant works and introduce the datasets we will use for the experimental evaluation. Reflecting the structure of this thesis, this chapter is divided in two: the first part focuses on visual object recognition, more specifically on unsupervised domain adaptation methods for classification, 
while the second part will focus on the robotics problem of RGB-D recognition - a modality in which learning to transfer the knowledge is of primary importance.

The domain adaption literature (section \ref{sec:domain_adap_related}) will present shallow, deep and adversarial methods. We will focus on the different methods used in single-source, multi-source domain adaptation and domain generalization. 

Section \ref{sec:robo_related}, on RGB-D recognition, will review methods for classifying objects using Depth only and then on how Depth and RGB cues are often integrated together. As the transfer learning method we will present in section~\ref{sec:deco}  has drawn inspiration these techniques, we will also present some literature pertaining image colorization.
\newline

\textbf{Chapter \ref{chap:DA}} will delve into the details of the three unsupervised domain adaptation methods we propose in this thesis. 
%The chapter 
%starts by giving a brief problem statement, then 
%presents the datasets we will test on and then goes into the details of each solution.

Section \ref{sec:da_feature_align} presents \textbf{\DIAL} a method for single and multi-source domain adaption based on the use of special domain alignment layers, which reduce the domain shift at the feature level.

In section \ref{sec:da_image_align} we present \textbf{SBADA-GAN}, a GAN based method for single source domain adaptation. In this approach we tackle the domain bias directly at the image level, projecting the source images into the target style and vice versa.

\textbf{ADAGE}, presented in section \ref{sec:da_hybrid_align}, builds on both previous approaches: it is a method for single, multi-source domain adaption and domain generalization. We suggest that for some things it is easier to reduce the domain shift in feature space, while for other it is easier in image space: ADAGE does both at the same time by using an adversarial loss to learn a domain-agnostic representation for the images and features. 
\newline

%\textbf{Chapter \ref{chap:auxiliary}} starts with a description of the problem we are dealing with: why do we need special techniques to efficiently classify RGB-D data? It then details the datasets and protocols used for evaluation. After this, we finally present the two alternative methods we came up with to solve the issue. 
\textbf{Chapter \ref{chap:auxiliary}} describes two alternative methods which can be used lo learn an effective image classifier for modalities where data is scarce.

Section \ref{sec:depthnet} presents a case study on the use of $3D$ CAD models to generate synthetic depth maps as a proxy for real data. We trained a CNN (the \textbf{DepthNet})  on the virtual depth images and evaluated its performance on real-life robotic RGB-D datasets.

In section \ref{sec:deco} we present $\mathbf{(DE)^2CO}$, our Deep Depth Colorization method for converting depth images to RGB. We show this transformation allows us to effectively use models pretrained on ImageNet with great success on RGB-D datasets.
\newline

The thesis concludes with a summary discussion and remarks on possible future directions of research in
\textbf{chapter \ref{chap:conclusions}}.

\section{Publications}
The following list gives an overview of the author's publications in chronological order; note that a few of these published papers (marked with a * ) are not included in this thesis.
\begin{itemize}
\item Kuzborskij, I., Maria Carlucci, F., \& Caputo, B. (2016). When naive bayes nearest neighbors meet convolutional neural networks. In Proceedings of the IEEE Conference on Computer Vision and Pattern Recognition (pp. 2100-2109).*

\item Carlucci, F. M., Russo, P., \& Caputo, B. (2017, May). A deep representation for depth images from synthetic data. In Robotics and Automation (ICRA), 2017 IEEE International Conference on (pp. 1362-1369). IEEE.

\item Carlucci, F. M., Porzi, L., Caputo, B., Ricci, E., \& Bulò, S. R. (2017, September). Just dial: Domain alignment layers for unsupervised domain adaptation. In International Conference on Image Analysis and Processing (pp. 357-369). Springer, Cham. \textbf{(oral - Best Student Paper)}*

\item D’Innocente, A., Carlucci, F. M., Colosi, M., \& Caputo, B. (2017, July). Bridging between computer and robot vision through data augmentation: a case study on object recognition. In International Conference on Computer Vision Systems (pp. 384-393). Springer, Cham. \textbf{(Best Paper Award finalist)}*

\item Carlucci, F. M., Porzi, L., Caputo, B., Ricci, E., \& Bulò, S. R. (2017, October). AutoDIAL: Automatic Domain Alignment Layers. In Proceedings of the IEEE International Conference on Computer Vision (pp. 5077-5085).

\item Carlucci, F. M., Russo, P., \& Caputo, B. (2018). $DE^2CO$: Deep Depth Colorization. IEEE Robotics and Automation Letters, 3(3), 2386-2393. (also presented at the IEEE International Conference on Robotics and Automation).

\item Russo, P., Carlucci, F. M., Tommasi, T., \& Caputo, B. (2018). "From source to target and back: symmetric bi-directional adaptive GAN." In Proceedings of the IEEE Conference on Computer Vision and Pattern Recognition.
\end{itemize}

\textit{Note that an extended version of "AutoDIAL" is currently pending review for publication in the IEEE Transactions on Pattern Analysis and Machine Intelligence journal}

\subsection{Technical Reports pending review:}
\textit{to be submitted to the IEEE Conference on Computer Vision and Pattern Recognition (2019)}
\begin{itemize}
    \item Carlucci, F. M., Russo, P., Tommasi, T., \& Caputo, B. (2018). Agnostic Domain Generalization. arXiv preprint arXiv:1808.01102.
    \item Carlucci, F. M., D'Innocente, A., Caputo, B., \& Tommasi. T. "Jigsaw Domain Generalization." arXiv preprint (2018).
\end{itemize}

\chapter{Background and Related Work}
\label{chap:related}
\textit{
This chapter is divided in two sections: one related to the unsupervised domain adaptation problem (in the context of image classification), and the other pertaining RGB-D recognition, with a specific focus on the depth modality.
Each section starts with the problem formulation, continues with a review of related work and finally presents the datasets on which the proposed algorithms will be tested.
%In this chapter we review the problem of learning on a target with few labeled data (such that training a model from scratch is non optimal) by exploiting auxiliary sources.
%We examine how and why we must overcome the data bias between domains and investigate different means of transfer learning. We will focus both on methods used in Computer Vision literature and will also look at the specific needs and techniques used in Robot Vision 
}

\section{Computer Vision: Domain Adaptation}
\label{sec:domain_adap_related}
%Unsupervised domain adaptation focuses on the 
%challenging 
%scenario where
%labeled data are only available in the source domain.
\textit{In this section, first we present a definition of the unsupervised domain adaptation problem and then review previous works. We consider traditional approaches based on shallow models, methods based on deep architectures and more recent techniques based on the generative adversarial paradigm. For each, we describe works on single-source, multi-source domain adaptation and domain generalization. In conclusion we present the datasets commonly used by the community, which we will use to evaluate our solutions}
\newline

\subsection{Unsupervised Domain Adaptation}
\label{def:DA}
In the introduction of this thesis we gave an intuition of what domain adaptation is and why is it relevant to the deployment of well performing recognition systems in the wild. Here we provide a more formal description:

Domain adaptation \cite{tommasi2013learning} aims at solving the learning problem on a target domain $T$ exploiting information from a source domain $S$, when both the domains and the corresponding tasks are not the same. 
More specifically, while the tasks have identical label sets $Y^s = Y^t$ they possess (slightly) different conditional distributions $P^s(Y |X) \sim P^t(Y|X)$. 
The domains are different in terms of marginal data distribution $P^s(X) \neq P^t(X)$ , and/or in feature spaces $X^s \neq X^t$. 
Our goal is to estimate a predictor from $S$ and $T$ that can be used to classify
sample points from the target domain.
Domain adaptation has been studied in two main settings: one is the semi-supervised case, where the target presents few labeled data, while the other is the unsupervised case that
considers only unlabeled examples for the target. In both cases, the source set is generally rich
in labeled samples. In this thesis we will focus on the unsupervised case.
We can then define the source as $S=\{(x_1,y_1), ..., (x_n,y_n)\} \subset \set X\times\set Y$ and the target as $T=\{\hat{x}_1, ..., \hat{x}_m\} \subset \set X$

\paragraph{Multi-source}is a natural extension of the the single source domain adaption setting where multiple source domains are considered instead of a single one. In this case we can redefine the source as:
$S=\{(x_1,y_1,d_1), ..., (x_n,y_n, d_n)\} \subset \set X\times\set Y\times\set D$. Generally speaking, the various source domains are also different in terms of marginal data distribution.

\paragraph{Domain Generalization}(DG) is an even more challenging variation. In this setting we usually have multiple sources but target data is not available during the training process. In order to best perform on the, unseen, target, DG methods attempt to train the most general and robust classifier possible on the source domains.
%The goal here is to learn the most general classifier on the sources, in order to make it perform well con the, unseen, target.

\subsection{Literature survey}
\subsubsection{Single source domain adaptation}
Traditional approaches addressed the problem of reducing the discrepancy between the source
and the target distributions by considering two main strategies. The first is based on instance re-weighting \cite{huang2006correcting,chu2013selective,yamada2012no,gong2013connecting,zeng2014deep}:  
source samples are assigned different importance according to their similarity to the target data. The re-weighted samples are then used to learn a classification model for the target domain. Following this scheme, Huang \etal\cite{huang2006correcting} introduced Kernel Mean Matching, a nonparametric
method to set source sample weights without explicitly estimating the data distributions.
Gong \etal\cite{gong2013connecting} proposed to automatically discover a set of landmark datapoints, \ie
to identify the source samples which are most similar to target instances, and used them to create
domain-invariant features. Chu \etal \cite{chu2013selective} unified the tasks of source sample selection
and classifier learning within a single optimization problem. 
While these works considered hand-crafted features, recently similar ideas have been
applied to deep models. For instance, Zeng \etal \cite{zeng2014deep} described 
an unsupervised domain adaptation approach for pedestrian detection using deep autoencoders to weight the importance of source training samples.

A second strategy for unsupervised domain adaptation is based on feature alignment, \ie source and target data are projected in a common subspace in order to reduce the distance among the associated distributions. This approach attracted considerable interest in the past years and several different methods have been 
proposed, both considering shallow models \cite{gong2012geodesic,long2013transfer,fernando2013unsupervised} and deep architectures \cite{long2015learning,tzeng2015simultaneous,ganin2014unsupervised,ghifary2016deep,bousmalis2016domain}.

\begin{figure}
    \centering
    \includegraphics[width=0.9\textwidth]{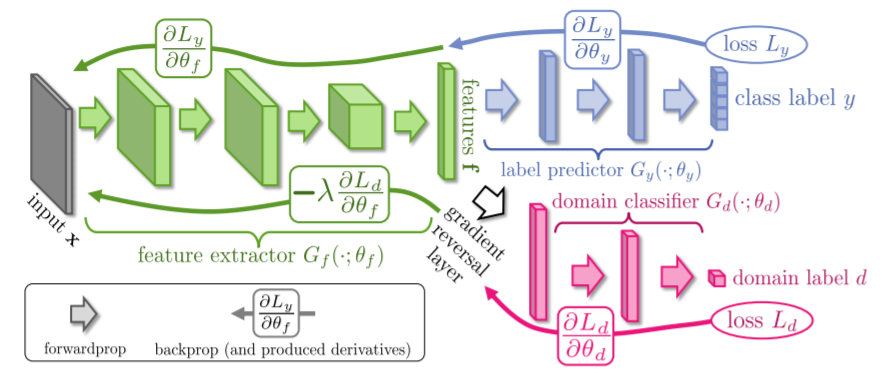}
    \caption{The proposed architecture for the use of the reverse gradient, from "Unsupervised Domain Adaptation by Backpropagation"~\cite{ganin2014unsupervised}. Our ADAGE framework, in section~\ref{sec:da_hybrid_align}, is influenced by this work and depends on the use of gradient reversal layers.}
    \label{fig:ganin_arc}
\end{figure}

Focusing on recent deep domain adaptation methods, different schemes have been considered to align feature representations. Earlier works proposed to reduce the domain shift, while learning deep representations, by modeling source and target data distributions in terms of their first 
\cite{long2015learning,long2016unsupervised,long2016deep,venkateswara2017deep} and second order \cite{sun2016deep} statistics. Other works proposed to learn domain-agnostic deep features within a domain-adversarial setting \cite{tzeng2015simultaneous,ganin2014unsupervised}. Haeusser \textit{et al.} \cite{haeusser17} described a methodology to learn domain invariant features by reducing the distance among source and target samples of each class, instead of considering the whole sets. Sankaranarayanan \textit{et al.} \cite{sankaranarayanan2017generate} proposed to learn an
embedding that is robust to the shift between source and target distributions by using a combination of a
classification loss and an image generation procedure modeled using a variant of Generative Adversarial Networks (GANs)\cite{Goodfellow:GAN:NIPS2014}. Other methods attempted to reduce the distribution mismatch between source and target data by embedding into a neural network specific distribution normalization layers ~\cite{li2016revisiting,mancini2018boosting,mancini2018kitting}.

Our approach in section \ref{sec:da_feature_align} belongs to the category of methods employing domain normalization layers for domain adaptation. Similarly to previous work we exploit Batch Normalization in the context of domain adaptation and propose to reduce the discrepancy between source and target distributions by introducing our DA-layers. However, we significantly depart from previous works \cite{li2016revisiting,mancini2018boosting,mancini2018kitting}, as our DA layers allow to automatically tune the required degree of adaptation at each level of the deep network. Furthermore, we also introduce a prior over the network parameters in order to fully benefit from the target samples during training.
%On the other hand, \cite{sankaranarayanan2017generate} is feeding the feature embeddings of the images to a GAN\cite{Goodfellow:GAN:NIPS2014} and then backpropagating the gan discriminator loss back to the classification network, thus aligning the features.

Entropy regularization for semi-supervised learning was originally introduced in \cite{grandvalet2004semisupervised}. This concept has been then exploited in deep neural networks with the name of learning with pseudo-labels \cite{zhu2005semi}. The main idea is to use high confidence predictions as labels for unlabeled samples in order to improve the performance of the classifier \cite{zhu2005semi}.
Recently, some domain adaptation methods \cite{russo17sbadagan,saito2017asymmetric,long2016unsupervised,long2016deep,venkateswara2017deep} have exploited this technique for learning deep representations. However, in section \ref{sec:da_feature_align} we demonstrate that this approach is especially effective when applied in combination with our proposed DA-layers.

%GANS
More recently, several approaches have attempted to reduce the domain shift by directly transforming images using Generative Adversarial Networks~\cite{Goodfellow:GAN:NIPS2014} (GANs).
%Generative Adversarial Networks are composed of two modules, a generator and a discriminator. The generator synthesizes samples whose distribution closely matches that of real data, while the discriminator distinguishes real from generated samples. 
Vanilla GANs are agnostic to the training samples labels, while conditional 
GAN variants \cite{Mirza:cGAN:arXiv2014} exploit the class annotation as additional information to both the generator and the discriminator.
Some works used multiple GANs: in CoGAN \cite{Liu:coGAN:NIPS16} two generators 
and two discriminators are coupled by weight-sharing to learn the joint distribution of images in two different domains without using pair-wise data.
Cycle-GAN \cite{CycleGAN2017}, Disco-GAN \cite{DBLP:conf/icml/KimCKLK17} and UNIT \cite{liu2017unsupervised} encourage the mapping between two domains to be well covered by imposing transitivity: the mapping in one direction followed by the mapping in the opposite direction should arrive where it started. 
Initially GAN methods were not exploited for domain adaptation, but recently several approaches have attempted to reduce the domain shift by directly transforming images using this approach \cite{Bousmalis:Google:CVPR17, Shrivastava:arXiv:16,Taigman2016UnsupervisedCI, hoffman2017cycada,russo17sbadagan}. 
\cite{Bousmalis:Google:CVPR17} proposed a GAN-based approach that adapts source images 
to appear as if drawn from the target domain; the classifier trained 
on such data outperformed several domain adaptation methods. 
\cite{Taigman2016UnsupervisedCI} introduced a method to generate source images that resemble the target ones, with the extra consistency constraint that the same transformation should keep the target samples identical.
All these methods focus on the source-to-target image generation, not considering the inverse procedure, from target to source, which we, in section \ref{sec:da_image_align}, show instead to be beneficial.

It must be noted that, while deep generative models are very effective in specific applications (\eg adaptation from synthetic to real images\cite{Shrivastava:arXiv:16,Bousmalis:Google:CVPR17}, digit recognition \cite{Bousmalis:Google:CVPR17, Shrivastava:arXiv:16, Taigman2016UnsupervisedCI, hoffman2017cycada, russo17sbadagan}), their performance in other settings is limited. % but have yet to scale to more general DA datasets.

\begin{figure}[b]
    \centering
    \includegraphics[width=1.0\textwidth]{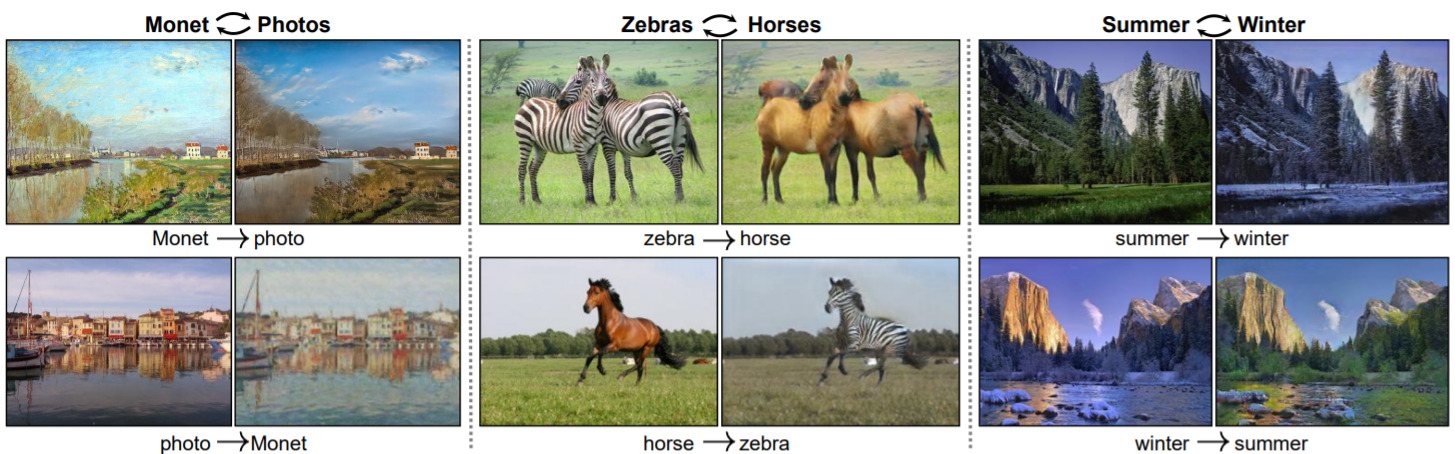}
    \caption{Image samples from CycleGAN\cite{CycleGAN2017}. Given any two unordered image collections X and Y , their algorithm learns to automatically “translate” an image from one into the other and vice versa. Our SBADA-GAN (section \ref{sec:da_image_align}) approach can be seen as natural extension of this method to the task of domain adaptation.}
    \label{fig:cyclegan_examples}
\end{figure}
%Experiments presented in Section~\ref{experiments} show the significant added value of our idea.

\subsubsection{Multi-source domain adaptation}
A reasonable assumption when building visual recognition systems is that we may have access to more than one source domain. In this context, the study of multi-source domain adaptation algorithms is especially important. 

This multi-source setting (Multi Domain Adaptation, \MDA), which originated from A-SVM~\cite{yang2007cross}, was initially studied from a theoretical point of view, focusing on theorems indicating how to optimally sub-select the data to be used in learning the source models \cite{Crammer_JMLR08}, or  proposing principled rules for combining the source-specific classifiers and obtain the ideal target class prediction \cite{Mansour_NIPS09}. 
Several other works followed this direction in the shallow learning framework. 
presenting algorithms based on the combination of source hypotheses \cite{Rita_NIPS2011,Duan_2009}.
A different set of multi-source adaptation approaches are based on learning new data representations, defining a mapping to a latent feature space shared between domains. Some of them are created for single sources but can be easily extended to multiple ones, as the feature replication method of \cite{daume2007} and the unsupervised approach based on Grassman manifolds presented in \cite{Gopalan2011}, which supports multiple sources by first computing the Karcher mean of the sources and then exploiting the standard single source version of the algorithm. 
Other works presented instead dedicated solutions for multiple sources \cite{hoffman_eccv12,JhuoLLC12,DuanTNNLS2012,Liu_ICDM16}.

Recently some deep architectures have been proposed to tackle 
this challenge. In particular, Xu \etal \cite{cocktail_CVPR18} proposed an approach derived from \cite{ganin2016domain} which exploits an adversarial domain discriminator branch for learning domain agnostic features. %, according to the rule in \cite{mansour2009domain}. 
A similar multi-way adversarial strategy was also  introduced in \cite{MDAN_ICLRW18}. Mancini \etal \cite{mancini2018boosting} proposed an approach for multi-source domain adaptation where latent domains are automatically discovered in the training set. Our work in section \ref{sec:da_feature_align} differs significantly from these previous works, as we propose a multi-source domain adaptation method based on DA-layers which automatically learns at each network layer the optimal degree of adaptation.

\textbf{Domain Generalization}
Domain Generalization (DG) is a newer line of research first studied by Blanchard et al.~\cite{blanchard2011generalizing}; in this setting, no transductive access to the target data is allowed, thus the main objective is to look across multiple sources for shared factors in the hypothesis that they will hold also for any new target domain. 
A number of shallow methods have been presented in the years. ~\cite{shallowDG} propose a projection-based algorithm, Domain-Invariant Component Analysis
(DICA) which extends Kernel PCA by incorporating the distributional
variance in order to reduce the dissimilarity across domains and the central
subspace. Khosla et al. \cite{khosla2012undoing} proposed a multi-task max-margin
classifier that explicitly encodes dataset-specific biases in feature space. These biases are used to push the dataset-specific weights to be similar to the global weights. Ghifary et al.~\cite{ghifary2017scatter} proposed Scatter Component Analyis (SCA), a representation learning algorithm for both domain adaptation and domain generalization. SCA, based on a the scatter measure which operates on reproducing kernel Hilbert space, finds a representation that trades between
maximizing the separability of classes, minimizing the mismatch between domains, and maximizing the separability of data.

Deep learning methods usually search for shared factors either at \emph{model-level} to regularize the learning process on the sources, or at \emph{feature-level} to learn some domain-shared representation. 
When focusing on deep DG methods, model-level strategies are presented in \cite{MassiRAL} and \cite{hospedalesPACS}. 
The first work proposes a weighting procedure on the source models, while the second aims at separating the source knowledge into domain-specific and domain-agnostic sub-models.
A meta-learning approach was recently presented in \cite{MLDG_AAA18}: it starts by creating virtual testing domains within each source mini-batch and then it trains a network to minimize the classification loss, while also ensuring that the taken direction leads to an improvement on the virtual testing loss.
Regarding feature-based methods, \cite{doretto2017} proposes to exploit a Siamese architecture to learn an embedding space where samples from different source domains but same labels are projected nearby, while samples from different domains and different labels are mapped far apart. Both the works \cite{DGautoencoders,Li_2018_CVPR} exploit deep learning autoencoders for domain generalization still focusing on representation learning.
%but the proposed method exploits image reconstruction: multi-task autoencoders are trained so that, given an
%image from one domain it is possible to reconstruct its analogous for all domains. 
A new way to tackle DG was recently presented in \cite{DG_ICLR18}. Instead of aiming at the reduction of domain-specific signals, this work introduces a form of data augmentation based on domain-guided perturbations of the input instances. 
%More in details, a domain classifier is
%initially trained to distinguish among the multiple sources, then for every source sample its loss is used to perturb the sample in the direction that changes the same loss the most. 
A label classifier is then trained on both the original instance and the data produced by the described augmentation process.

As a final remark, we note that, despite for both DA and DG there
exist multiple methods based on features and representation learning, image-adaptive solutions are exclusive for the single source domain adaptation setting. The only methods in DG that somehow involve images are \cite{DGautoencoders,DG_ICLR18} but the focus is either on source-to-source reconstruction or on data augmentation. With \adage, a method we introduce in section \ref{sec:da_hybrid_align}, we introduce a joint image and feature adaptation method: it learns how to project images into a network-understandable agnostic visual space that paves the way for completing the domain gap closure in the feature space. 

\subsection{Datasets}
\label{sec:da_datasets}
\textit{The following section presents the datasets and the protocols which will be used to evaluate the DA methods proposed in chapter \ref{chap:DA}. We first present some simple datasets, commonly used by generative adversarial methods, and then describe the more complex scenarios on which DA methods are routinely tested}

\subsubsection{Digits like datasets}
\label{ss:da_digits_datasets}
\textit{The following datasets are fairly simple, with moderate intra class variability. They are commonly used to evalute GAN based methods, which, to date, still struggle with more complex data}
\newline
\begin{figure}
    \centering
    \includegraphics[width=1.0\textwidth]{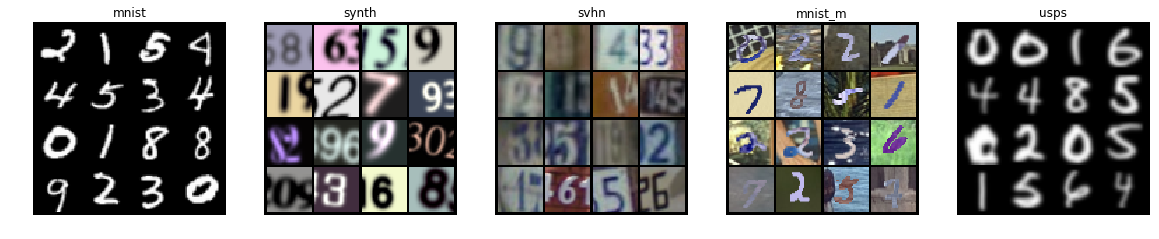}
    \caption{Samples from the digits datasets. From left to right: MNIST, SYNTH, SVHN, MNIST-M and USPS}
    \label{fig:data_digits_samples}
\end{figure}

\textbf{MNIST}~\cite{lecun1998gradient} is the dataset on which the first convolutional neural network was trained. It has a training set of $60k$ examples, and a test set of $10k$ examples. It is a subset of a larger set available from NIST. The digits have been size-normalized and centered in a fixed-size image. 
The images contain a single digit numbers on a black background. 

\textbf{MNIST-M}~\cite{Ganin:DANN:JMLR16} is a variant where the background is substituted by a 
randomly extracted patch obtained from color photos of BSDS500~\cite{arbelaez2011contour}.

\textbf{USPS}~\cite{friedman2001elements} is a digit dataset automatically scanned from envelopes 
by the U.S. Postal Service containing a total of $9,298$ $16\times16$ pixel grayscale samples; 
the images are centered, normalized and show a broad range of font styles. 

\textbf{SVHN}~\cite{netzer2011reading} is the challenging real-world Street View House 
Number dataset. It contains over $600$k $32\times32$ pixel 
color samples, while we focused on the smaller version of almost $100$k %$99,289$ 
cropped digits. Besides presenting a great variety of shapes and textures, images from this dataset 
often contain extraneous numbers in addition to the labeled, centered one.

\textbf{SYNTH Digits} this collection~\cite{Ganin:DANN:JMLR16} 
consists of $500$k images generated from Windows$^{TM}$ fonts by varying the
text (that includes different one-, two-, and three-digit numbers), positioning, orientation,
background and stroke colors, and the amount of blur

\textbf{Synth Signs} the Synthetic Signs collection~\cite{Moiseev:2013} contains $100k$  samples of common street signs obtained from Wikipedia and artificially transformed to simulate various imaging conditions. 

\textbf{GTSRB}The German Traffic Signs Recognition Benchmark (GTSRB, \cite{stallkamp2011german}) consists of $51,839$ cropped images of German traffic signs.

\begin{figure}
    \centering
    \includegraphics[width=0.5\textwidth]{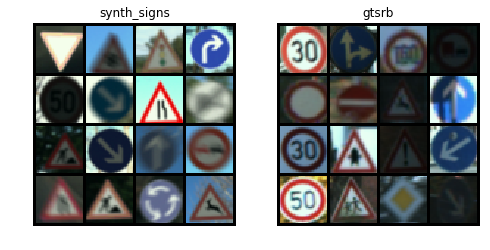}
    \caption{Samples from the signs datasets. From left to right: Synth Signs and GTSRB}
    \label{fig:data_signs_samples}
\end{figure}

\subsubsection{Real life datasets}
\label{ss:da_real_datasets}
\textit{These datasets are more complex and unconstrained; they better mimic real world applications. Most of them are commonly used for evaluation of recent deep learning based methods.}
\newline

\begin{figure}
    \centering
    \includegraphics[width=1.0\textwidth]{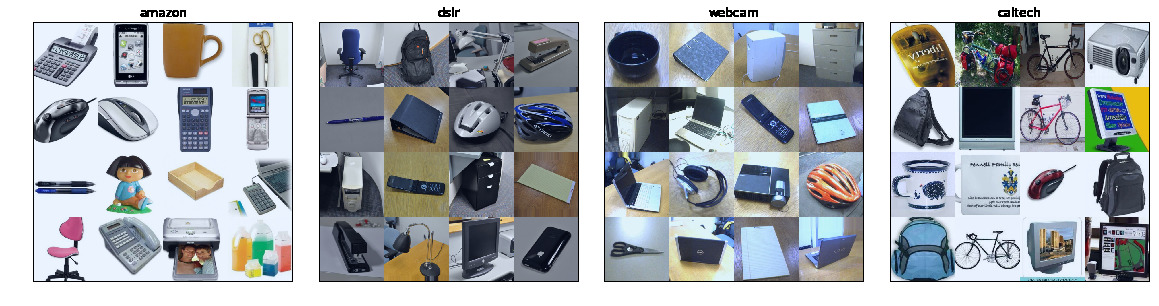}
    \caption{Samples from the Office and Caltech settings. From left to right: Amazon, DSLR, Webcam and Caltech}
    \label{fig:data_officecalt_samples}
\end{figure}

The \textbf{Office 31}\cite{saenko2010adapting} dataset has been the standard benchmark for testing domain-adaptation methods for many years.
It contains $4652$ images organized in $31$ classes from three different domains: Amazon (A), DSRL (D) and Webcam (W).
Amazon images are collected from \texttt{amazon.com}, Webcam and DSLR images were manually gathered in an office environment. As can be seen in Fig.~\ref{fig:data_officecalt_samples}, the domain gap is pretty large between Amazon and the remaining two, and small between DSLR and Webcam.
In our experiments we consider all possible source\slash{}target combinations of these domains and adopt the \emph{full protocol} setting \cite{gong2013connecting}, \ie we train on the entire labeled source and unlabeled target data and test on annotated target samples.
For the multi-source setting, we instead follow the protocol described in~\cite{cocktail_CVPR18}.

\begin{figure}
    \centering
    \includegraphics[width=1.0\textwidth]{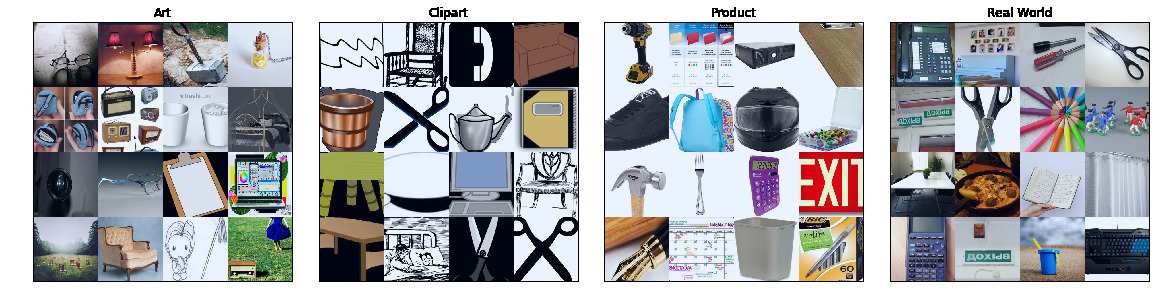}
    \caption{Samples from the Office-Home setting. From left to right: Art, Clipart, Product and Real World}
    \label{fig:data_officehome_samples}
\end{figure}

\textbf{Office-Home} \cite{venkateswara2017deep} is a recently released dataset, built to overcome the size limitation of previous domain adaptation settings.
It consists of $4$ domains, each domain containing $65$ matching categories of everyday objects from the home and  office environment. Looking at Fig. \ref{fig:data_officehome_samples} we can see how there exists a large domain gap between domains containing drawings and domains containing photos.
In total this dataset is composed by $15.500$ images gathered from the web.
The domains are: \textit{Art}, \textit{Clipart}, \textit{Product} and \textit{Real-World}.

The \textbf{Office-Caltech} \cite{gong2012geodesic} dataset is obtained by selecting the subset of $10$ common categories in the Office31 and the Caltech256\cite{griffin2007caltech} datasets.
It contains $2533$ images of which about half belong to Caltech256.
Each of Amazon (A), DSLR (D), Webcam (W) and Caltech256 (C) are regarded as separate domains.
In our experiments we only consider the source\slash{}target combinations containing C as either the source or target domain.

To further perform an analysis on a large-scale dataset, we also consider the \textbf{Cross-Dataset Testbed} introduced in \cite{tommasi2014testbed} and specifically the \textbf{Caltech-ImageNet} setting.
This dataset was obtained by collecting the images corresponding to the $40$ classes shared between the Caltech256 (C) and the Imagenet (I) \cite{deng2009imagenet} datasets.
To facilitate comparison with previous works \cite{tzeng2015simultaneous,tommasi2016learning,sun2016return} we perform experiments in two different settings.
The first setting, adopted in \cite{tommasi2016learning,tzeng2015simultaneous}, considers 5 splits obtained by selecting 5534 images from ImageNet and 4366 images from Caltech256 across all 40 categories.
The second setting, adopted in \cite{sun2016return}, uses 3847 images for Caltech256 and 4000 images for ImageNet.

\begin{figure}
    \centering
    \includegraphics[width=1.0\textwidth]{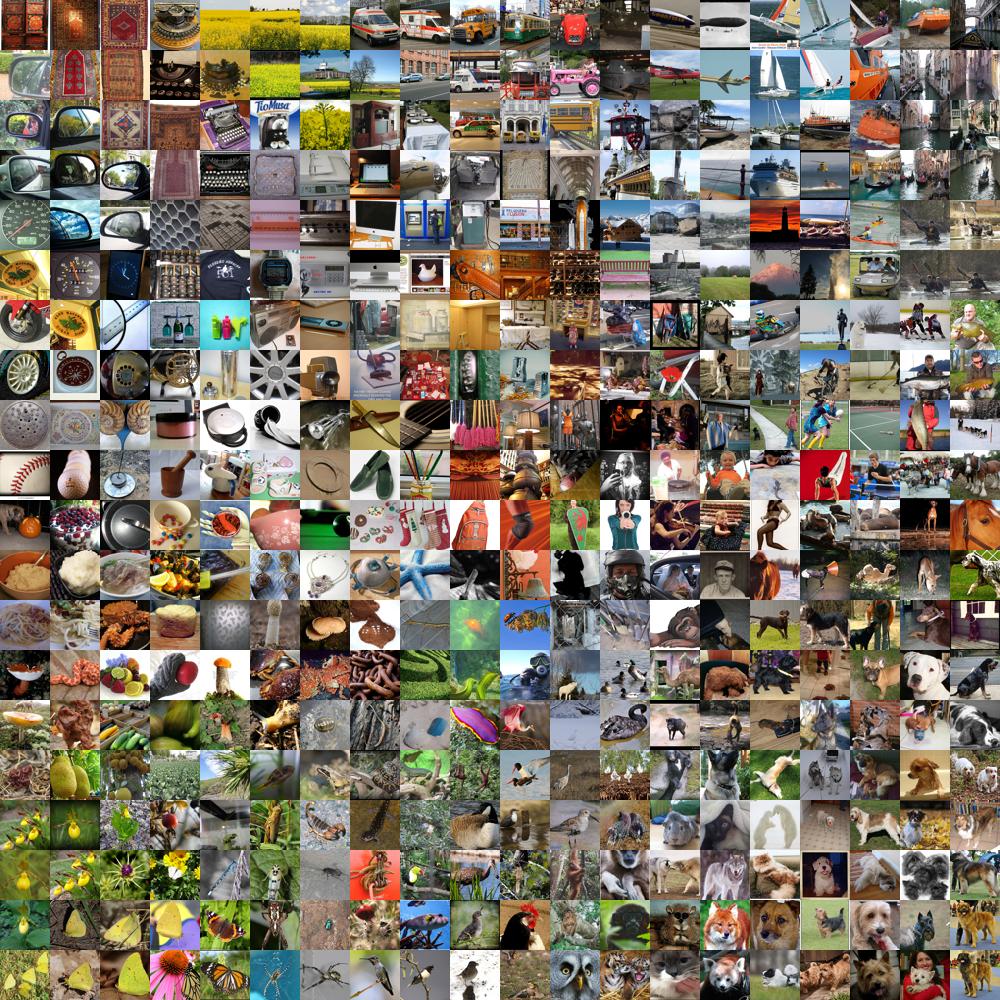}
    \caption{$1000$ samples from the ImageNet dataset}
    \label{fig:data_imagenet_samples}
\end{figure}

\section{Robotic Vision: RGB-D recognition}
\label{sec:robo_related}

% \subsection{Transfer Learning}
\textit{
We start this section by explaining why RGB-D recognition is important and how it is related to this thesis. We then give an overview of how this problem has been tackled by others and, in the end, we present the RGB-D datasets commonly used by the community which we will use to evaluate our proposed solutions.}
\newline

\subsection{Problem statement}
% RGB-D data is almost always available to robotic systems, thanks to the low cost of depth sensors (i.e. Microsoft Kinect), but it is rarely used for recognition. This is due to many reasons, the main one being that there exists no pretrained model for Depth data and existing datasets are too small to train from scratch.
\begin{figure}
    \centering
    \includegraphics[trim=0 0 0 1cm,clip,width=0.4\textwidth]{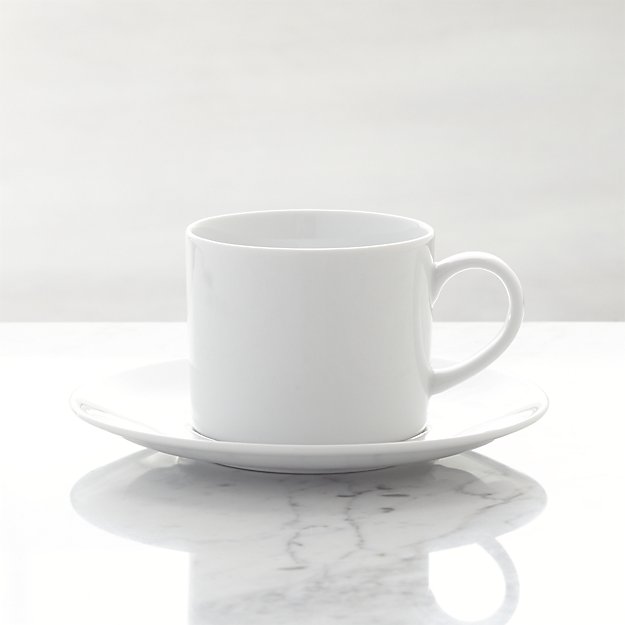}
    \includegraphics[width=0.4\textwidth]{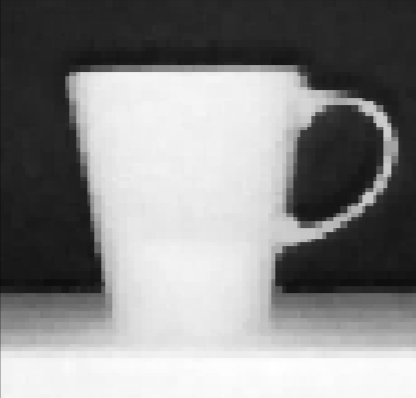}
    \caption{To the left: an RGB image of a white cup, on a white plate, sitting on a white table with a white background. Similar scenes lack in texture and are extremely hard to correctly classify. To the right: the depth image of a similar scene. Note how the cup is much more visible against the background and the gradient on the table gives us geometrical intuition.  }
    \label{fig:white_cup}
\end{figure}

Perception of the outside world is probably one of the key ingredients for the success of autonomous robots interacting with an unstructured and unconstrained environment. Planning, reasoning and accurate motion control all depend on the agent having a correct internal representation of the space it's moving in. 
Adding multiple sensing modalities is one way to make the perception task easier.
For example, RGB recognition systems have been know to struggle with low-contrast, textureless, images, such as the left picture in Fig. \ref{fig:white_cup}, but depth cameras can produce images of the same scene which are much easier to interpret.
Indeed, due to their low cost and ease of use, depth cameras are one of the most commonly available sensors in robotics. Depth is used for many things\cite{han2013enhanced}, from SLAM \cite{endres2012evaluation} to segmentation\cite{hermans2014dense,hernandez2012detecting}, to pose estimation\cite{rafi2015semantic,keskin2013real}. As discussed above, the depth modality can also clearly contribute to the recognition performance, as it can perceive different features (ex. geometrical cues) from the RGB ones. 

In section \ref{ss:robo_related_real} we will survey most recent works dealing with RGB-D recognition. Almost all of them share one commonality: they use an ImageNet pretrained network for RGB perception and then exploit multiple workarounds to deal with the Depth modality. Why is this necessary? Simply put, most RGB-D datasets are small (in semantic variability if not in sheer size) and a pretrained network of some type is needed to avoid overfitting issues. Here lies the problem: it is extremely unlikely we will ever have an RGB-D dataset comparable in size to ImageNet.

Most large RGB datasets (ImageNet included) were farmed from the web, which is a true treasure for this kind of data - but datasets for other modalities (Depth, Infrared, Multispectral cameras) will never be found online with the same ease. For these cases we must accept that we either train a model on synthetic data (which we explore in section \ref{sec:depthnet}) or we find a smart, possibly data-driven, way to exploit what we learned on RGB for these other modalities (which we will describe in section \ref{sec:deco}).

Classical DA approaches are not easily applicable here: firstly, our \textit{target}, the Depth modality, is fully labeled (so even semi-supervised approaches do not really fit) and secondly our images are not immediately usable by RGB pretrained models, at least without some sort of preprocessing.

\subsection{Literature survey}
\label{ss:robo_related_real}
\begin{figure}
    \centering
    \includegraphics[width=1.0\textwidth]{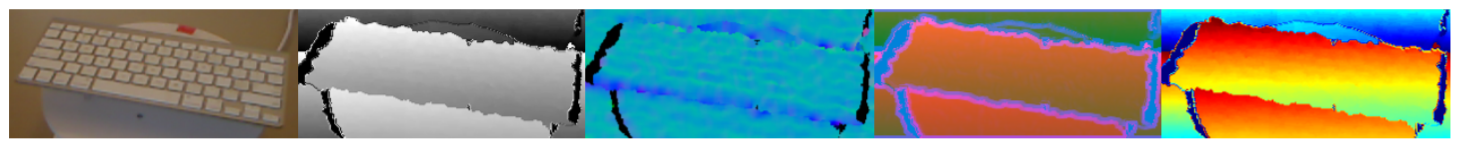}
    \caption{Different approaches for color encoding of depth images. From left to right: RGB, depth-gray, surface normals, HHA, colorjet. (image from "Multimodal Deep Learning for Robust RGB-D Object Recognition"~\cite{eitel2015multimodal}}
    \label{fig:eitel_mappings}
\end{figure}
Object recognition from RGB-D data traditionally relied on hand-crafted features such as SIFT \cite{lowe2004distinctive} and spin images \cite{lai2011large}, combined together through vector quantization in a Bag-of-Words encoding \cite{lai2011large}. This heuristic approach has been surpassed by end-to-end feature learning architectures, able to define suitable features in a data-driven fashion   \cite{blum2012learned,socher2012convolutional,asif2015efficient}. All these methods have been designed to cope with a limited amount of training data (of the order of $10^3-10^4$ depth images), thus they are able to only partially exploit the generalization abilities of deep learning as feature extractors experienced in the computer vision community \cite{krizhevsky2012imagenet,sharif2014cnn}, where databases of $10^6$ RGB images like ImageNet \cite{russakovsky2015imagenet} or Places \cite{zhou2014learning} are available.

An alternative route is that of re-using deep learning architectures trained on ImageNet through pre-defined encoding \cite{gupta2014learning} or colorization. Colorization of depth images can be seen as a transfer learning process across modalities, and several works explored this avenue within the deep learning framework: 
%since\cite{schwarz2015rgb} re-defined the state of the art in the field, this last approach has been actively and successfully investigated. Eitel et al \cite{eitel2015multimodal} proposed a parallel CNN architecture, one for the depth channel and one for the RGB one, combined together in the final layers through a late fusion scheme. 
HHA, a method proposed by Gupta, Saurabh, et al. \cite{gupta2014learning}, is a mapping where one channel encodes the horizontal disparity, one the height above ground and the third the pixelwise angle between the surface normal and the gravity vector. 
Schwarz et al ~\cite{schwarz2015rgb} proposed a  colorization pipeline where colors are assigned to the image pixels according to the distance of the vertexes of a rendered mesh to the center of the object. Bo et al.\cite{Bo} convert the depth map to surface normals and then re-interprets it as RGB values, while Aekerberg et al. \cite{aakerberg2017depth} builds on this approach and suggests an effective preprocessing pipeline to increase performance.
Besides the naive grayscale method, the rest of the mentioned colorization schemes are computationally expensive.  Eitel et al ~\cite{eitel2015multimodal} used a color mapping technique known as \textit{ColorJet},  showing this simple method to be competitive with more sophisticated approaches.

In the context of RGB-D object detection, a recent stream of works explicitly addressed cross modal transfer learning through sharing of weights across architectures \cite{hoffman2016icra}, \cite{hoffman2016learning} and \cite{Gupta_2016_CVPR}. 
This last work is conceptually close to one approach we will present in section \ref{sec:deco}, as it proposes to learn how to transfer RGB extracted information to the Depth domain through distillation \cite{hinton2015distilling}. While  \cite{Gupta_2016_CVPR} has proved very successful in the object detection realm, it presents some constraints that might potentially be problematic in object recognition, from the requirement of paired RGB-D images, to specific  data preprocessing and preparation for training. As opposed to this, our algorithm does not require explicit pairing of images in the two modalities, can be applied successfully on raw pixel data and does not require other data preparation for training.

Some approaches coupled non linear learning methods with various forms of spatial encodings \cite{cheng2015semi,cheng2014semi,cheng2015convolutional,li2015beyond}.
Hasan et al \cite{zaki2016convolutional} pushed further this multi-modal approach, proposing an architecture merging together RGB, depth and 3D point cloud information. Another notable feature is the encoding of an implicit multi scale representation through a rich coarse-to-fine feature extraction approach.

All these works, and others \cite{zaki2016convolutional,carlucci2016deep}, make use of an ad-hoc mapping for converting depth images into three channels.
This conversion is vital as the dataset has to be compatible with the pre-trained CNN. Depth data is encoded as a 2D array where each element represents an approximate distance between the sensor and the object. Depth information is often depicted and stored as a single monochrome image.  Compared to regular RGB cameras, the depth resolution is relatively low, especially when the frame is cropped to focus on a particular object. 
In section \ref{sec:deco} we will address this issue, by avoiding heuristic choices in our approach and instead relying instead on an end-to-end, residual based deep architecture to learn the optimal mapping for the cross modal knowledge transfer.

% Most of works in object recognition, against whom we will compare, are evaluated on one single database, with Washington being the standard choice in the robot vision literature. This raises concerns about the generality of these methods, especially considering their hand-crafted nature. We circumvent this issue by evaluating \DECO\ on three different databases.

Our work is also related to the colorization of grayscale images using deep nets. 
Cheng et al ~\cite{cheng2015deep} proposed a colorization pipeline based on three different hand-designed feature extractors to determine the features from different levels of an input image.  Larsson et al ~\cite{larsson2016learning} used an architecture consisting of two parts. The first part is a fully convolutional version of VGG-16 used as  feature extractor, and the second part is a fully-connected layer with 1024 channels predicting the distributions of hue and the chroma for each pixel given its feature descriptors from the previous level. 
Iizuka et al ~\cite{IizukaSIGGRAPH2016} proposed an end-to-end network  able to learn global and local features,  exploiting the classification labels for better image colorization. Their architecture  consists of several networks followed by fusion layer for the colorization task.
%\textbf{ADDED new/2017 literature}
Sun et al. \cite{sun2017weakly} propose to use large scale CAD rendered data to leverage depth information without using low level features or colorization. In Asif et al. \cite{asif2017rgb}, hierarchical cascaded forests were used for computing grasp poses and perform object classification, exploiting several different features like orientation angle maps, surface normals and depth information colored with \textit{Jet} method.
Our work differs from this last research thread in the specific architecture proposed, and in its main goal, as here we are interested in learning optimal mapping for categorization rather than for colorization of grayscale images. 

All these works build on top of CNNs pre-trained over ImageNet, for all modal channels. Thus, the very same filters are used to extract features from all of them. As empirically successful as this might be, it is a questionable strategy, as RGB and depth images are perceptually very different, and as such they would benefit from approaches able to learn data-specific features (fig. \ref{fig:depthnet_first_layer_weights}). 

Another possibility, which we explore in section \ref{sec:depthnet} is to train the model on synthetic data; it is usually much cheaper to build a computer generated dataset than a real one and, often times, it allows for a greater variability.
Our method matches this challenge, learning RGB features from RGB data and depth features from synthetically generated data, within a deep learning framework. The use of realistic synthetic data in conjunction with deep learning architectures is a promising emerging trend \cite{papon2015semantic,maturana2015voxnet,wu20153d,handa2014benchmark}. We are not aware of previous work attempting to use synthetic data to learn depth representations, with or without deep learning techniques. 

\subsection{Datasets}
\label{sec:rgbd_datasets}
\textit{The datasets and the protocols used to evaluate the methods proposed in chapter \ref{chap:auxiliary} are listed here}
\newline

\begin{figure}
    \centering
    \includegraphics[width=1.0\textwidth]{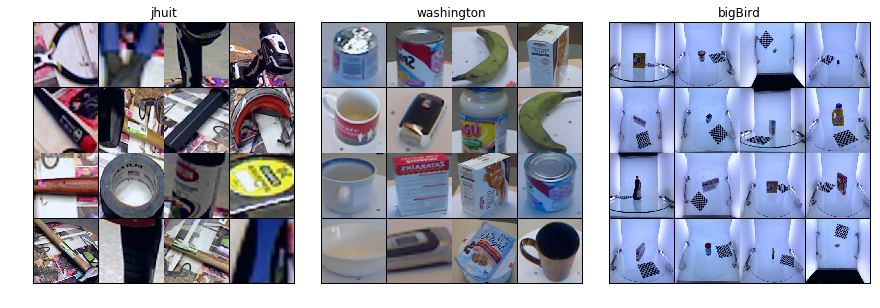}
    \caption{Samples from the Jhuit\cite{jhuit}, Washington-50\cite{washington} and BigBIRD\cite{singh2014bigbird} datasets, RGB modality. Note how items in the BigBIRD datasets are very small: for all our Depth recognition experiments we used the provided masks to crop the images (see Fig. \ref{fig:data_rgbd_d_samples}).}
    \label{fig:data_rgbd_samples}
\end{figure}

\begin{figure}
    \centering
    \includegraphics[width=1.0\textwidth]{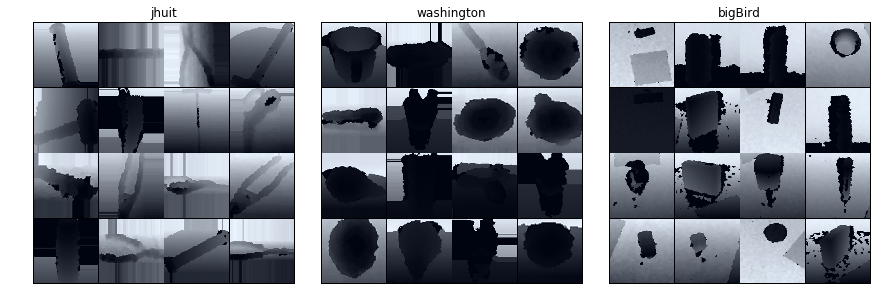}
    \caption{Samples from the Jhuit\cite{jhuit}, Washington\cite{washington} and BigBIRD\cite{singh2014bigbird} datasets, Depth modality. For visualization purposes, the 16bit raw data has been normalized to 8bits.}
    \label{fig:data_rgbd_d_samples}
\end{figure}

We considered three datasets (samples can be seen in Fig \ref{fig:data_rgbd_samples} and \ref{fig:data_rgbd_d_samples}): the \textbf{Washington RGB-D} \cite{washington}, the \textbf{JHUIT-50} \cite{jhuit} and the \textbf{BigBIRD} \cite{singh2014bigbird} object datasets, which are the main public datasets for RGB-D object recognition. Each of them contains paired RGB and Depth images.

The first consists of $41,877$ RGB-D images organized into $300$ instances divided in $51$ classes. 
We performed experiments on the object categorization setting, where we followed the evaluation protocol defined in \cite{washington}. Following the guidelines defined by the authors\cite{washington} we crop the images with the provided bounding boxes.

The JHUIT-50  is a challenging recent dataset that focuses on the problem of fine-grained classification.  It contains $50$ object instances, often very similar with each other (e.g. 9 different kinds of screwdrivers). As such, it presents different recognition challenges compared to the Washington database. Here we followed the evaluation procedure defined in \cite{jhuit}.

BigBIRD is the biggest of the datasets we considered: it contains $121$ object instances and $75.000$ images. Unfortunately, it is an extremely unforgiving dataset for evaluating depth features: many objects are extremely similar, and many are boxes, which are indistinguishable without texture information. To partially mitigate this, we grouped together all classes annotated with the same first word:
%containing the same four starting letters: f
for example \textit{nutrigrain apple cinnamon} and \textit{nutrigrain blueberry} were grouped into \textit{nutrigrain}. With this procedure, we reduced the number of classes to 61 (while keeping all of the samples). 
As items are quite small (check the rightmost image in Fig.\ref{fig:data_rgbd_samples})
, we used the object masks provided by \cite{singh2014bigbird} to crop around the object (results are in Fig. \ref{fig:data_rgbd_d_samples}). Evaluation-wise, we followed the  protocol defined in \cite{jhuit}.

% \subsection{Synthetic Data Generation}
% \input{depthnet_related}

\chapter{Learning to see across domains}
\label{chap:DA} 
\textit{This chapter presents different strategies that allow us to deal with the Unsupervised Domain Adaptation problem. Three methods are presented: \DIAL, SBADA-GAN and ADAGE. The first uses a novel layer, the Domain Alignment layer, which structurally enforces statistic similarity between domains by applying a transformations that maps all data sources into a predefined distribution. The second is GAN\cite{Goodfellow:GAN:NIPS2014} based method which tackles the domain shift issue in image space. The last method tweaks both the images and the feature representation to have all domains lie in one common, agnostic, subspace.}

\section{Feature Alignment: \DIAL}
\label{sec:da_feature_align}
\textit{One of the main challenges for developing visual recognition systems working in the wild is to devise computational models immune from the domain shift problem, \ie accurate when test data are drawn from a (slightly) different data distribution than training samples. In the last decade, several research efforts have been devoted to devise algorithmic solutions for this issue. Recent attempts to mitigate domain shift have resulted into deep learning models for domain adaptation which learn domain-invariant representations by introducing appropriate loss terms, by casting the problem within an adversarial learning framework or by embedding into deep network specific domain normalization layers. 
This section describes a novel approach for unsupervised domain adaptation. Similarly to previous works we propose to align the learned representations by embedding them into appropriate network feature normalization layers. Opposite to previous works, our \emph{Domain Alignment Layers} are designed not only to match the source and target feature distributions but also to \emph{automatically} learn the degree of feature alignment required at different levels of the deep network. 
Differently from most previous deep domain adaptation methods, our approach is able to operate in a multi-source setting. Thorough experiments on four publicly available benchmarks confirm the effectiveness of our approach.}
\newline
%%%%%%%%%%%%%% Intro %%%%%%%%%%%%%%%%%%%%%

The quest to create intelligent systems able to adapt to varying environmental conditions is a deep-seated human pursuit. Focusing on computer vision, adaptation is a crucial aspect, as visual recognition systems are required to operate under different conditions, corresponding to varying illuminations, changing point of view, diverse image resolution, etc. Therefore, it is not surprising that over the years researchers have devoted significant efforts into developing algorithms and techniques for domain adaptation. 

Domain adaptation methods attempt to address the so-called \textit{domain shift} problem, \ie the fact that in many real world applications there is a discrepancy between the distributions of training and test samples. 
This distribution mismatch adversely affects the performance of visual recognition models.
Despite the significant advances 
%in computer vision 
brought by deep learning, several works have shown that the domain shift problem is alleviated when using deep feature representations, but not solved \cite{Transfer2}.

In the last few years the research community has devoted significant efforts in addressing %domain shift. 
this issue.
In particular, several previous studies have considered the problem of \textit{unsupervised domain adaptation}, where only the training samples from a source domain have labels and the goal is to predict the labels of test samples in a target domain.
%In this context, the specific problem of unsupervised domain adaptation, \ie no labeled data are available in the target domain, deserves special attention.
This problem is highly relevant in many applications as annotating data is not only a costly and time-consuming operation but, in some cases, it is not possible at all.
Several approaches have been proposed for unsupervised domain adaptation, both considering hand-crafted features~\cite{huang2006correcting,gong2013connecting,gong2012geodesic,long2013transfer,fernando2013unsupervised} and learned deep representations~\cite{long2015learning,tzeng2015simultaneous,ganin2014unsupervised,long2016unsupervised,ghifary2016deep,li2016revisiting,carlucci2017just}.
Focusing on deep architectures, several strategies have been proposed in the last few years.
Some methods attempt to reduce the discrepancy among source and target distributions by learning features which are invariant to the domain shift. Earlier works propose to minimize the Maximum Mean Discrepancy (MMD)~\cite{long2015learning,long2016unsupervised}: the distributions of the learned source and target representations are optimized to be as similar as possible by minimizing the distance between their mean embeddings.
Other studies \cite{tzeng2015simultaneous,ganin2014unsupervised} consider a domain-confusion loss, \ie learn domain-agnostic representations by introducing an auxiliary classifier predicting if a sample comes from the source or from the target domain.
More recently, researchers have also investigated alternative directions. For instance, Ghifary \textit{et al.}~\cite{ghifary2016deep} designed an encoder-decoder network to jointly learn source labels and reconstruct unsupervised target images. Some methods tackle the domain shift problem by transforming images using Generative Adversarial Networks (GANs) \cite{Bousmalis:Google:CVPR17, Shrivastava:arXiv:16, Taigman2016UnsupervisedCI,hoffman2017cycada,russo17sbadagan}. 
\cite{bousmalis2016domain,li2016revisiting,carlucci2017just}, Other works investigate the possibility of reducing the distribution discrepancy among source and target domain by embedding into a neural network specific distribution normalization layers ~\cite{li2016revisiting,carlucci2017just,mancini2018boosting,mancini2018kitting}. These methods are especially interesting since, differently from most previous works learning deep domain-invariant features, they do not consider in the optimization additional loss terms for minimizing the distance among distributions (\eg MMD, domain-confusion), thus they do not require to tune the associated hyper-parameters. Furthermore, opposite to previous approaches based on GANs, the optimization process has more favorable convergence properties.

\begin{figure}[tb]
  \centering
  \includegraphics[width=0.9\textwidth]{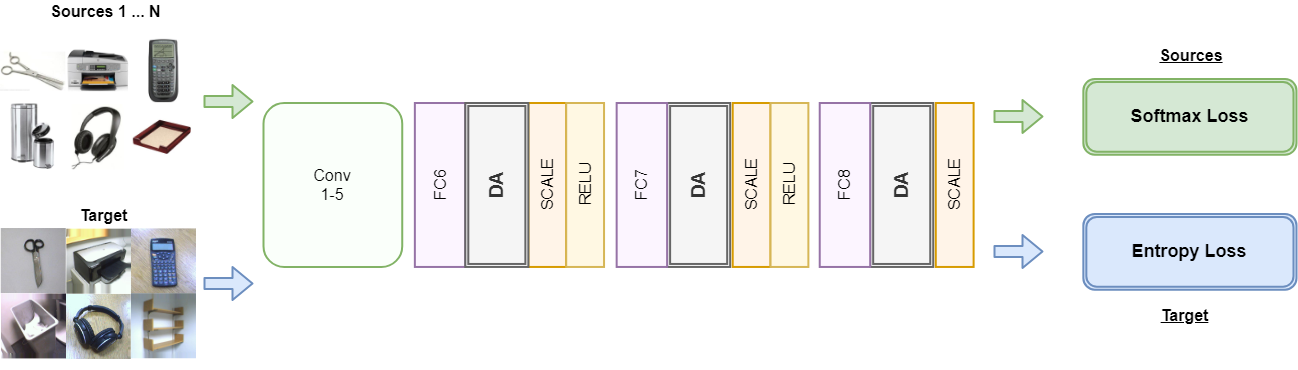}
  \caption{\DIAL as applied on \Alex~\cite{krizhevsky2012imagenet}. Source and target images are provided as input to the network. After passing through the first layers, they enter our \DAL where source and target distributions are aligned. 
  The DA-Layer is shown in detail in Fig. ~\ref{fig:da_detail}.
  After flowing through the whole network, source samples contribute to a Softmax loss, while target samples contribute to an Entropy loss, which promotes classification models which maximally separate unlabeled data. Note that we use multiple \DALs to align learned feature representations at different levels.}
  \label{fig:dial_teaser}
\end{figure}
\begin{figure}
\centering
\includegraphics[width=0.8\columnwidth]{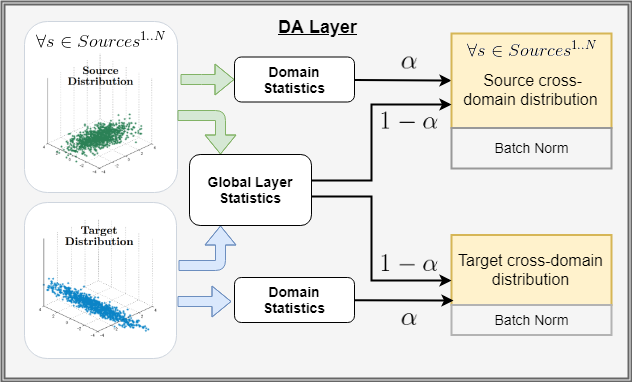}
\caption{The DA-layer learns the global statistics of all domains and normalizes the source and target mini-batches according to the computed mean and variance, different for each domain (see subsection ~\ref{sec:dial_predictors}). The amount by which each distribution is influenced by the global one and therefore the degree of domain alignment, depends on a parameter, $\alpha \in [0.0, 1.0]$, which is also automatically learned.}
\label{fig:da_detail}
\end{figure}

The approach described in this section belongs to the latter category of methods. Specifically, we introduce novel \textit{Domain Alignment} layers (\DALs) (Fig.~\ref{fig:dial_teaser} and ~\ref{fig:da_detail}) which are embedded at different levels of a deep architecture to align the learned source and target feature distributions to a canonical one.
Different from previous works based on distribution normalization layers \cite{li2016revisiting, mancini2018boosting,mancini2018kitting} which decide \emph{a priori} which layers should be adapted, we endow our DA-layers with the ability to \emph{automatically} learn the degree of alignment that should be pursued at different levels of the network.
This is to our knowledge the first work that tries to pursue this objective. 
Furthermore, inspired by ~\cite{grandvalet2004semisupervised}, we propose to leverage information from the target domain to construct a prior distribution on the network parameters, biasing the learned solution towards models that are able to separate well the classes in the target domain (see subsection~\ref{ss:training}).
Our DA-layers and the considered prior distribution work in synergy during learning: the first aligning the source and target feature distributions, the second encouraging the network to learn features that lead to maximally separated target classes.
We call our algorithm \DIAL -- DomaIn Alignment Layers with Automatic alignment parameters.
Our extensive experimental evaluation demonstrates that \DIAL greatly alleviates the domain discrepancy and outperforms state of the art techniques on popular domain adaptation benchmarks: Office-31~\cite{saenko2010adapting}, Office-Caltech~\cite{gong2012geodesic}, the Caltech-ImageNet setting of the Cross-Dataset Testbed\cite{tommasi2014testbed} and the recent Office-Home \cite{venkateswara2017deep} dataset.

This section extends our earlier works \cite{carlucci2017auto,carlucci2017just} through considering a multi-source domain adaptation setting. In particular we modify the proposed DA-layers in order to enable feature normalization of data coming from multiple domains. Moreover, we enrich the related works, provide additional details on the proposed method and significantly expand our experimental results and analysis considering an additional large scale dataset (\ie the Office-Home \cite{venkateswara2017deep} dataset) and more recent deep architectures. 

To summarize, our contribution in this work are threefold.
First, we present an approach for unsupervised domain adaptation, based on the introduction of DA-layers to explicitly address the domain shift problem, which act in synergy with an entropy loss which exploits unsupervised target data during learning.
Our solution simultaneously aligns feature representations and learns where and to which extent adaptation should take place.
We also show that the proposed approach can be naturally extended to a multi-source setting.
Second, in contrast to previous works optimizing domain discrepancy regularization terms~\cite{long2016unsupervised,tzeng2015simultaneous,ganin2014unsupervised,long2015learning}, our DA-layers do not require any additional hyper-parameters.
Third, we perform an extensive experimental analysis on four different benchmarks.
We find that our unsupervised domain adaptation approach outperforms state-of-the-art methods and can be applied to different CNN architectures, consistently improving their performance in domain adaptation problems.

The remainder of this section is organized as follows.
%We first introduce related work in Section ~\ref{related}. 
The proposed DA-layers and our novel unsupervised domain adaptation approach are described in subsection ~\ref{subsec:dial_method}. The experimental results and analysis are provided in subsection ~\ref{subsec:dial_experiments}, and we conclude in subsection ~\ref{subsec:dial_conclusions}.

%%%%%%% Method %%%%%%%%%%%%%%%%%%%%%%
\subsection{Automatic DomaIn Alignment Layers}
\label{subsec:dial_method}

Let $\set X$ be the input space (\eg images) and $\set Y$ the output space (\eg image categories) of our learning task.
In the typical Unsupervised Domain Adaptation (UDA) setting, we are given labeled data points from a \emph{source} domain 
and the goal is to train a classifier for data points coming from a \emph{target} domain.
A particular variant of UDA, which will be addressed in this section, considers \emph{multiple} source domains instead of a single one. 
%The domains are identified via conditional distributions $p^s_{\mathtt{xy|d}}$ defined over $\set X\times\set Y$ and conditioned on a domain variable in $\set D$ (including source and target domains)
%are identified via probability distributions $p^s_{\mathtt{xy}}$ and $p^t_\mathtt{xy}$, respectively, defined over $\set X\times\set Y$.
Source and target distributions are in general different and unknown, but we are provided with an \emph{i.i.d} sample $\set S=\{(x_1,y_1, d_1),\dots,(x_n,y_n,d_n)\}\subset \set X\times\set Y\times\set D$ from the source domains and an unlabeled \emph{i.i.d} sample $\set T=\{\hat x_1,\dots,\hat x_m\}\subset\set X$ from the target domain, where $\set D$ is the set of source domain labels. 
%We will denote by $\set S_d\subset\set S$ the subset of sample points in $\set S$ belonging to domain $d\in\set D$.

Our goal is to estimate a predictor from $\set S$ and $\set T$ that can be used to classify sample points from the target domain.
This task is particularly challenging because on one hand we lack direct observations of labels from the target domain and on the other hand the discrepancy between the source and target domain distributions prevents a predictor trained on $\set S$ to be readily applied to the target domain. 

A number of state of the art methods try to reduce the domain discrepancy by performing some form of alignment at the feature or classifier level.
In particular, the recent, most successful methods try to \emph{couple} the training process and the domain adaptation step within \emph{deep} neural architectures~\cite{ganin2014unsupervised,long2016unsupervised,long2015learning}, as this solution enables alignments at different levels of abstraction.
The approach we propose embraces the same philosophy, while departing from the assumption that domain alignment can by pursued by applying the \emph{same} predictor to the source and target domains.
This is motivated by an impossibility theorem~\cite{ben2010impossibility}, which intuitively states that no learner relying on the \emph{covariate shift} hypothesis, and achieving a low discrepancy between the source and target unlabeled distributions, is guaranteed to succeed in domain adaptation without further relatedness assumptions between training and target distributions.
For this reason, we assume that the source and target predictors are in general \emph{different} functions.
Nonetheless, all predictors depend on a common parameter $\theta$ belonging to a set $\Theta$, which couples them explicitly, while not being directly involved in the alignment of the source and target domains.
This contrasts with the majority of state of the art methods that augment the loss function used to train their predictors with a regularization term penalizing discrepancies between source and target representations (see, \eg~\cite{ganin2014unsupervised,long2016unsupervised,long2015learning}). 
The perspective we take is different and is close in spirit to AdaBN~\cite{li2016revisiting}.
It consists in hard-coding the desired domain-invariance properties into the source and target predictors through the introduction of so-called \emph{Domain-Alignment layers} (\DALs).
%Also, we distinguish from previous approaches in the field, by trying to sidestep the problem of deciding which layers should be aligned, and to what extent.
% We achieve this goal by endowing the architecture with the ability to \emph{automatically} tune the degree of alignment that should be considered in each domain-alignment layer.
Moreover, we sidestep the problem of deciding which layers should be aligned, and to what extent, by endowing the architecture with the ability to \emph{automatically} tune the degree of alignment that should be considered in each domain-alignment layer.
The rest of this section is devoted to providing the details of our method, which extends our previous work~\cite{carlucci2017auto} to the case of multiple source domains.

\subsubsection{Domain-Specific Predictors}
\label{sec:dial_predictors}
Each domain is associated with a predictor that is implemented as a deep neural network and each domain-specific predictor is applied only to sample points from the corresponding domain. All predictors are actually almost identical, as they share the same structure and the same weights (given by the parameter $\theta$).
However, the networks contain also a number of special layers, the \DALs, which implement a domain-specific operation.
Indeed, the role of such layers is to apply a data transformation that aligns the observed input distribution with a reference distribution.
Since in general the input distributions of the domain-specific predictors differ, while the reference distribution stays the same, we have that the predictors undergo different transformations in the corresponding \DALs.
Consequently, the source and target predictors de facto implement different functions, which is important for the reasons given in subsec.~\ref{subsec:dial_method}.

The actual implementation of our \DALs is inspired by AdaBN~\cite{li2016revisiting}, where Batch Normalization layers are used to independently align source and target distributions to a standard normal distribution, by matching the first- and second-order moments.
The approach they propose consists in training on the source domain a network having BN-layers, thus obtaining the source predictor, and deriving the target predictor as a post-processing step, which re-estimates the BN statistics using target samples only. 
Accordingly, the source and target predictors share the same network parameters but have different BN statistics, thus rendering the predictors different functions.

The approach we propose sticks to the same idea of using BN-layers to align domains, but we introduce fundamental changes.
One limitation of AdaBN is that the target observations have no influence on the network parameters, as they are not observed during training.
Our approach overcomes this limitation by coupling the network parameters to target and source observations at training time.
This is achieved in two ways: first we introduce an entropy-based prior distribution for the network parameters based on the target observations; second, we endow the architecture with the ability of learning the degree of adaptation by introducing a parametrized, cross-domain bias to the input distribution of each domain-specific \DAL. 
The rest of this subsection is devoted to describe the new layer, while we defer to the next subsection the description of the prior distribution.

\paragraph{\DAL.}
The goal of our \DALs is to align the input distribution to a reference distribution. Here, we consider the case of having a standard normal as reference distribution, but other distributions can be considered~\cite{carlucci2017just}. Under this assumption, our layer takes the same form of Batch Normalization, but as opposed to BN, which computes first and second-order moments from the input distribution -- in our case samples from the same domain in the mini-batch --, we let the latter statistics to be contaminated by samples from the other domains, thus introducing a cross-domain bias. 
Since the domain-specific predictors share the same network topology, each \DAL in one predictor has a matching \DAL in all other domain-specific predictors.
Let $x$ denote an input to the \DAL of a given domain for a given feature channel and spatial location.
Assume $q$ to be the distribution of $x$ and assume $\bar q$ to be the distribution of inputs to all matching \DALs.
Let $q_\alpha=\alpha q + (1-\alpha)\bar q$ be the cross-domain distribution mixed by a factor $\alpha\in[0,1]$, which ranges from considering only a domain-specific distribution $q$ if $\alpha=1$, to considering the mixture distribution $\bar q$ of all domains if $\alpha=0$.
Then, the output of the \DAL is given by
\begin{equation}
\label{eqn:dal}
\mathtt {DA}(x;\alpha) = \frac{x-\mu_\alpha}{\sqrt{\epsilon+\sigma_\alpha^2}},\,\,\,
\end{equation}
where $\epsilon>0$ is a small number to avoid numerical instabilities in case of zero variance, $\mu_{\alpha}=\mathtt {E}_{x\sim q_\alpha}[x]$, $\sigma^2_{\alpha}=\mathtt {Var}_{x\sim q_\alpha}[x]$.
Akin to BN, we estimate the statistics based on the mini-batch and derive similarly the gradients through the statistics (see Supplementary Material).
Note that in this work we assume the $\alpha$ parameter to be learned and shared between matching \DALs of each domain-specific predictor. However, $\alpha$'s can differ across \DALs within the same predictor.

The rationale behind the introduction of the mixing factor $\alpha$ is that we can move from having an independent alignment of the domains akin to AdaBN, when $\alpha=1$, to having a coupled normalization when $\alpha=0$.
In the former case the \DAL computes different functions in each domain-specific predictor and is equivalent to considering a full degree of domain alignment.
The latter case, instead, yields the same function in each predictor thus transforming the domains equally, which yields no domain alignment.
Since the mixing parameter $\alpha$ is not fixed a priori but learned during the training phase, we obtain as a result that the network can decide how strong the domain alignment should be at each level of the architecture where \DAL is applied.
More details about the actual CNN architectures used to implement the domain-specific predictors are given in subsection~\ref{ss:dial_setup}.

\subsubsection{Training}\label{ss:training}
During the training phase we estimate the parameter $\theta$, which holds the neural network weights shared by the domain predictors including the mixing factors $\alpha$s pertaining to the \DALs, using the observations $\set S$ from the source domains and $\set T$ from the target domain.
As we stick to a discriminative model, the unlabeled target observations cannot be employed to express the data likelihood. However, we can exploit $\set T$ to construct a prior
distribution of the parameter $\theta$. Accordingly, we shape a posterior distribution of $\theta$ given the observations $\set S$ and $\set T$ as
\begin{equation}
    P(\theta|\set S,\set T)\propto P(y_{\set S}|x_{\set S},d_{\set S},\theta)P(\theta|\set T)\,,%\prod_{(x,y)\in\set S}P(y|x,\theta)\,,
    \label{eq:posterior}
\end{equation}
where $y_{\set S}=\{y_1,\dots,y_n\}$ and $x_{\set S}=\{x_1,\dots,x_n\}$ collect the labels and data points of the observations in $\set S$, respectively, and $d_\set S=\{d_1,\ldots,d_n\}$ collects the source domain labels.
For notational convenience, we dropped dependences that are induced by \DALs only and we will keep this simplification in the rest of the section (\eg the prior term in~\eqref{eq:posterior} depends also on $x_\set S$ and $d_\set S$ and the likelihood term depends also on $\set T$). 
The posterior distribution is maximized over $\Theta$ to obtain a maximum a posteriori estimate $\hat\theta$ of the parameter used in the source and target predictors:
\begin{equation}
    \theta^* \in\argmax_{\theta\in\Theta} P(\theta|\set S,\set T)\,.
    \label{eq:MAP}
\end{equation}
The term $P(y_{\set S}|x_{\set S},d_\set S,\theta)$ in~\eqref{eq:posterior} represents the likelihood of $\theta$ with respect to observations from the source domains,
while
%given a sample $(x,y)$ from the source domain. This corresponds to \eg the $y$th output element of a neural network having a soft-max layer on top. The prior term 
$P(\theta|\set T)$ is the prior term depending on the target domain sample, which acts as a regularizer in the classical learning theory sense. Both terms actually, depend on all domains due to the cross-domain statistics that we have in our \DALs for $0\leq\alpha<1$ and are estimated from observations from the source \emph{and} target domains.

The likelihood decomposes into the following product over observation, due to the data sample \emph{i.i.d.} assumption:
\begin{equation}
    P(y_{\set S}|x_{\set S},d_\set S,\theta)=\prod_{i=1}^nf_{d_i}^\theta(y_i;x_i)\,,
    \label{eq:likelihood}
\end{equation}
where $f^\theta_{d_i}(y_i;x_i)$ is the probability that sample point $x_i$ takes label $y_i$ according to the predictor for domain $d_i$.

Before delving into the details of the prior term, we would like to remark on the absence of an explicit component in the probabilistic model that tries to align the source and target domains.
This is because in our model the domain-alignment step is taken over by each predictor, independently, via the domain-alignment layers as shown in the previous subsection.

\paragraph{Prior distribution.}
The prior distribution of the parameter $\theta$ shared by the source and target predictors is constructed from the observed, target data distribution. This choice is motivated by the theoretical possibility of squeezing more bits of information from unlabeled data points insofar as they exhibit low levels of class overlap~\cite{oneill1978normal}. Accordingly, it is reasonable to bias a priori a predictor based on the degree of label uncertainty that is observed when the same predictor is applied to the target samples. Uncertainty in this sense can be measured for an hypothesis $\theta$ in terms of the following empirical entropy of the target predictor
\begin{equation}
    h(\theta|\set T)=-\frac{1}{m}\sum_{i=1}^m\sum_{y\in\set Y}f^\theta_t(y;\hat x_i)\log f^\theta_t(y;\hat x_i)\,,
    \label{eq:h}
\end{equation}
where $f_t(y;\hat x_i)$ represents the probability that sample point $\hat x_i$ takes label $y$ according to the target predictor.

It is now possible to derive a prior distribution $P(\theta|\set T)$ in terms of the label uncertainty measure $h(\theta|\set T)$
by requiring the prior distribution to maximize the entropy under the constraint $\int h(\theta|\set T)P(\theta|\set T)d\theta=\varepsilon$, 
where the constant $\varepsilon>0$ specifies how small the label uncertainty should be on average. This yields a concave, variational optimization problem with solution:
\begin{equation}
    P(\theta|\set T)\propto\exp\left( -\lambda\, h(\theta|\set T) \right)\,,
    \label{eq:prior}
\end{equation}
where $\lambda$ is the Lagrange multiplier corresponding to $\varepsilon$. 
%This prior distribution has been introduced in the context of semi-supervised learning in~\cite{grandvalet2004semisupervised} and % and employed for unsupervised domain adaptation in~\cite{long2016unsupervised}.
The resulting prior distribution  
satisfies the desired property of preferring models that exhibit well separated classes (\ie having lower values of $h(\theta|\set T)$), thus enabling 
the exploitation of the information content of unlabeled target observations within a discriminative setting~\cite{grandvalet2004semisupervised}.

Prior distributions of this kind have been adopted also in other works~\cite{long2016unsupervised} in order to exploit more information from the target distribution, but have never been used before in conjunction to explicit domain alignment methods (\ie not based on additional regularization terms such as MMD and domain-confusion) like the one we are proposing.
%\elisa{motivation}

\paragraph{Inference.}
Once we have estimated the optimal network parameters $\theta^*$ by solving~\eqref{eq:MAP}, we can remove the dependence of the target predictor on $\set T$ on source domain observations.
In fact, after fixing $\theta^*$, the input distribution to each \DAL also becomes fixed, and we can thus compute and store the required statistics once at all, akin to standard BN. 

\subsubsection{Implementation Notes}
\DAL can be implemented as a mostly straightforward modification of standard Batch Normalization.
We refer the reader to the supplementary material for a complete derivation.
In our implementation in particular, we treat each set of matching \DALs  in the domain specific predictors as a single network layer which simultaneously computes the normalization functions in Equation~\eqref{eqn:dal} for all domains and learns the $\alpha$ parameter.
During training, each \DAL receives the domain labels corresponding to its input samples, and is thus able to easily differentiate between them.
Similarly to standard BN, we keep separate moving averages of each domain's statistics and for the global statistics.
Since $\alpha$ is constrained to $[0,1]$, we clip its value in the allowed range in each forward pass of the network.

By replacing the optimization problem in~\eqref{eq:MAP} with the equivalent minimization of the negative logarithm of $P(\theta|\set S, \set T)$ and combining \eqref{eq:posterior}, \eqref{eq:likelihood}, \eqref{eq:h} and \eqref{eq:prior} we obtain a loss function $L(\theta)=L^s(\theta) + \lambda L^t(\theta)$, where:
\begin{align*}
\label{eq:loss}
  L^s(\theta) &= - \frac{1}{n}\sum_{i=1}^n \log f_{d_i}^\theta(y_i;x_i)\,,\\
  L^t(\theta) &= - \frac{1}{m}\sum_{i=1}^m\sum_{y\in\set Y}f_t^\theta(y;\hat x_i)\log f_t^\theta(y;\hat x_i)\,.
\end{align*}
The term $L^s(\theta)$ is the standard log-loss applied to the source samples, while $L^t(\theta)$ is an entropy loss applied to the target samples.
The second term can be implemented by feeding $f_t^\theta(y;x^t_i)$ to both inputs of a cross-entropy loss layer, where supported by the deep learning toolkit of choice.
% In our implementation, based on Caffe~\cite{jia2014caffe} \samuel{o PyTorch?}, we obtain it by slightly modifying the existing \texttt{SoftmaxLoss} layer.\footnote{The source code is available at \url{https://github.com/ducksoup/autodial}}

\subsection{Experiments}
\label{subsec:dial_experiments}

In this section we extensively evaluate our approach and compare it with state of the art unsupervised domain adaptation methods.
We also provide a detailed analysis of the proposed framework, performing a sensitivity study and demonstrating empirically the effect of our contributions.
Note that all the results in the following are reported as averages over five training/testing runs.
Full details on the datasets and the used protocol can be found in subsection ~\ref{ss:da_real_datasets}.

\subsubsection{Networks and Training}
\label{ss:dial_setup}
We apply the proposed method to three widely used CNNs architectures, \ie AlexNet~\cite{krizhevsky2012imagenet}, \VGGf{}\cite{chatfield2014return}, and Inception-BN~\cite{ioffe2015batch}.
We train our networks using mini-batch stochastic gradient descent with momentum, as implemented in the Caffe library, using the following meta-parameters: weight decay $5\times 10^{-4}$, momentum $0.9$, initial learning rate $10^{-3}$.
We augment the input data by scaling all images to $256\times 256$ pixels, randomly cropping $227\times 227$ pixels (for AlexNet and \VGGf) or $224\times 224$ pixels (Inception-BN) patches and performing random flips.
In all experiments we choose the parameter $\lambda$ by cross-validation on the source set according to the protocol in~\cite{long2016unsupervised}.

AlexNet~\cite{krizhevsky2012imagenet} and \VGGf~\cite{chatfield2014return} are two well-know architectures with five convolutional and three fully-connected layers, denoted as \texttt{fc6}, \texttt{fc7} and \texttt{fc8}.
The outputs of \texttt{fc6} and \texttt{fc7} are commonly used in the domain-adaptation literature as pre-trained feature representations~\cite{Transfer2,sun2016return} for traditional machine learning approaches.
In our experiments we modify both architectures by appending a \DAL to each fully-connected layer.
Differently from the original networks, we \emph{do not} perform dropout on the outputs of \texttt{fc6} and \texttt{fc7}.

We initialize the network parameters from a publicly-available model trained on the ILSVRC-2012 data, we freeze all the convolutional layers, and increase the learning rate of \texttt{fc8} by a factor of $10$.
During training, each mini-batch contains a number of source and target samples proportional to the size of the corresponding dataset, while the batch size remains fixed at 256.
We train for a total of 60 epochs (where ``epoch'' refers to a complete pass over the source set), reducing the learning rate by a factor $10$ after 54 epochs.

Inception-BN~\cite{ioffe2015batch} is a very deep architecture obtained by concatenating ``inception'' blocks.
Each block is composed of several parallel convolutions with batch normalization and pooling layers.
To apply the proposed method to Inception-BN, we replace each batch-normalization layer with a \DAL.
Similarly to AlexNet, we initialize the network's parameters from a publicly-available model trained on the ILSVRC-2012 data and freeze the first three inception blocks.
The $\alpha$ parameter is also fixed to a value of $0$ in the \DALs of the first three blocks, which is equivalent to preserving the original batch normalization layers.
Due to GPU memory constraints, we use a much smaller batch size than for AlexNet and fix the number of source and target samples in each batch to, respectively, 32 and 16.
In the Office-31 experiments we train for 1200 iterations, reducing the learning rate by a factor 10 after 1000 iterations.
% \textbf{README: questo è il protocollo seguito in DIAL, vogliamo scriverlo? e' un po' strano che qui usiamo le epoche quando per il resto usiamo le iterazioni $\rightarrow$ }
%In the reference distribution ablation we instead train for 20 epochs, reducing the learning rate by a factor 10 every $33\,\%$ of the total number of iterations.
In the Cross-Dataset Testbed experiments we train for 2000 iterations, reducing the learning rate after 1500.

\subsubsection{Analysis of the proposed method}
\label{sec:dial_results-analysis}

\begin{figure}
  \centering
  \includegraphics[width=0.8\columnwidth]{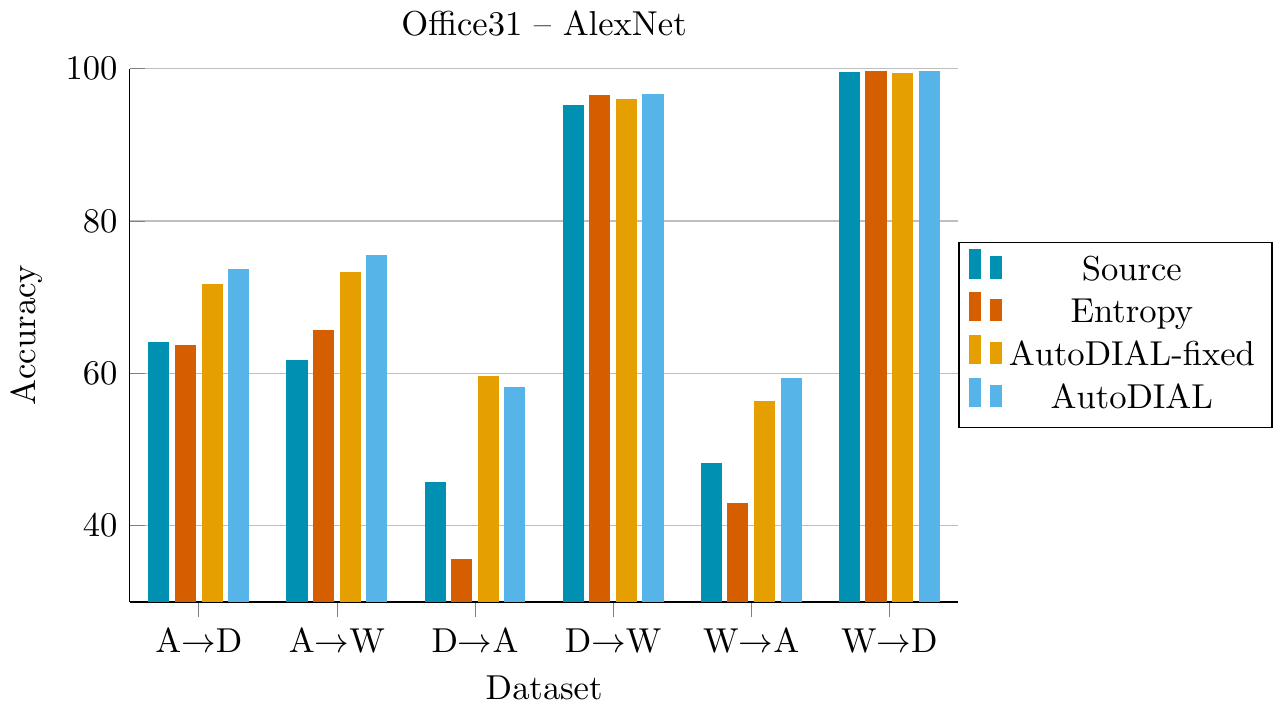}
  \caption{Accuracy on the Office31 dataset when considering different architectures based on AlexNet. 
  }
  \label{fig:dial_ablation}
\end{figure}

\begin{figure*}[ht!]
  \centering
  \includegraphics[height=4.3cm]{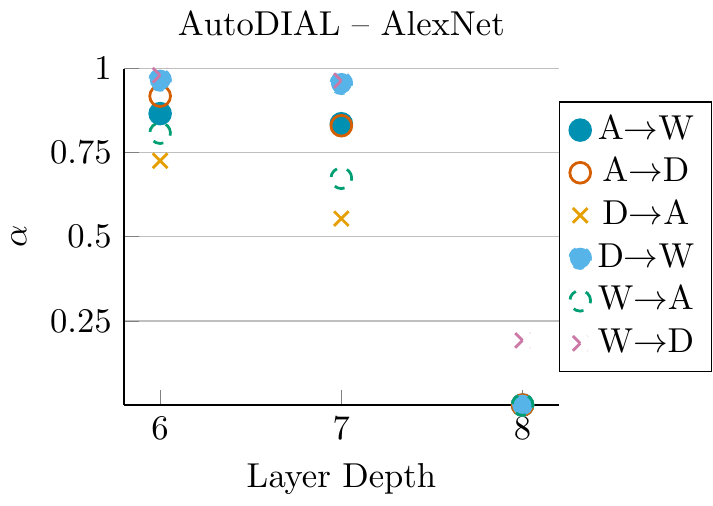}\hfill
  \includegraphics[height=4.3cm]{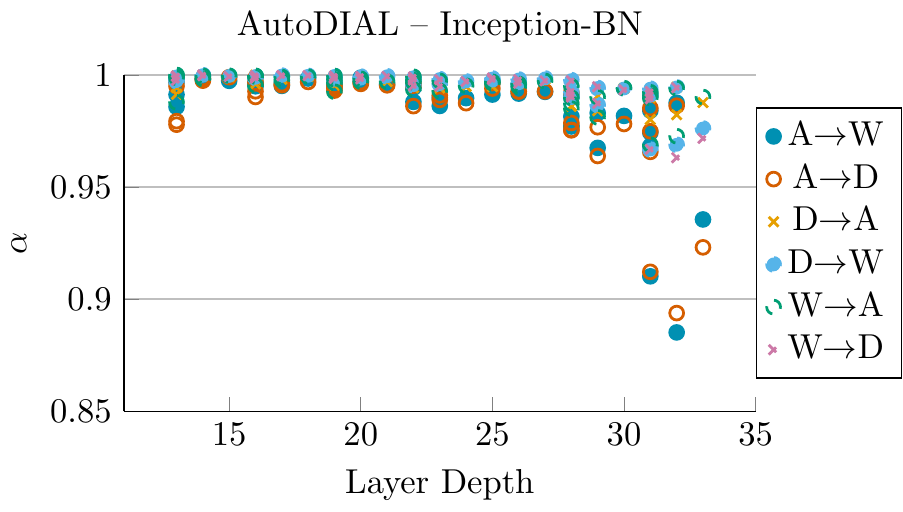}\hfill
  \includegraphics[height=4.3cm]{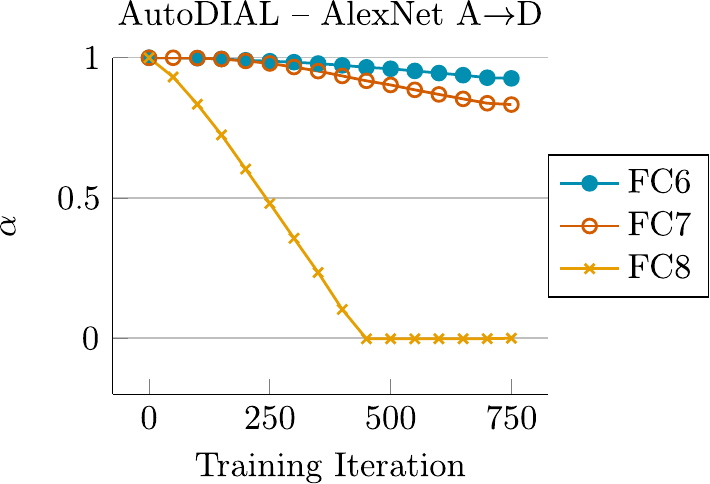}
  \caption{$\alpha$ parameters learned on the Office31 dataset, plotted as a function of layer depth (left and center) and training iteration (right).}
  \label{fig:dial_alpha}
\end{figure*}

Here we conduct an in-depth analysis of the full proposed approach, evaluating the impact of our three main contributions: i) aligning features by matching source and target distributions to a reference one; ii) learning the adaptation coefficients $\alpha$; iii) applying an entropy-based regularization term.

As a first set of experiments, we perform an ablation study on the Office31 dataset and report the results in Fig.~\ref{fig:dial_ablation}.

Specifically, we compare the performance of four variations of the \Alex network: trained on source data only (Source); with the addition of the entropy loss (Entropy); with \DALs and $\alpha$ fixed to 1 (\DIAL-fixed); with \DALs and learned $\alpha$ (\DIAL).
Here the advantage of learning $\alpha$ is evident, as \DIAL outperforms \DIAL-fixed in all but one of the experimental settings.
Interestingly, the addition of the entropy term by itself seems to have mixed effects on the final accuracy: in D$\rightarrow$A and W$\rightarrow$A in particular, the performance drastically decreases in Entropy compared to Source.
This is not surprising as these two settings correspond to cases where the number of labeled source samples is very limited and the domain shift is more severe. However, using \DALs in conjunction with the entropy loss always leads to a sizable performance increase.
These results confirms the validity of our contribution: the entropy regularization term is especially beneficial when source and target data representations are aligned.

In Fig.~\ref{fig:dial_alpha} we plot the values of $\alpha$ learned by the \DALs in \DIALAlex{} and \DIALInception{} on the Office31 dataset.
In both networks we observe that lower layers tend to learn values closer to $1$, \ie require an higher degree of adaptation compared to the layers closer to the classifier.
This behavior, however, seems to be more pronounced in \DIALAlex{} compared to \DIALInception{}.
Our results agree with recent findings in the literature \cite{aljundi2016lightweight}, according to which lower layers in a network are subject to domain shift even more than the very last layers.
During training, the $\alpha$ are able to converge to their final values in a few iterations (Fig.~\ref{fig:dial_alpha}, right).

\subsubsection{Comparison with State of the Art methods}
\label{sec:dial_results-comparison}

\begin{table}
  \begin{adjustbox}{width=1.2\textwidth,center=\textwidth}
  \begin{tabular}{lr*{6}{c}c}
    \toprule
    \multirow{2}{*}{Method} & \small{Source} & Amazon & Amazon & DSLR   & DSLR   & Webcam & Webcam & \multirow{2}{*}{Average}\\
                            & \small{Target} & DSLR   & Webcam & Amazon & Webcam & Amazon & DSLR   & \\
    \midrule
    \multicolumn{2}{l}{\Alex{} -- source~\cite{krizhevsky2012imagenet}}
      & $63.8$ & $61.6$ & $51.1$ & $95.4$ & $49.8$ & $99.0$ & $70.1$ \\
    \multicolumn{2}{l}{DDC~\cite{tzeng2014deep}}
      & $64.4$ & $61.8$ & $52.1$ & $95.0$ & $52.2$ & $98.5$ & $70.6$ \\
    \multicolumn{2}{l}{DAN~\cite{long2015learning}}
      & $67.0$ & $68.5$ & $54.0$ & $96.0$ & $53.1$ & $99.0$ & $72.9$ \\
    \multicolumn{2}{l}{ReverseGrad~\cite{ganin2014unsupervised}}
      & $67.1$ & $72.6$ & $54.5$ & $96.4$ & $52.7$ & $99.2$ & $72.7$ \\
   %   & 67.1 $ $$ & $72.6$ & 54.5 & $96.4$ & 52.7 & $99.2$ & 72.7 \\
     %     \multicolumn{2}{l}{Fua~\cite{}}
     % & -- & $76.0$ & -- & $96.7$ & -- & $99.6$ & -- \\
                \multicolumn{2}{l}{DRCN~\cite{ghifary2016deep}}
      & $66.8$ & $68.7$ & $56.0$ & $96.4$ & $54.9$ & $99.0$ & $73.6$ \\
    \multicolumn{2}{l}{RTN~\cite{long2016unsupervised}}
      & $71.0$ & $73.3$ & $50.5$ & $\mathbf{96.8}$ & $51.0$ & $\mathbf{99.6}$ & $73.7$\\
    \multicolumn{2}{l}{JAN~\cite{long2016deep}}
      & $71.8$ & $74.9$ & $\mathbf{58.3}$ & $96.6$ & $55.0$ & ${99.5}$ & $76.0$ \\
    \midrule
    %\multicolumn{2}{l}{\oldDIALAlex $sparse$}
      %& $72.4$ & $\mathbf{76.5}$ & $\mathbf{58.6}$ & $\mathbf{97.0}$ & $55.9$ & ${99.4}$ & $76.5$ \\
      \multicolumn{2}{l}{\DIALAlex}
      & $\mathbf{73.6}$ & $\mathbf{75.5}$ & $58.1$ & $96.6$ & $\mathbf{59.4}$ & ${99.5}$ & $\mathbf{77.1}$ \\
    \bottomrule
  \end{tabular}
  \end{adjustbox}
  \caption{AlexNet-based approaches on Office31 / full sampling protocol.}
  \label{tab:dial_office31-alex}
\end{table}

\begin{table}
  \begin{adjustbox}{width=1.2\textwidth,center=\textwidth}
  \begin{tabular}{lr*{6}{c}c}
    \toprule
    \multirow{2}{*}{Method} & \small{Source} & Amazon & Amazon & DSLR   & DSLR   & Webcam & Webcam & \multirow{2}{*}{Average}\\
                            & \small{Target} & DSLR   & Webcam & Amazon & Webcam & Amazon & DSLR   & \\
    \midrule
    %61.33	58.33	46.5	93.83	47.67	98.7	67.73
    \multicolumn{2}{l}{\VGGf{} -- source~\cite{chatfield2014return}}
      & $61.3$ & $58.3$ & $46.5$ & $93.8$ & $47.7$ & $98.7$ & $67.7$ \\
    \multicolumn{2}{l}{GFK~\cite{venkateswara2017deep}}
      & $48.6$ & $52.1$ & $41.8$ & $89.2$ & $49.0$ & $93.2$ & $62.3$ \\
    \multicolumn{2}{l}{TCA~\cite{venkateswara2017deep}}
      & $51.0$    & $49.4$ & $48.1$ & $93.1$ & $48.8$ & $96.8$ & $64.5$ \\
    \multicolumn{2}{l}{CORAL~\cite{venkateswara2017deep}}
      & $54.4$ & $51.7$  & $48.3$ & $96.0$ & $47.3$ & $98.6$ & $66.0$ \\
    \multicolumn{2}{l}{JDA~\cite{venkateswara2017deep}}
      & $59.2$ & $58.6$ & $51.4$ & $96.9$ & $52.3$ & $97.8$ & $69.4$ \\
    \multicolumn{2}{l}{DAN~\cite{venkateswara2017deep}}
      & $67.0$ & $67.8$  & $50.4$ & $95.9$ & $52.3$ & $99.4$  & $72.1$ \\
    \multicolumn{2}{l}{DANN~\cite{venkateswara2017deep}}
      & $\mathbf{72.9}$ & $72.7$  & $56.3$ & $96.5$ & $53.2$  & $\mathbf{99.4}$  & $75.2$ \\
    \multicolumn{2}{l}{DAH-e~\cite{venkateswara2017deep}}
      & $66.3$ & $66.2$ & $56.0$ & $94.6$ & $53.9$ & $97.0$ & $72.3$ \\
    \multicolumn{2}{l}{DAH~\cite{venkateswara2017deep}}
      & $66.5$ & $68.3$  & $55.5$ & $96.1$  & $53.0$ & $98.8$  & $73.0$ \\
    \midrule
    \multicolumn{2}{l}{\DIALVGGf}
    %72.2  & 73.1  & 61.0  & 97.6  & 57.9  & 98.3  & 76.7
      & $72.2$ & $\mathbf{73.1}$ & $\mathbf{61.0}$ & $\mathbf{97.6}$ & $\mathbf{57.9}$ & ${98.3}$ & $\mathbf{76.7}$ \\
    \bottomrule
  \end{tabular}
  \end{adjustbox}
    \caption{VGGf-based approaches on Office31 / full sampling protocol.}
  \label{tab:dial_office31-vggf}
\end{table}

\begin{table}
  \begin{adjustbox}{width=1.2\textwidth,center=\textwidth}
  \begin{tabular}{lr*{12}{c}c}
    \toprule
    \multirow{2}{*}{Method} & \small{Source} & Ar & Ar & Ar & Cl & Cl & Cl & Pr & Pr & Pr & Rw & Rw & Rw  & \multirow{2}{*}{Average}\\
                            & \small{Target} & Cl & Pr & Rw & Ar & Pr & Rw & Ar & Cl & Rw & Ar & Cl & Pr& \\
    \midrule
    \multicolumn{2}{l}{\VGGf{} - source\cite{chatfield2014return}}
      & $30.1$  & $42.9$  & $54.4$  & $31.6$  & $45.8$  & $47.9$  & $29.8$  & $30.1$  & $55.0$  & $43.9$  & $35.6$  & $62.6$  & $42.5$ \\
    \multicolumn{2}{l}{GFK~\cite{venkateswara2017deep}}
      & $21.6$  & $31.7$  & $38.8$  & $21.6$  & $34.9$  & $34.2$  & $24.5$  & $25.7$  & $42.9$  & $32.9$  & $29.0$  & $50.9$  & $32.4$ \\
    \multicolumn{2}{l}{TCA~\cite{venkateswara2017deep}}
      & $19.9$  & $32.1$  & $35.7$  & $19.0$  & $31.4$  & $31.7$  & $21.9$  & $23.6$  & $42.1$  & $30.7$  & $27.2$  & $48.7$  & $30.3$ \\
    \multicolumn{2}{l}{CORAL~\cite{venkateswara2017deep}}
      & $27.1$  & $36.2$  & $44.3$  & $26.1$  & $40.0$  & $40.3$  & $27.8$  & $30.5$  & $50.6$  & $38.5$  & $36.4$  & $57.1$  & $37.9$ \\
    \multicolumn{2}{l}{JDA~\cite{venkateswara2017deep}}
      & $25.3$  & $36.0$  & $42.9$  & $24.5$  & $40.2$  & $40.9$  & $26.0$  & $32.7$  & $49.3$  & $35.1$  & $35.4$  & $55.4$  & $37.0$ \\
    \multicolumn{2}{l}{DAN~\cite{venkateswara2017deep}}
      & $30.7$  & $42.2$  & $54.1$  & $32.8$  & $47.6$  & $49.8$  & $29.1$  & $34.1$  & $56.7$  & $43.6$  & $38.3$  & $62.7$  & $43.5$ \\
    \multicolumn{2}{l}{DANN~\cite{venkateswara2017deep}}
      & $33.3$  & $43.0$  & $54.4$  & $32.3$  & $49.1$  & $49.8$  & $30.5$  & $38.1$  & $56.8$  & $44.7$  & $42.7$  & $64.7$  & $44.9$ \\
    \multicolumn{2}{l}{DAH-e~\cite{venkateswara2017deep}}
      & $29.2$  & $35.7$  & $48.3$  & $33.8$  & $48.2$  & $47.5$  & $29.9$  & $38.8$  & $55.6$  & $41.2$  & $45.0$  & $59.1$  & $42.7$ \\
    \multicolumn{2}{l}{DAH~\cite{venkateswara2017deep}}
      & $31.6$  & $40.8$  & $51.7$  & $34.7$  & $51.9$  & $52.8$  & $29.9$  & $\mathbf{39.6}$  & $60.7$  & $45.0$  & $45.1$  & $62.5$  & $45.5$ \\
    \midrule
    \multicolumn{2}{l}{\DIALVGGf}
    %72.2  & 73.1  & 61.0  & 97.6  & 57.9  & 98.3  & 76.7
      & $\mathbf{35.7}$  & $\mathbf{53.7}$  & $\mathbf{61.6}$  & $\mathbf{38.9}$  & $\mathbf{58.7}$  & $\mathbf{61.3}$  & $\mathbf{37.8}$  & $39.1$  & $\mathbf{65.8}$  & $\mathbf{48.5}$  & $\mathbf{46.2}$  & $\mathbf{70.0}$  & $\mathbf{51.4}$ \\
    \bottomrule
  \end{tabular}
  \end{adjustbox}
  \caption{VGGf-based approaches on Office-Home\cite{venkateswara2017deep}. \{Art (Ar), Clipart (Cl),
Product (Pr), Real-World (Rw)\}}
  \label{tab:dial_office_home-vggf}
\end{table}

\begin{table}
\begin{adjustbox}{width=1.2\textwidth,center=\textwidth}
  \begin{tabular}{lr*{6}{c}c}
    \toprule
    \multirow{2}{*}{Method} & \small{Source} & Amazon & Amazon & DSLR   & DSLR   & Webcam & Webcam & \multirow{2}{*}{Average}\\
                            & \small{Target} & DSLR   & Webcam & Amazon & Webcam & Amazon & DSLR   & \\
    \midrule
    \multicolumn{2}{l}{\Inception{} -- source~\cite{ioffe2015batch}}
      & $70.5$ & $70.3$ & $60.1$ & $94.3$ & $57.9$ & $\mathbf{100.0}$ & $75.5$ \\
    \multicolumn{2}{l}{AdaBN~\cite{li2016revisiting}}
      & $73.1$ & $74.2$ & $59.8$ & $95.7$ & $57.4$ & $99.8$ & $76.7$ \\
    \multicolumn{2}{l}{AdaBN + CORAL~\cite{li2016revisiting}}
      & $72.7$ & $75.4$ & $59.0$ & $96.2$ & $60.5$ & $99.6$ & $77.2$ \\
    \multicolumn{2}{l}{DDC~\cite{tzeng2014deep}}
      & $73.2$ & $72.5$ & $61.6$ & $95.5$ & $61.6$ & $98.1$ & $77.1$ \\
    \multicolumn{2}{l}{DAN~\cite{long2015learning}}
      & $74.4$ & $76.0$ & $61.5$ & $95.9$ & $60.3$ & $98.6$ & $77.8$ \\
    \multicolumn{2}{l}{JAN~\cite{long2016deep}}
      & $77.5$ & $78.1$ & $\mathbf{68.4}$ & $96.4$ & $\mathbf{65.0}$ & $99.3$ & $80.8$ \\
    \midrule
    %\multicolumn{2}{l}{\oldDIALInception $BN$}
    %  & $\mathbf{87.3}$ & $82.9$ & $63.1$ & $\mathbf{98.2}$ & $62.6$ & $99.9$ & $\mathbf{82.4}$ \\
    \multicolumn{2}{l}{\DIALInception}
      & $\mathbf{82.3}$ & $\mathbf{84.2}$ & $64.6$ & $\mathbf{97.9}$ & $64.2$ & $99.9$ & $\mathbf{82.2}$ \\
    \bottomrule
  \end{tabular}
  \end{adjustbox}
  \caption{Inception-based approaches on Office31 / full sampling protocol.}
  \label{tab:dial_office31-inception}
\end{table}

\begin{table}[!th]
\begin{adjustbox}{width=1.2\textwidth,center=\textwidth}
  \begin{tabular}{lr*{6}{c}c}
  \toprule
  \multirow{2}{*}{Method} & \small{Source} & Amazon & Webcam & DSLR & Caltech & Caltech   & Caltech  & \multirow{2}{*}{Average} \\
                          & \small{Target} & Caltech & Caltech   & Caltech & Amazon & Webcam & DSLR & \\
  \midrule
  \multicolumn{2}{l}{AlexNet -- source~\cite{krizhevsky2012imagenet}} 
    & $83.8$ & $76.1$ & $80.8$ & $91.1$ & $83.1$ & $89.0$ & $84.0$\\
  \multicolumn{2}{l}{DDC~\cite{tzeng2014deep}} 
    & $85.0$ & $78.0$ & $81.1$ & $91.9$ & $85.4$ & $88.8$ & $85.0$\\
  \multicolumn{2}{l}{DAN~\cite{long2015learning}} 
    & $85.1$ & $84.3$ & $82.4$ & $92.0$ & $90.6$ & $90.5$ & $87.5$\\
  \multicolumn{2}{l}{RTN~\cite{long2016unsupervised}} 
    & $\mathbf{88.1}$ & $86.6$ & $84.6$ & $93.7$ & $\mathbf{96.9}$ & $\mathbf{94.2}$ & $\mathbf{90.6}$\\
  % \multicolumn{2}{l}{RTN (no RES)~\cite{long2016unsupervised}} 
  %   & $87.8$ & $84.8$ & $83.4$ & $93.2$ & $96.6$ & $93.9$ & $89.9$\\
  \midrule
  \multicolumn{2}{l}{\DIALAlex}
    & $87.4$ & $\mathbf{86.8}$ & $\mathbf{86.9}$ & $\mathbf{94.3}$ & $96.3$ & $90.1$ & $90.3$\\
   % \midrule
%  \multicolumn{2}{l}{\DIALAlex{} CP}     & $\mathbf{88.1}$ & $\mathbf{87.2}$ & $\mathbf{88.1}$ & $\mathbf{94.5}$ & $\mathbf{97.1}$ & $90.5$ & $\mathbf{90.9}$\\
  %\midrule
  %\multicolumn{2}{l}{\DIALInception}
  %  & \boldmath{$93.2$} & \boldmath{$90.5$} & \boldmath{$90.0$} & \boldmath{$95.1$} & \boldmath{$98.5$} & \boldmath{$96.9$} & \boldmath{$94.0$}\\
  \bottomrule
  \end{tabular}
  \end{adjustbox}
  \caption{Office-Caltech results using the full protocol.}
  \label{tab:dial_office-caltech}
\end{table}

\begin{table}
  \centering
  \begin{tabular}{lr*{2}{c}}
    \toprule
    \multirow{2}{*}{Method} & \small{Source} & Caltech  & Imagenet \\
                            & \small{Target} & Imagenet & Caltech  \\
    \midrule
    \multicolumn{2}{l}{SDT~\cite{tzeng2015simultaneous}}
      & -- & $73.6$ \\
    \multicolumn{2}{l}{Tommasi \etal~\cite{tommasi2016learning}}
      & -- & $75.4$ \\
    \midrule
    \multicolumn{2}{l}{\Inception{} -- source~\cite{ioffe2015batch}}
      & $82.1$ & $88.4$ \\
    \multicolumn{2}{l}{AdaBN~\cite{li2016revisiting}}
      & $82.2$  & $87.3$ \\
    \midrule
    \multicolumn{2}{l}{\DIALInception}
      & $\textbf{85.2}$ & $\textbf{90.5}$ \\
    \bottomrule
  \end{tabular}
    \caption{Cross-Dataset Testbed results using the protocol in~\cite{tommasi2014testbed}.}
  \label{tab:dial_ci40-default}
\end{table}

\begin{table}
  \centering
  \begin{tabular}{lr*{2}{c}}
    \toprule
    \multirow{2}{*}{Method} & \small{Source} & Caltech  & Imagenet \\
                            & \small{Target} & Imagenet & Caltech  \\
    \midrule
    \multicolumn{2}{l}{SA~\cite{fernando2013unsupervised}}
      & $43.7$ & $52.0$ \\
    \multicolumn{2}{l}{GFK~\cite{gong2012geodesic}}
      & $52.0$ & $58.5$ \\
    \multicolumn{2}{l}{TCA~\cite{pan2011domain}}
      & $48.6$ & $54.0$ \\
    \multicolumn{2}{l}{CORAL~\cite{sun2016return}}
      & $66.2$ & $74.7$ \\
    \midrule
    \multicolumn{2}{l}{\Inception{} -- source~\cite{ioffe2015batch}}
      & $82.1$ & $88.4$ \\
    \multicolumn{2}{l}{AdaBN~\cite{li2016revisiting}}
      & $81.9$  & $86.5$ \\
    \midrule
    \multicolumn{2}{l}{\DIALInception}
      & $\mathbf{84.2}$ & $\mathbf{89.8}$ \\
    \bottomrule
  \end{tabular}
  \caption{Cross-Dataset Testbed results using the protocol in~\cite{sun2016return}.}
  \label{tab:dial_ci40-saenko}
\end{table}

In this section we compare our approach with state-of-the art deep domain adaptation methods. We first consider the Office-31 dataset.
The results of our evaluation, obtained embedding the proposed DA-layers in the AlexNet, \VGGf{} and the Inception-BN networks as explained in subsection~\ref{ss:dial_setup}, are summarized in Tables~\ref{tab:dial_office31-alex}, ~\ref{tab:dial_office31-vggf} and~\ref{tab:dial_office31-inception}, respectively.
As baselines, we consider: Deep Adaptation Networks (DAN) \cite{long2015learning}, Deep Domain Confusion (DDC) \cite{tzeng2014deep}, the ReverseGrad network \cite{ganin2014unsupervised}, Residual Transfer Network (RTN) ~\cite{long2016unsupervised}, Joint Adaptation Network (JAN) \cite{long2016deep}, Deep Reconstruction-Classification Network
(DRCN) \cite{ghifary2016deep} and AdaBN~\cite{li2016revisiting} with and without CORAL feature alignment \cite{sun2016return}. The results associated to the baseline methods are derived from the original papers.
When using the \VGGf{} network we instead compare with Deep Hashing Networks\cite{venkateswara2017deep} and the baselines reported in their work.
%We compare these baselines to the AlexNet and Inception-BN networks modified with our approach as explained in subsection~\ref{sec:dial_setup}.
%In the tables our approach is denoted as \DIALAlex and \DIALInception.
As a reference, we further report the results obtained considering standard AlexNet, \VGGf{} and Inception-BN networks trained only on source data. %, and with AlexNet trained using the proposed approach but without domain-alignment layers %(\DIALAlexNet-No DA).

%Unsurprisingly, the classification accuracy of the two methods based on hand-crafted features is greatly inferior to all other approaches.
Among the deep methods based on the AlexNet architecture, \DIALAlex \ shows the best average performance, clearly demonstrating the benefit of the proposed adaptation strategy.
Similar results are found in the experiments with \VGGf{} and Inception-BN networks, where our approach also outperforms all baselines.

It is interesting to compare \DIAL with the AdaBN method \cite{li2016revisiting}, as this approach also develops from a similar intuition than ours. Our results clearly demonstrate the added value of our contributions: the introduction of the alignment parameters $\alpha$, together with the adoption of the entropy regularization term, produce a significant boost in performance.
It is interesting to note that the relative increase in accuracy from the source-only Inception-BN to \DIALInception is higher than that from the source only AlexNet to \DIALAlex.
The considerable success of our method in conjunction with Inception-BN can be attributed to the fact that, differently from AlexNet, this network is pre-trained with batch normalization, and thus initialized with weights that are already calibrated for normalized features.
% In appendix ~\ref{sec:dial_office31}, we also provide \DIALAlex results on Office 31, using the classical\cite{saenko2010adapting} sampling protocol and show that our method can successfully deal with less samples.

In our second set of experiments we analyze the performance of several approaches on the Office-Caltech dataset.
The results are reported in Table~\ref{tab:dial_office-caltech}.
We restrict our attention to methods based on deep architectures and, for a fair comparison, we consider all AlexNet-based approaches.
Here we report results obtained with DDC~\cite{tzeng2014deep}, DAN~\cite{long2015learning}, and the recent Residual Transfer Network (RTN) in~\cite{long2016unsupervised}.
As it is clear from the table, our method and RTN have roughly the same performance (90.6$\%$ vs 90.4$\%$ on average), while they significantly outperform the other baselines.

As a third set of experiments we evaluate our method on the $12$ settings of Office-Home. As shown in table ~\ref{tab:dial_office_home-vggf}, \DIAL comfortably outperforms the previous state of the art on this setting.

For our multi-source experiments, we benchmark our approach on Office, testing on each of the three domains, while using the remaining two as sources (results in table ~\ref{tab:dial_office31-multi}).
Both our variants, $\alpha$ free or fixed, outperform the existing state of the art in this setting.

Finally, we perform some large scale experiments on the Caltech-ImageNet subset of the Cross-Dataset Testbed of \cite{tommasi2014testbed}. 
As explained above, to facilitate comparison with previous works which have also considered this dataset we perform experiments in two different settings. As baselines we consider Geodesic Flow Kernel (GFK) \cite{gong2012geodesic}, Subspace Alignment (SA) \cite{fernando2013unsupervised}), CORAL \cite{sun2016return}, Transfer Component Analysis (TCA)~\cite{pan2011domain}, Simultaneous Deep Transfer (SDT)~\cite{tzeng2015simultaneous}, and the recent method in \cite{tommasi2016learning}.
Table ~\ref{tab:dial_ci40-default} and Table ~\ref{tab:dial_ci40-saenko} show our results.

The proposed approach significantly outperforms previous methods and sets the new state of the art on this dataset.
The higher performance of our method is not only due to the use of Inception-BN but also due to the effectiveness of our contributions. 
Indeed, the proposed alignment strategy, combined with the adoption of the entropy regularization term, %unique combination of the proposed regularization term (see Equation~\eqref{eq:prior}) with our \DALs 
makes our approach more effective than previous adaptation techniques based on Inception-BN, \ie AdaBN \cite{li2016revisiting}.

\begin{table}[tbp]
\centering
\begin{tabular}{lcccc}
\toprule
Source:           & A-W  & A-D  & D-W  & Mean \\
Target:           & Dslr    & Webcam    & Amazon    &      \\
\midrule
Source only                   & $98.2$ & $92.7$ & $51.6$ & $80.8$ \\
sFRAME~\cite{xie2015learning} & $54.5$ & $52.2$ & $32.1$ & $46.3$ \\
SGF~\cite{Gopalan2011}  & $39.0$ & $52.0$ & $28.0$ & $39.7$ \\
DCTN~\cite{cocktail_CVPR18}        & $\mathbf{99.6}$ & $96.9$ & $54.9$ & $83.8$ \\
\midrule
AdaBN (Inception-BN)~\cite{li2016revisiting}
                              & 94.2 & $\mathbf{97.2}$ & $59.3$ & $83.6$ \\
\midrule
%\oldDIALAlex\cite{massimilianoPAMI} & $94.8$ & $95.8$ & $\mathbf{62.9}$ & $84.5$ \\
%Our best DIAL     & $97.2$ & $95.8$ & $62.7$ & $85.2$ \\
\DIALAlex & $97.2$ & $95.3$ & $62.7$ & $\mathbf{85.1}$ \\
\bottomrule
\end{tabular}
\caption{\DIAL results on the multi-source Office31 setting}
\label{tab:dial_office31-multi}
\end{table}

%%%%%%%%%%%%%%%%%%%%%% Conclusions %%%%%%%%%%%%%%%%%%%%%%%%%%%%

\subsection{Conclusion}
\label{subsec:dial_conclusions}
We presented \DIAL, a novel framework for unsupervised, deep domain adaptation.
The core of our contribution is the introduction of novel Domain Alignment layers, which reduce the domain shift by matching source and target distributions to a reference one. Our DA-layers are endowed with a set of alignment parameters, also learned by the network, which allow the CNN not only to align the source and target feature representations but also to automatically decide at each layer the required degree of adaptation. 
Our framework exploits target data both by computing statistics in the DA-layers and by introducing an entropy loss which promotes classification models with high confidence on unlabeled samples.
%We evaluated the proposed approach devising a simple implementation of our \DALs based on batch normalization.
The results of our experiments demonstrate that our approach outperforms state of the art domain adaptation methods.

While this section focuses on the challenging problem of unsupervised domain-adaptation, our approach can be also exploited in a semi-supervised setting. Future works will be devoted to analyze the effectiveness of \DIAL in this scenario. 

\section{Image Alignment: SBADA-GAN}
\label{sec:da_image_align}
\textit{The effectiveness of GANs in producing images according to a specific 
visual domain has shown potential in  unsupervised domain adaptation. 
Source labeled images have been modified to mimic target samples for 
training classifiers in the target domain, 
and inverse mappings from the target to the source domain have also been evaluated. }

\textit{In this section we aim at getting the best of both worlds
by introducing a symmetric mapping among domains. We jointly optimize bi-directional 
image transformations combining them with target self-labeling. 
We define a new class consistency loss that aligns the generators in the two directions, 
imposing to preserve the class identity of an image passing through both domain mappings.
A detailed analysis of the reconstructed images, a thorough ablation study
and extensive experiments on six different settings confirm the power
of our approach. }
\newline

The ability to generalize across domains is challenging when there is ample labeled data on which to train a deep network (source domain), but no annotated data for the target domain. To attack this issue, a wide array of methods have been proposed, most of them aiming at reducing the shift between the source and target distributions (see Sec.~\ref{sec:domain_adap_related} for a review of previous work). An alternative is mapping the source data into the target domain,
either by modifying the image representation ~\cite{Ganin:DANN:JMLR16} or by directly generating a new version of the source images ~\cite{Bousmalis:Google:CVPR17}. Several authors proposed approaches that follow both these strategies by building over Generative Adversarial Networks (GANs) ~\cite{Goodfellow:GAN:NIPS2014}. 
A similar but inverse method maps the target data into the source domain, where there is already an abundance of labeled images ~\cite{Hoffman:Adda:CVPR17}.

We argue that these two mapping directions should not be alternative, but complementary. 
Indeed, the main ingredient for adaptation is the ability of transferring successfully the style of one domain to the images of the other. This, given a fixed generative architecture, will depend on the application: there may be cases where mapping from the source to the target is easier, and cases where it is true otherwise. By pursuing both directions in a unified architecture, we can obtain a system more robust and more general than previous adaptation algorithms. 

\begin{figure}
\includegraphics[width=1.0\textwidth]{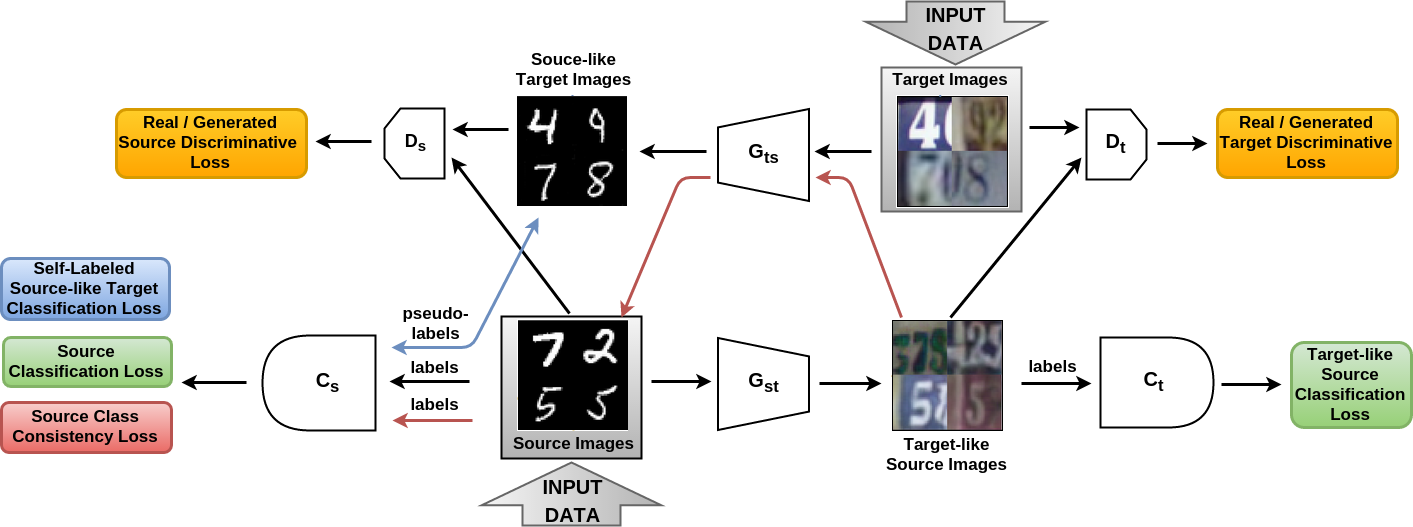}
{\caption{SBADA-GAN, training: the data flow starts from the Input Data arrows. The bottom and top row show respectively the source-to-target and target-to-source symmetric directions. The generative models $G_{st}$ and $G_{ts}$ transform the source images to the target domain and  vice versa.
 $D_{s}$ and $D_{t}$ discriminate real from generated images of source and target.  Finally the classifiers  $C_{s}$ and $C_{t}$ are trained to recognize respectively the original source images and their target-like transformed versions.
The bi-directional blue arrow indicates that the source-like target images are automatically annotated and the assigned pseudo-labels are re-used by the classifier $C_s$. The red arrows describe the class consistency condition by which source images transformed to the target domain through $G_{st}$ and back to the source domain through $G_{ts}$ should maintain their ground truth label.}
\label{fig:sbadagan}}
\end{figure}

With this idea in mind, we designed SBADA-GAN: Symmetric Bi-directional ADAptive Generative Adversarial Network.
Its features are (see Figure ~\ref{fig:sbadagan}): 

\begin{itemize}

\item it exploits two generative adversarial losses that encourage the network to produce 
target-like images from the source samples and source-like images from the target samples.
Moreover, it jointly minimizes two classification losses, one on the original 
source images and the other on the transformed target-like source images;

\item it uses the source classifier to annotate the source-like transformed target images. 
Such pseudo-labels 
help regularizing the same classifier while improving 
the target-to-source generator model by backpropagation;

\item it introduces a new semantic constraint on the source images: the \textit{class consistency loss}. It 
imposes that by mapping source images towards the target domain and then again towards the 
source domain they should get back to their ground truth class.
This last condition is less restrictive than a standard reconstruction loss 
~\cite{CycleGAN2017,DBLP:conf/icml/KimCKLK17}, as it deals 
only with the image annotation and not with the image appearance. Still, our experiments 
show that it is highly effective in aligning the domain mappings in the two directions;  

\item at test time the two trained classifiers are used respectively
on the original target images and on their source-like transformed version. The two 
predictions are integrated to produce the final annotation.
\end{itemize}

Our architecture yields realistic image reconstructions while  competing against previous 
state-of-the-art classifiers and exceeding them on four out of six different 
unsupervised adaptation settings. 
An ablation study showcasing the importance of each component in the architecture, and investigating 
the robustness with respect to its hyperparameters, sheds light on the inner workings of the approach, while providing  further evidence of its value.

%%%%%%%%%%%%%%%%%%% METHOD

\paragraph{Model}
We focus on unsupervised cross domain classification. Let us start from a dataset $\bm{X}_s=\{\bm{x}^i_s,y^i_s\}_{i=0}^{N_s}$ drawn from a labeled source domain $\mathcal{S}$, and a dataset $\bm{X}_t=\{\bm{x}^j_t\}_{j=0}^{N_t}$ from a different unlabeled target domain $\mathcal{T}$, sharing the same set of categories.  The task is to maximize the classification accuracy on $\bm{X}_t$ while training on $\bm{X}_s$.
To reduce the domain gap, we propose to adapt the source images such that they appear as sampled from 
the target domain by training a generator model $G_{st}$ that maps any source samples 
$\bm{x}^i_s$ to its target-like version $\bm{x}^i_{st}=G_{st}(\bm{x}^i_s)$ defining the set 
$\bm{X}_{st}=\{\bm{x}^i_{st},y^i_s\}_{i=0}^{N_s}$ (see Figure ~\ref{fig:sbadagan}, bottom row).
The model is also augmented with a discriminator $D_{t}$ and a classifier $C_{t}$. 
The former takes as input the target images $\bm{X}_t$ and target-like source transformed images $\bm{X}_{st}$, 
learning to recognize them as two different sets. The latter takes as input each of 
the transformed images $\bm{x}^i_{st}$ and learns to assign its task-specific label $y^i_s$. 
During the training procedure for this model, information about the domain recognition likelihood 
produced by $D_{t}$ is used adversarially to guide and optimize the performance of the generator $G_{st}$. 
Similarly, the generator also benefits from backpropagation in the classifier training procedure.

Besides the source-to-target transformation, we also consider the inverse target-to-source
direction by using a symmetric architecture (see Figure ~\ref{fig:sbadagan}, top row).
Here any target image $\bm{x}^j_t$ is given as input to a generator model $G_{ts}$ transforming it to its 
source-like version $\bm{x}^j_{ts}=G_{ts}(\bm{x}^j_t)$, defining the set $\bm{X}_{ts}=\{\bm{x}^j_{ts}\}_{j=0}^{N_t}$. 
As before, the model is augmented with a discriminator $D_{s}$ which takes as input both $\bm{X}_{ts}$ and 
$\bm{X}_s$ and learns to recognize them as two different sets, adversarially helping the generator. 

Since the target images are unlabeled, no classifier can be trained in the target-to-source
direction as a further support for the generator model. We overcome this issue by \emph{self-labeling}
(see Figure ~\ref{fig:sbadagan}, blue arrow).
The original source data $\bm{X}_s$ is used to train a classifier $C_{s}$.
Once it has reached convergence, we apply the learned model to annotate each of the source-like
transformed target images $\bm{x}^j_{ts}$. These samples, with the assigned pseudo-labels 
$y^j_{t_{self}}=\argmax_y(C_{s}(G_{ts}(\bm{x}^j_t))$, 
are then used transductively as input to $C_{s}$ while information about the performance of the model 
on them is backpropagated to guide and improve the generator $G_{ts}$.
Self-labeling has a long track record of success for domain adaptation: it proved to be effective 
both with shallow models ~\cite{BruzzonePAMI, HabrardPS13, Morvant15}, as well as with the most recent 
deep architectures ~\cite{TRUDA-NIPS16_savarese,Tzeng_ICCV2015,saito2017asymmetric}.
In our case the classification loss on pseudo-labeled samples is combined with our other losses, which 
helps making sure we move towards the optimal solution: in case of a moderate domain shift, the 
correct pseudo-labels help to regularize the learning process, while in case of large domain shift, 
the possible mislabeled samples do not hinder the performance (see Sec. ~\ref{subsec:sbadagan_ablation}
for a detailed discussion on the experimental results).

Finally, the symmetry in the source-to-target and target-to-source transformations is enhanced by 
aligning the two generator models such that, when used in sequence, they bring a sample back to its
starting point. Since our main focus is classification, we are interested in preserving the 
class identity of each sample rather than its overall appearance. Thus, instead of a standard image-based
reconstruction condition we introduce a \emph{class consistency} condition (see Figure ~\ref{fig:sbadagan}, red arrows).
Specifically, we impose that any source image $\bm{x}^i_s$ adapted to the target domain through $G_{st}(\bm{x}^i_s)$ 
and transformed back towards the source domain through $G_{ts}(G_{st}(\bm{x}^i_s))$ is correctly classified by 
$C_{s}$. This condition helps by imposing a further joint optimization of the two generators. 

\paragraph{Learning}
Here we formalize the description above. To begin with, we specify that the generators
take as input a noise vector $\bm{z}\in \mathcal{N}(0,1)$ besides the images, this allows
some extra degree of freedom to model external variations. We also better define
the discriminators as $D_{s}(\bm{x})$, $D_{t}(\bm{x})$ and the classifiers 
as $C_{s}(\bm{x})$, $C_{t}(\bm{x})$. Of course each of these models depends from its parameters 
but we do not explicitly indicate them to simplify the notation. For the same reason
we also drop the superscripts $i,j$. 

The source-to-target part of the network optimizes the following objective function:
\begin{equation}
\min_{{G_{st}}, {C_{t}}} \max_{{D_{t}}} ~~ \alpha \mathcal{L}_{D_t}(D_{t},G_{st}) + 
										   \beta \mathcal{L}_{C_t}(G_{st},C_{t})~,
\end{equation}
where the classification loss $\mathcal{L}_{C_t}$ is a standard \emph{softmax cross-entropy} 
\begin{align}
\mathcal{L}_{C_t}(G_{st},C_{t}) = \mathbb{E}_{\substack{\{\bm{x}_s, \bm{y}_s\}\sim \mathcal{S}\\\bm{z}_s\sim noise}}[{-\bm{y}_{s}}\cdot\log({\hat{\bm{y}}_s})]~,
\label{eq:sbadagan_class}
\end{align}
evaluated on the source samples transformed by the generator $G_{st}$, so that 
$\hat{\bm{y}}_s=C_{t}(G_{st}(\bm{x}_s, \bm{z}_s))$ and $\bm{y}_s$  is the one-hot encoding of the class label $y_s$. 
For the discriminator, instead of the less robust binary cross-entropy,
we followed ~\cite{mao2016multi} and chose a \emph{least square} loss:
\begin{align}
\mathcal{L}_{D_{t}}(D_{t},G_{st})  = & \mathbb{E}_{\bm{x}_t \sim T}[(D_{t}(\bm{x}_t) - 1)^2 ] + \nonumber \\
                                     & \mathbb{E}_{\substack{\bm{x}_s \sim S\\\bm{z}_s\sim noise}}[(D_{t}(G_{st}(\bm{x}_s, \bm{z}_s)))^2]~. 
                                     \label{eq:sbadagan_discr}
\end{align}

The objective function for the target-to-source part of the network is:
\begin{align}
\min_{{G_{ts}}, {C_{s}} } \max_{{D_{s}}} ~~ & \gamma \mathcal{L}_{D_s}(D_{s},G_{ts}) + \nonumber \\
                                            & \mu    \mathcal{L}_{C_s}(C_{s}) + \eta   \mathcal{L}_{self}(G_{ts},C_{s})~,
\end{align}
where the discriminative loss is analogous to eq. (~\ref{eq:sbadagan_discr}), while the
classification loss is analogous to eq. (~\ref{eq:sbadagan_class}) but evaluated on the original source samples
with $\hat{\bm{y}}_s = C_{s}(\bm{x}_s)$, 
thus it neither has any dependence on the generator that transforms the target samples $G_{ts}$, 
nor it provides feedback to it.
The \emph{self} loss is again a classification softmax cross-entropy: 
\begin{align}
\mathcal{L}_{self}(G_{ts},C_{s}) = \mathbb{E}_{\substack{\{\bm{x}_t, \bm{y}_{t_{self}}\} \sim \mathcal{T}\\\bm{z}_t\sim noise}}[-\bm{y}_{t_{self}}\cdot \log({\hat{\bm{y}}_{t_{self}}})]~.
\end{align}
where $\hat{\bm{y}}_{t_{self}} = C_{s}(G_{ts}(\bm{x}_t,\bm{z}_t))$ and $\bm{y_{t_{self}}}$ is the one-hot vector
encoding of the assigned label ${y_{t_{self}}}$. 
This loss back-propagates to the generator $G_{ts}$ which is encouraged to preserve the annotated category 
within the transformation.

Finally, we developed a novel \textit{class consistency} loss by minimizing the error of the classifier
$C_{s}$ when applied on the concatenated transformation of $G_{ts}$ and $G_{st}$ to produce
$\hat{\bm{y}}_{cons}= (C_{s}(G_{ts}(G_{st} (\bm{x}_s, \bm{z}_s), \bm{z}_t)))$:
\begin{align}
\mathcal{L}_{cons}(G_{ts},G_{st},C_{s})= %= ~~~~~~~~~~~~~~~~~~~~~~~~~~~~~~~~~~~~~~~~~~~~~~~~~~~~~~\nonumber\\
\mathbb{E}_{\substack{\{\bm{x}_s, \bm{y}_s\}\sim S\\\bm{z}_s,\bm{z}_t \sim noise}}[{-\bm{y}_s}\cdot\log({\hat{\bm{y}}_{cons}})]~.
\end{align}
This loss has the important role of aligning the generators in the two directions and strongly 
connecting the two main parts of our architecture. 

By collecting all the presented parts, we conclude with the complete SBADA-GAN loss: 
\begin{align}
\mathcal{L}_{SBADA-GAN}(G_{st},G_{ts},C_{s},C_{t},D_{s},D_{t})= ~~~~~~~~~~~~~~\nonumber\\
\alpha \mathcal{L}_{D_{t}}+ \beta \mathcal{L}_{C_{t}}+ \gamma \mathcal{L}_{D_{s}} + \mu \mathcal{L}_{C_{s}}+
\eta \mathcal{L}_{self} + \nu \mathcal{L}_{cons}~.                              
\end{align}                            
Here $(\alpha, \beta, \gamma,\mu,\eta,\nu)\geq0$ are weights that control the interaction of the loss terms.  
While the combination of six different losses might appear daunting, it is not unusual ~\cite{Bousmalis:DSN:NIPS16}.
Here, it stems from the symmetric bi-directional nature of the overall architecture. Indeed each 
directional branch has three losses as it is custom practice in the GAN-based domain adaptation literature ~\cite{Hoffman:Adda:CVPR17,Bousmalis:Google:CVPR17}. 
Moreover, the ablation study reported in Sec. ~\ref{subsec:sbadagan_ablation} indicates that the system is remarkably robust to 
changes in the hyperparameter values.

\begin{figure}[tb]
\includegraphics[width=\textwidth]{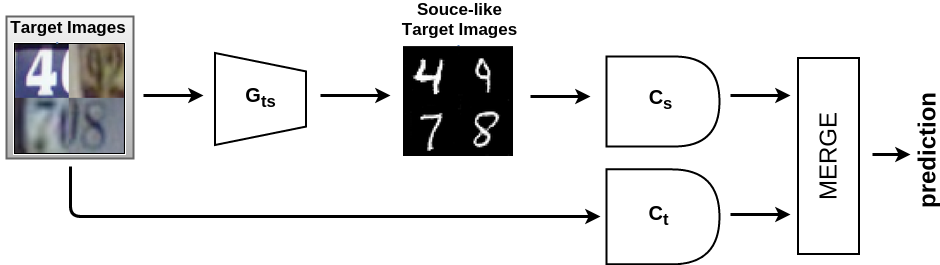}
{\caption{SBADA-GAN, test: the two pre-trained classifiers
are applied respectively on the target images and on the transformed source-like target images.
Their outputs are linearly combined for the final prediction.}
\label{fig:cycada_like_test}}
\end{figure}

\paragraph{Testing}
The classifier $C_{t}$ is trained on $\bm{X}_{st}$ generated 
images that mimic the target domain style, and is then tested on the original target samples $\bm{X}_t$.
The classifier $C_{s}$ is trained on $\bm{X}_s$ source data, and then tested on 
$\bm{X}_{ts}$ samples, that are the target images modified to mimic the source domain style.
These classifiers make mistakes of different type assigning also a different confidence rank to 
each of the possible labels. 
Overall the two classification models complement each other. 
We take advantage of this with a simple ensemble method $~\sigma C_s(G_{ts}(\bm{x}_t,\bm{z}_t)) + \tau C_t(\bm{x}_t)~$ which linearly 
combines their probability output, providing a further gain in performance.
A schematic illustration of the testing procedure is shown in Figure ~\ref{fig:cycada_like_test}.
We set the combination weights $\sigma,\tau$ through cross validation (see Sec. ~\ref{subsec:sbadagan_implem} for further details).

%%%%%%%%%%%%% EVALUATION %%%%%%%%%%%%%%%%%%%%

\begin{table*}[tbp]
\begin{adjustbox}{width=1.0\textwidth,center=\textwidth}
\begin{tabu} to 0.9\textwidth {@{}l@{}cccccc}
\hline
\rowfont{\scriptsize}
& \textbf{MNIST$\rightarrow$ USPS}  & \textbf{USPS$\rightarrow$MNIST} 
& \textbf{MNIST$\rightarrow$MNIST-M} & \textbf{SVHN$\rightarrow$MNIST} 
& \textbf{MNIST$\rightarrow$SVHN} & \textbf{Synth Signs$\rightarrow$GTSRB} \\
\hline
\rowfont{\small}Source Only                 			& 78.9   			& 57.1 $\pm$ 1.7    & 63.6      & 60.1 $\pm$ 1.1	& 26.0 $\pm$ 1.2 	& 79.0 \\
\hline
\rowfont{\small}CORAL ~\cite{Sun:CORAL:AAAI16} 				& 81.7      		&  -                 & 57.7      & 63.1        	& -			& 86.9 \\
\rowfont{\small}MMD ~\cite{Tzeng_ICCV2015}     				& 81.1      		&  -                 & 76.9      & 71.1         & -         & 91.1 \\
\rowfont{\small}DANN ~\cite{Ganin:DANN:JMLR16} 				& 85.1      		& 73.0 $\pm$ 2.0    & 77.4      & 73.9          & 35.7      & 88.7 \\
\rowfont{\small}DSN ~\cite{Bousmalis:DSN:NIPS16}				& 91.3      		&  -                & 83.2      & 82.7          & -         & 93.1 \\
\rowfont{\small}CoGAN ~\cite{Liu:coGAN:NIPS16}  				& 91.2      		& 89.1 $\pm$ 0.8    & 62.0      & not conv.     &  -     & -  \\
\rowfont{\small}ADDA ~\cite{Hoffman:Adda:CVPR17} 			& 89.4 $\pm$ 0.2    & 90.1 $\pm$ 0.8    & -     & 76.0 $\pm$ 1.8   & -     & - \\
\rowfont{\small}DRCN ~\cite{WenLi:ECCV2016}      			& 91.8 $\pm$ 0.1   & 73.7 $\pm$ 0.1   & -      & 82.0 $\pm$ 0.2  & 40.1 $\pm$ 0.1  & - \\
\rowfont{\small}PixelDA ~\cite{Bousmalis:Google:CVPR17}      & 95.9             & -             & 98.2      &  -              & -            & - \\
\rowfont{\small}DTN ~\cite{Taigman2016UnsupervisedCI} & -    & -                & -            &  84.4      & -            & - \\
\rowfont{\small}TRUDA ~\cite{TRUDA-NIPS16_savarese} & -      & -        & 86.7                 & 78.8       & 40.3    & - \\
\rowfont{\small}ATT ~\cite{saito2017asymmetric} & -      & -        & 94.2                 & 86.2       & 52.8    & 96.2 \\
\rowfont{\small}UNIT ~\cite{liu2017unsupervised} & 95.9      & 93.5        & -                 & 90.5       & -    & - \\
\rowfont{\small}DA$_{ass}$ fix. par. ~\cite{haeusser17}     & -         & -             & 89.5           & 95.7             &  -           & 82.8  \\
\rowfont{\small}DA$_{ass}$ ~\cite{haeusser17}               & -         & -             & 89.5           & \textbf{97.6}             &  -          & \textbf{97.7} \\
\hline
\rowfont{\small}\text{Target Only}            			& 96.5          & 99.2 $\pm$ 0.1    & 96.4     & 99.5     &  96.7     & 98.2        \\
\hline
\rowfont{\small}SBADA-GAN $C_t$          & 96.7         & 94.4         & 99.1           & 72.2             & 59.2 &   95.9      \\
\rowfont{\small}SBADA-GAN $C_s$         & 97.1         & 87.5         & 98.4           & 74.2              & 50.9  &  95.7     \\
\rowfont{\small}SBADA-GAN              & \textbf{97.6} & \textbf{95.0}         & \textbf{99.4}    & 76.1    & \textbf{61.1} & 96.7 \\
\hline
\hline
\rowfont{\small}GenToAdapt ~\cite{sankaranarayanan2017generate} & 92.5 $\pm$ 0.7    & 90.8 $\pm$ 1.3  & -       & 84.7 $\pm$ 0.9   & 36.4 $\pm$ 1.2   &- \\
\rowfont{\small}CyCADA ~\cite{hoffman2017cycada} & 94.8 $\pm$ 0.2            & 95.7  $\pm$ 0.2  & -       & 88.3 $\pm$ 0.2  &  -   &- \\
\rowfont{\small}Self-Ensembling ~\cite{visdawinners} & 98.3 $\pm$ 0.1  & 99.5 $\pm$ 0.4   & -       & 99.2 $\pm$ 0.3  &  42.0 $\pm$ 5.7   & 98.3 $\pm$ 0.3 \\
% %\text{Linear comb confidence} & 97.61              &               & 99.25           &                   &               \\
\hline
\end{tabu}
\end{adjustbox}
\caption{
Comparison against previous work.
SBADA-GAN $C_t,C_s$ reports respectively the accuracies produced by the classifier trained in the target domain space, 
and the results produced by training in the source domain space and testing on the target images mapped to this space. 
SBADA-GAN reports the results obtained by a weighted 
combination of the softmax outputs of these two classifiers. 
Note that all competitors convert SVHN to grayscale, while we deal with the more complex original RGB version. 
The last three rows report results from online available pre-print papers.}
\label{table:sbadagan_results}
\end{table*}

\subsection{Datasets and Adaptation Scenarios}
We evaluate SBADA-GAN on several unsupervised adaptation 
scenarios
, considering the following widely used datasets and settings (dataset details can be found in section ~\ref{sec:da_datasets}):
\begin{description}[wide=0\parindent]
\item[MNIST $\rightarrow$ MNIST-M:] We follow the evaluation protocol of ~\cite{Bousmalis:DSN:NIPS16,Bousmalis:Google:CVPR17,Ganin:DANN:JMLR16}.
\item[MNIST $\leftrightarrow$ USPS:]
We follow the evaluation protocol of ~\cite{Bousmalis:Google:CVPR17}.
\item[SVHN $\leftrightarrow$ MNIST:] Most previous works simplified the 
data by considering a grayscale version, instead we apply our method to the original RGB images. 
We resize the MNIST images to $32\times32$ pixels and use the protocol by 
~\cite{Bousmalis:DSN:NIPS16,WenLi:ECCV2016}.
We also test SBADA-GAN on a traffic sign scenario.\\
\item[Synth Signs $\rightarrow$ GTSRB:]
Both databases contain samples from $43$ 
%different 
classes, thus defining a larger classification task than
that on the $10$ digits.  We adopt the protocol proposed in ~\cite{haeusser17}.
\end{description}

\subsection{Implementation details}
\label{subsec:sbadagan_implem}
We composed SBADA-GAN starting from two symmetric GANs, each with an architecture\footnote{See all the model 
details in the supplementary material.} analogous to that used in ~\cite{Bousmalis:Google:CVPR17}.

The model is coded\footnote{Code available at https://github.com/engharat/SBADAGAN} in python and we ran all our 
experiments in the Keras framework ~\cite{chollet2017}. We use the ADAM ~\cite{kingma2014adam}
optimizer with learning rates for the discriminator and the generator both set to $10^{-4}$.
The batch size is set to $32$ and we train the model for $500$ epochs not noticing any overfitting, which suggests that further epochs might be beneficial.
The $\alpha$ and $\gamma$ loss weights (discriminator losses) are set to $1$, 
$\beta$ and $\mu$ (classifier losses) are set to $10$, 
to prevent that generator from indirectly 
switching labels (for instance, transform $7$'s into $1$'s). 
The class consistency loss weight $\nu$ is set to $1$.

All training procedures start with the self-labeling loss weight, $\eta$, set to zero, as 
this loss hinders convergence until the classifier is fully trained. After the model converges 
(losses stop oscillating, usually after $250$ epochs) $\eta$ is set to $1$ to further 
increase performance. 
Finally the parameters to combine the classifiers at test time are $\sigma \in [0,0.1,0.2, \ldots, 1]$ 
and $\tau=(1-\sigma)$ chosen on a validation set of 1000 random samples from the target in
each different setting.

\subsection{Quantitative Results}
Table ~\ref{table:sbadagan_results} shows results on our evaluation settings.
The top of the table reports
results
by thirteen competing 
baselines published over the last two years.
The Source-Only and Target-Only
rows contain reference results corresponding to the %na\"ive 
no-adaptation case and to the
target fully supervised case. 
For SBADA-GAN, besides the full method, we also report the accuracy obtained by the separate 
classifiers ($C_s$,$C_t$) before the linear combination.
%The last three rows show results that appeared recently in pre-prints available online.

SBADA-GAN improves over the state of the art in four out of six settings. 
In these cases the advantage with respect to its competitors is already visible in the 
separate $C_s$ and $C_t$ results and it increases when considering the full combination procedure. 
Moreover, the gain in performance of SBADA-GAN reaches up to $+8$ percentage points 
in the MNIST$\rightarrow$SVHN experiment. This setting was disregarded in many previous 
works: differently from its
%largely studied 
inverse SVHN$\rightarrow$MNIST, it requires a difficult adaptation from the 
grayscale handwritten digits domain to the widely variable and colorful street view house number domain.
Thanks to its bi-directionality, SBADA-GAN leverages on the inverse target to source mapping 
to produce highly accuracy results.

Conversely, in the SVHN$\rightarrow$MNIST case SBADA-GAN ranks eighth out of the thirteen baselines in terms of performance.  Our accuracy is on par with ADDA's ~\cite{Hoffman:Adda:CVPR17}: the two approaches share the same classifier architecture, although the number of fully-connected neurons of  SBADA-GAN is five time lower. Moreover, compared to DRCN~\cite{WenLi:ECCV2016}, the classifiers of SBADA-GAN are shallower with a reduced number of convolutional layers. 
Overall here SBADA-GAN suffers of some typical drawbacks of GAN-based domain adaptation methods: although the style of a domain can be easily transferred in the raw pixel space, the generative process does not have any explicit constraint on reducing the overall data distribution shift as instead done by the alternative feature-based domain adaptation approaches. Thus, methods like DA$_{ass}$~\cite{haeusser17}, DTN~\cite{Taigman2016UnsupervisedCI} and DSN~\cite{Bousmalis:DSN:NIPS16} deal better with the large domain gap of the SVHN$\rightarrow$MNIST setting.

Finally, in the Synth Signs $\rightarrow$ GTSRB experiment, SBADA-GAN is just slightly worse than DA$_{ass}$, but outperforms all the other competing methods. The comparison remains in favor of SBADA-GAN when considering that its performance is robust to hyperparameter variations (see Sec. ~\ref{subsec:sbadagan_ablation} for more details).

\subsection{Qualitative Results}
\begin{figure}[tb]
\centering
    \includegraphics[trim={3cm 10cm 11.5cm 9cm},clip, width=0.8\textwidth]{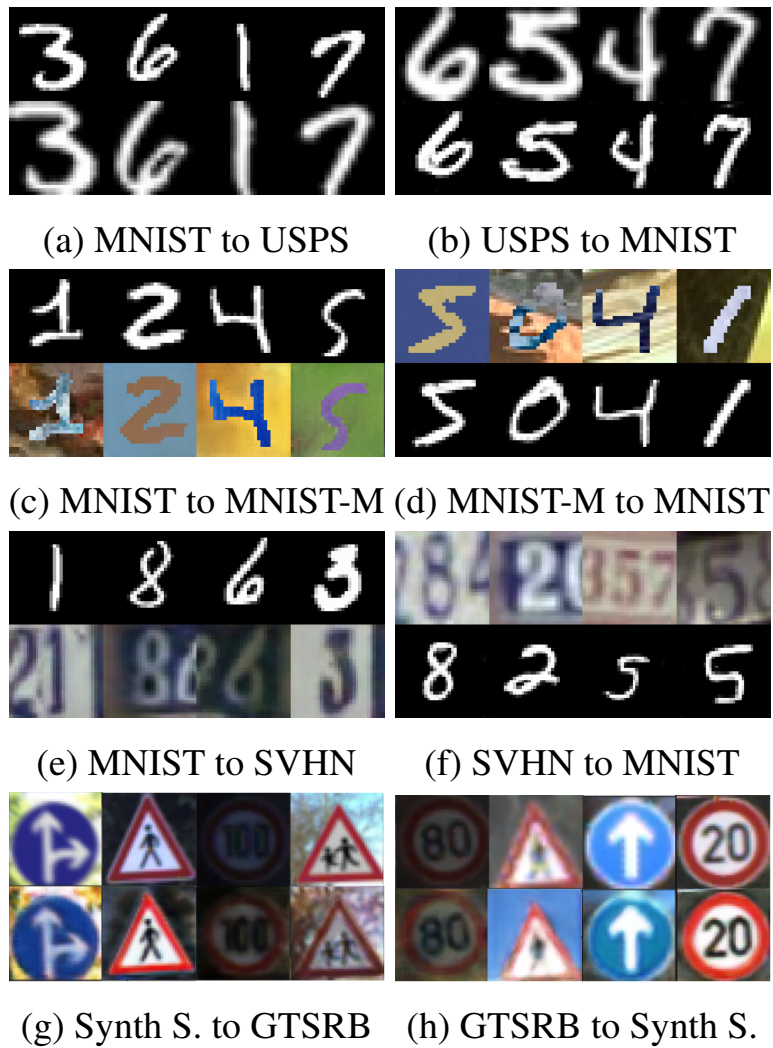}
\caption{Examples of generated digits: we show the image transformation
    from the original domain to the paired one as indicated under every sub-figure.
    For each of the (a)-(h) cases, the original/generated images are in the top/bottom row.
    }\label{fig:sbadagan_examples_best}
\end{figure}

To complement the quantitative evaluation, we look at the quality of the images generated by SBADA-GAN. First, we see from Figure ~\ref{fig:sbadagan_examples_best} how the generated images mimic the style of the chosen domain, even when going from the simple MNIST digits to the SVHN colorful house numbers.

\begin{figure}[tb]
    \centering
    \includegraphics[trim={4cm 9cm 4cm 8cm},clip, width=0.95\textwidth]{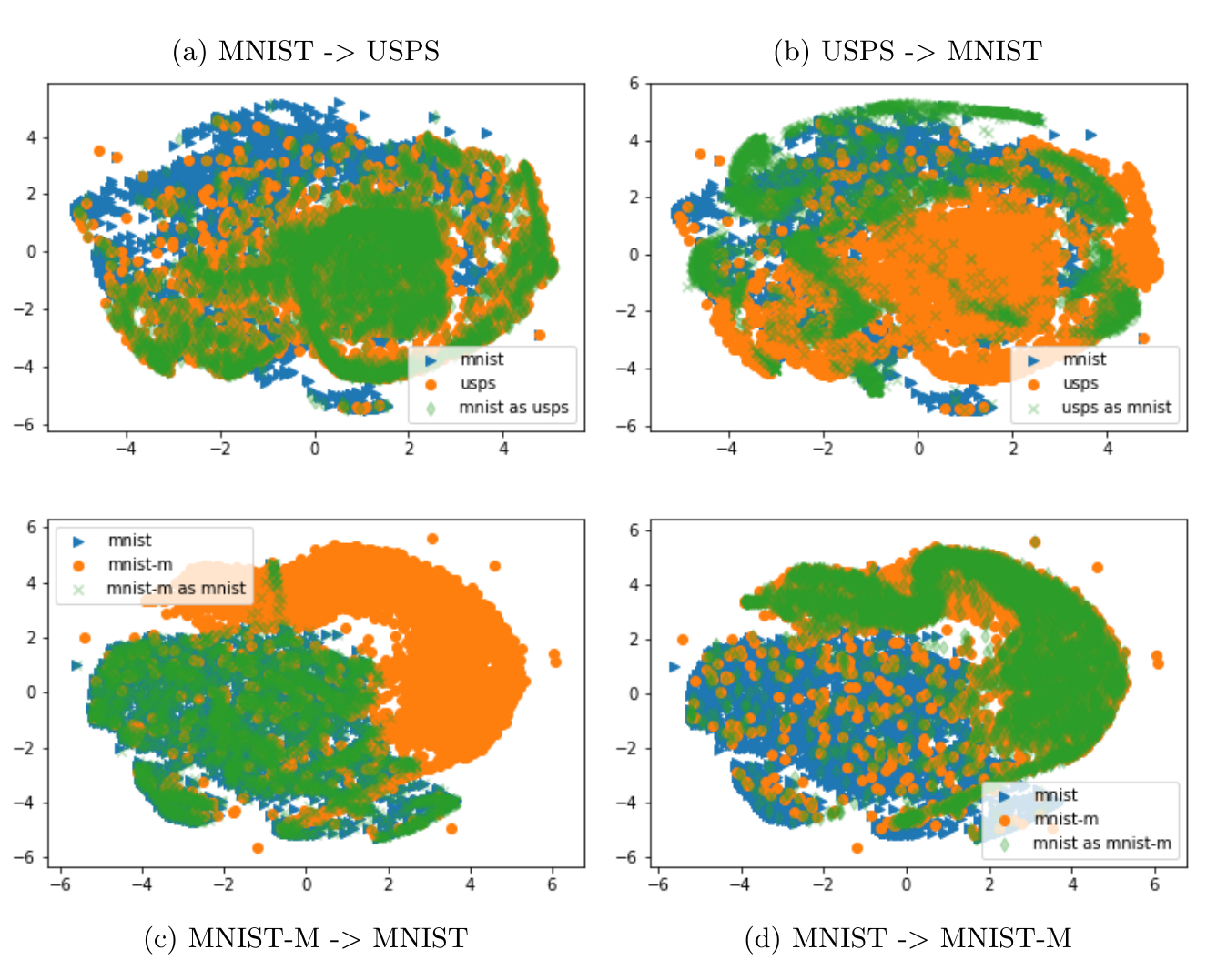}
    \caption{t-SNE visualization of source, target and source mapped to target images. 
    Note how the mapped source covers faithfully the target space both in the (a),(b)
    case with moderated domain shift and in the more challenging (c),(d) setting.}\label{fig:TSNE} 
\end{figure}

Visually inspecting the data distribution before and after domain mapping defines a second qualitative
evaluation metric. We use t-SNE~\cite{maaten2008visualizing} to project the data from their 
raw pixel space to a simplified 2D embedding. Figure ~\ref{fig:TSNE} shows 
that the transformed dataset tends to replicate faithfully
the distribution of the chosen final domain.

A further measure of the quality of the SBADA-GAN generators 
comes from the diversity of the produced images. Indeed, GAN's 
generators may collapse and output a single prototype that maximally fools the discriminators.
To evaluate the diversity of samples generated by SBADA-GAN we choose 
the Structural Similarity (SSIM, ~\cite{Wang:2004:IQA}) that correlates well with the human perceptual similarity judgments. 
Its values range between $0$ and $1$ with higher values  corresponding to more similar images. 
We follow the same procedure used in ~\cite{pmlr-v70-odena17a} by randomly choosing $1000$ 
pairs of generated images within a given class. 
We also repeat the evaluation over all the classes and calculate the average results.
Table ~\ref{table:sbadagan_data_ssim} shows the results of the mean SSIM metric
and indicates that the SBADA-GAN generated images not only mimic the same style, but also
successfully reproduce the variability of a chosen domain.

\begin{table}[tb]
\centering
\begin{tabu}{lcccc}
\hline
\rowfont{\scriptsize} Setting & S   & T map to S & S map to T & T   \\
\hline
\rowfont{\scriptsize} MNIST $\rightarrow$ USPS    & $0.206$  &   $0.219$      &   $0.106$      & $0.102$  \\
\rowfont{\scriptsize} MNIST $\rightarrow$ MNIST-M & $0.206$  &   $0.207$      &  $0.035$   & $0.032$  \\     
\rowfont{\scriptsize} MNIST $\rightarrow$ SVHN    & $0.206$  &   $0.292$      &  $0.027$       & $0.012$  \\
\rowfont{\scriptsize} Synth S. $\rightarrow$ GTSRB       & $0.105$  &   $0.136$      &  $0.128$       & $0.154$  \\
\hline
\end{tabu}
\caption{Dataset mean SSIM: this measure of data variability suggests 
that our method successfully generates images with not only the same style of a chosen domain, 
but also similar perceptual variability.}
\label{table:sbadagan_data_ssim} 
\end{table}

\subsection{Method Analysis}
\label{subsec:sbadagan_ablation}

\paragraph{Ablation Study}
Starting from the core source-to-target GAN module we analyze the effect of adding all the other 
model parts. At first we add the symmetric target-to-source GAN model.
These two parts are then combined and the domain transformation loop is closed 
by adding the class consistency condition. Finally the model is completed by introducing the 
target self-labeling procedure. We empirically test each of these model steps 
on the MNIST$\rightarrow$USPS setting and report the results in Table ~\ref{table:ablation}.
We see the gain achieved by progressively adding the different components, with the largest 
advantage obtained by the introduction of self-labeling. 

An analogous boost due to self-labeling is also visible in all the other experimental
settings with the exception of  MNIST$\leftrightarrow$SVHN, where the accuracy remains
unchanged if $\eta$ is equal or larger than zero. A further analysis reveals that here 
the recognition accuracy of the source classifier applied to the source-like transformed target images
is quite low (about $65\%$, while in all the other settings reaches $80-90\%$), thus the 
pseudo-labels cannot be considered reliable. Still, using them does not hinder the 
overall performance.
\begin{table}[htb]
\centering
\begin{tabu}{|c|c|c|c|c|c|@{}c@{}|}
\hline
\rowfont{\small}\multicolumn{2}{|c|}{S$\rightarrow$T} & \multicolumn{2}{c|}{T$\rightarrow$S}& Class & Self & \multirow{2}{*}{Accuracy}\\
\rowfont{\small}\multicolumn{2}{|c|}{GAN}      & \multicolumn{2}{c|}{GAN} & Consist. & Label. &\\
\cline{1-6}
\rowfont{\small}
\parbox[t]{2mm}{\rotatebox[origin=c]{90}{$\mathcal{L}_{D_{t}}$}}& %(D_{t},G_{st})
\parbox[t]{2mm}{\rotatebox[origin=c]{90}{$\mathcal{L}_{C_{t}}$}}& %(G_{st},C_{t})
\parbox[t]{2mm}{\rotatebox[origin=c]{90}{$\mathcal{L}_{D_{s}}$}}& %(D_{ts},G_{ts})
\parbox[t]{2mm}{\rotatebox[origin=c]{90}{$\mathcal{L}_{C_{s}}$}}& %(C_{s})
\parbox[t]{2mm}{\rotatebox[origin=c]{90}{$\mathcal{L}_{cons}$}}& %(G_{ts},C_{s})
\parbox[t]{2mm}{\rotatebox[origin=c]{90}{$\mathcal{L}_{self}$}}& %(G_{ts},G_{st},C_{s})
~~MNIST$\rightarrow$USPS~~\\						\hline
$\checkmark$ & $\checkmark$ &&&&&94.23 \\
&&$\checkmark$&$\checkmark$&&&91.55 \\
$\checkmark$&$\checkmark$&$\checkmark$&$\checkmark$&&&94.90 \\
$\checkmark$&$\checkmark$&$\checkmark$&$\checkmark$&$\checkmark$&&95.45 \\
$\checkmark$&$\checkmark$&$\checkmark$&$\checkmark$&$\checkmark$&$\checkmark$&97.60\\
\hline
\end{tabu}
\caption{Analysis of the role of each SBADA-GAN component. We ran experiments by turning on
the different losses of the model as indicated by the checkmarks. 
}
\label{table:ablation}
\end{table}
\begin{figure}[htb]
    \centering
    \includegraphics[trim={2.2cm 13cm 11cm 11.5cm},clip,width=0.9\textwidth] {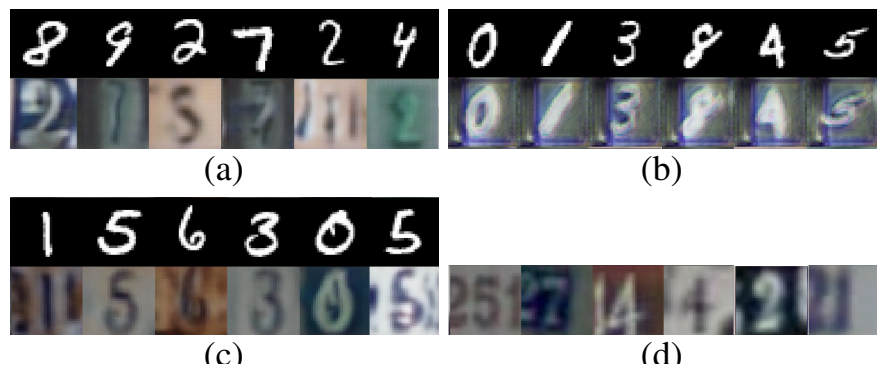}
    \caption{$G_{ts}$ outputs (lower line) and their respective inputs (upper line) obtained with: 
              (a) no consistency loss, (b) image-based cycle consistency loss ~\cite{CycleGAN2017,DBLP:conf/icml/KimCKLK17}, 
              (c) our class consistency loss. 
              In (d) we show some real SVHN samples as a reference.}\label{fig:sbadagan_rec_comparison}
\end{figure}

\begin{figure}[htb]
    \centering
    \includegraphics[trim={2.2cm 13cm 11.4cm 11.7cm},clip,width=0.9\textwidth] {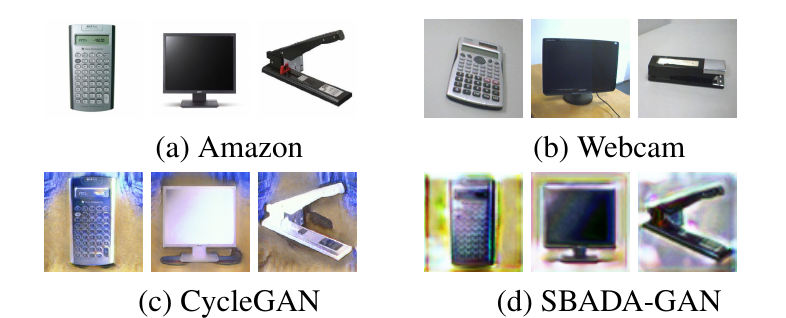}
    \caption{CycleGAN ~\cite{CycleGAN2017} vs SBADA-GAN on the Amazon-Webcam experiment of the 
    Office Dataset ~\cite{Saenko:2010}.}\label{fig:sbadagan_office}
\end{figure}

The crucial effect of the class consistency loss can be better observed by looking 
at the generated images and comparing them with those obtained in two alternative cases: 
setting $\nu=0$, \ie not using any consistency condition between the two generators $G_{st}$ and $G_{ts}$, 
or substituting our class consistency loss with the standard cycle consistency loss ~\cite{CycleGAN2017,DBLP:conf/icml/KimCKLK17}
based on image reconstruction.
For this evaluation we choose the MNIST$\rightarrow$SVHN case which has the strongest
domain shift and we show the generated images in Figure ~\ref{fig:sbadagan_rec_comparison}.
When the consistency loss is not activated, the $G_{ts}$ output images are realistic, but fail at reproducing the correct input digit and provide misleading information to the classifier. 
On the other hand, using the cycle-consistency loss preserves the input digit but fails in rendering a realistic sample in the correct domain style. 
Finally, our class consistency loss allows to preserve the distinct features belonging to a category while still leaving enough freedom to the generation process, thus it succeeds in both preserving the digits and rendering realistic samples.  

\paragraph{CycleGAN vs SBADA-GAN}
To further clarify the difference between the two methods, we remind that CycleGAN is unsupervised and works only when transferring style across similarly shaped categories (\eg horses$\rightarrow$zebras), not across domains. SBADA-GAN instead deals with domains containing multiple categories. The images samples in Figure ~\ref{fig:sbadagan_rec_comparison}(b) are indeed obtained with CycleGAN: training on them produces an accuracy of 25.5\%, much lower than the corresponding 61.1\%  of SBAD-GAN. 
Moreover, CycleGAN has a single transformed image as output, while SBADA-GAN exploits a noise vector as input producing multiple outputs for each input image: this is critical for classification as it provides variability through data augmentation, it avoids overfitting and eases generalization.
For completeness we also ran an experiment on the challenging Office Dataset ~\cite{Saenko:2010}: here both the images produced by CycleGAN and SBADA-GAN (see Figure ~\ref{fig:sbadagan_office}) are given as input to a pre-trained AlexNet and the classification accuracy is respectively 52.0\% and 50.7\%, both lower than the reference 61.6\% result produced by the baseline without adaptation. These results confirm the known difficulty of GAN-based method to deal with domain shifts due to poses and shapes.

\paragraph{Robustness Study}
SBADA-GAN is robust to the specific choice of the consistency loss weight $\nu$, given that it is different from zero. Changing it in $[0.1,1,10]$ induces a maximum variation of $0.6$ percentage points in accuracy over the different settings. An analogous evaluation performed on the classification loss weights $(\beta$,$\mu)$ reveals that changing them in the same range used for $\nu$ causes a maximum overall performance variation of $0.2$ percentage points. 
Furthermore SBADA-GAN is minimally sensitive to the  batch size used: halving it from $32$ to $16$ samples while keeping the same number of learning epochs reduces the performance only of about $0.2$ percentage points.
Such robustness is particularly relevant when compared to competing 
methods. For instance the most recent $DA_{ass}$~\cite{haeusser17} needs a perfectly balanced source and target 
distribution of classes in each batch, a condition difficult to satisfy in real world scenarios, and halving the 
originally large batch size reduces by $3.5$ percentage points the final accuracy. Moreover, changing the weights 
of the losses that enforce associations across domains with a range analogous to that used for the SBADA-GAN parameters 
induces a drop in performance up to $16$ accuracy percentage points\footnote{More details are provided in the supplementary material.}.

\subsection{Conclusion}
This section presented SBADA-GAN, an adaptive adversarial domain adaptation architecture that simultaneously maps  source samples into the target domain and vice versa with the aim to learn and use both classifiers at test time. To achieve this, self-labeling is exploited to regularize the classifier trained on the source, and  we impose a class consistency loss that improves greatly the stability of the architecture, as well as the quality of the reconstructed images in both domains. 

We explain the success of SBADA-GAN in several ways.
To begin with, thanks to the the bi-directional mapping we avoid deciding a priori which is the best strategy for a specific task. Also, the combination of the two network directions improves performance providing empirical evidence that they are learning different, complementary features.
Our class consistency loss aligns the image generators, allowing both domain transfers to influence each other.  Finally the self-labeling procedure boost the performance in case of moderate domain shift without hindering it in case of large domain gaps.

\section{A Hybrid Approach: ADAGE}
\label{sec:da_hybrid_align}
\textit{The ability to generalize across visual domains is crucial for the robustness of visual recognition systems in the wild. Several works have been dedicated to close the gap between a single labeled source domain and a target domain with transductive access to its data. In this section we focus on the wider domain generalization task involving multiple sources and seamlessly extending to unsupervised domain adaptation when unlabeled target samples are available at training time. We propose a hybrid architecture that we name ADAGE: it gracefully maps different source data towards an agnostic visual domain through pixel-adaptation based on a novel %layer-aggregative 
incremental architecture, and closes the remaining domain gap through feature adaptation. Both the adaptive processes are guided by adversarial learning. Extensive experiments show remarkable improvements compared to the state of the art.}
\newline

As we've seen before, Domain Adaptation (DA) is at its core the quest for principled algorithms enabling the generalization of visual recognition methods. Given at least a source domain for training, the goal is to achieve recognition results as good as those achievable on source test data on \emph{any}  other target domain, in principle belonging to a different probability distribution, without having prior access to labeled images. Solving this problem will represent a major step towards one of the key goals of computer vision, i.e. having machines able to answer the question `what do you see?' in the wild; hence, its increased popularity in the community over the last years (see section ~\ref{sec:domain_adap_related} for a review of recent work).

Since its definition ~\cite{Saenko:2010}, the most popular instantiation of the problem has assumed to have access to annotated data from a single source domain and to unlabeled data from a specific target domain. Still, there has been recently a growing interest on how to leverage over multiple source domains when unlabeled target data are available ~\cite{cocktail_CVPR18}, and even more on how to generalize over \emph{any}  possible target domain, when it is not possible to access target data of any sort a priori ~\cite{hospedalesPACS}. Intuitively, by leveraging over multiple sources it should be possible to design algorithms able to discard the specific style of each source domain, while capturing the generic content of the visual categories contained in all domains depicting such categories.   

Algorithm-wise, the community has attacked the problem with two disjoint strategies. The first is based on end-to-end architectures that minimize both a source classification and a domain shift measure. Specifically, these methods express the domain shift either in terms of difference in the domain statistics ~\cite{Tzeng:MMD:arxiv14,Sun:CORAL:AAAI16}, or  of 
adversarial domain discrimination ~\cite{Ganin:DANN:JMLR16,Hoffman:Adda:CVPR17}, or  of random walk transition probability among the samples of two 
domains ~\cite{haeusser17}. All of them aim at aligning the feature representation learned for the domains.
The other direction deals with image (instead of feature) transformation. The visual style of a domain, be it source, target or both, is modeled and 
transferred to images of the different one so that adaptation happens at the input level and standard classification networks
can be reused without further internal modifications ~\cite{russo17sbadagan,Bousmalis:Google:CVPR17}. While this second solution allows for 
human-understandable modifications, it should be noted that a visually pleasant transformation may not be what the network needs most to get the best possible adaptation.

In this work we focus on the scenario where multiple source domains are available at training time, with the aim of learning an \emph{agnostic} visual domain as well as the corresponding representation able to capture the intrinsic information carried by all domains while discarding the distinctive style of each individual source. To do so, we propose a hybrid architecture that sits at the intersection of the two algorithmic approaches outlined above, allowing to get the best of both worlds. Our intuition is that it is possible to reduce the domain shift by acting simultaneously on the image and on the feature space within an adversarial framework. This means dropping the condition of a human-understandable style transfer to the benefit of a new more 
network-understandable visual language. To this end, we propose an architecture where a novel incremental transformer %block  
maps the available images, guided by an adversarial loss,  towards an \emph{agnostic}, intermediate visual domain that retains the most important domain invariant information. Such agnostic images are then used as input to a multi-branch feature adaptation block, getting 
a substantial benefit with respect to separate feature-based and style-transfer based methods.
We call our architecture Agnostic Domain Generalization (ADAGE).
While in general we do not assume to have access to unlabeled target data, the architecture can be easily extended to the unsupervised multi-source domain adaptation scenario.

We test ADAGE on the domain generalization and unsupervised multi source  domain adaptation settings, comparing against 
recent approaches ~\cite{MLDG_AAA18,MDAN_ICLRW18,cocktail_CVPR18}. In all experiments, for both settings, ADAGE significantly outperforms the state of the art, proving the benefit of a hybrid approach. An ablation study and visualizations of the agnostic domain images complete our experimental study.

%%%%%%%%%%%%%%%%%%%%%%%%%%%%%%%%%
%%%%%%%%%%%ARCHITECTURE%%%%%%%%%%
%%%%%%%%%%%%%%%%%%%%%%%%%%%%%%%%%
\subsection{Agnostic Domain Generalization}
\label{sec:adage_architecture}

\begin{figure}[htb]
\centering
\includegraphics[width=0.95\textwidth]{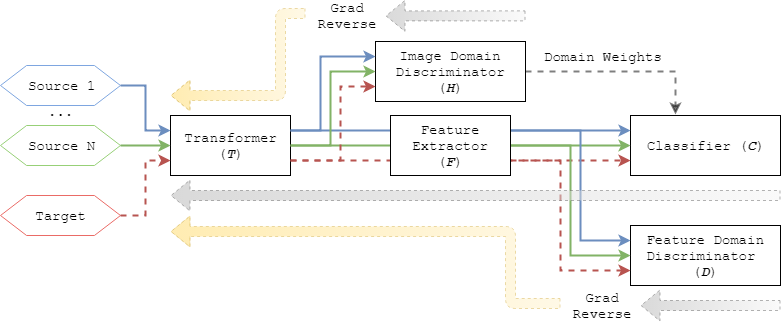}
\caption{A brief description of our architecture: all samples (including target ones, in the DA setting)
follow the same path in the network. Firstly they go through the Transformer $T$ that takes as input the 
original images and outputs the modified ones. Then the updated samples go through the Image Domain Discriminator $H$
and, at the same time, flow through the Feature Extractor $F$. Once converted to features, the samples 
progress both into the Classifier $C$ and through the Feature Domain Discriminator $D$. We use $H$ to get an estimate of domain similarity, which is used to bias the classification loss towards those sources which are more similar to the target.
The gradient from $H$ is inverted and flows through $T$ driving image modifications towards
domain confusion. Similarly, the gradient from $D$ also inverted, is backpropagated through $F$ and $T$
so that both the feature and the image dedicated blocks benefit from a further push towards the domain agnostic
space. The classification gradient travels through the whole network, excluding $H$ and $D$.}
\label{fig:adage_architecture}
\end{figure}

\begin{figure}[tb]
\centering
\includegraphics[width=0.95\textwidth]{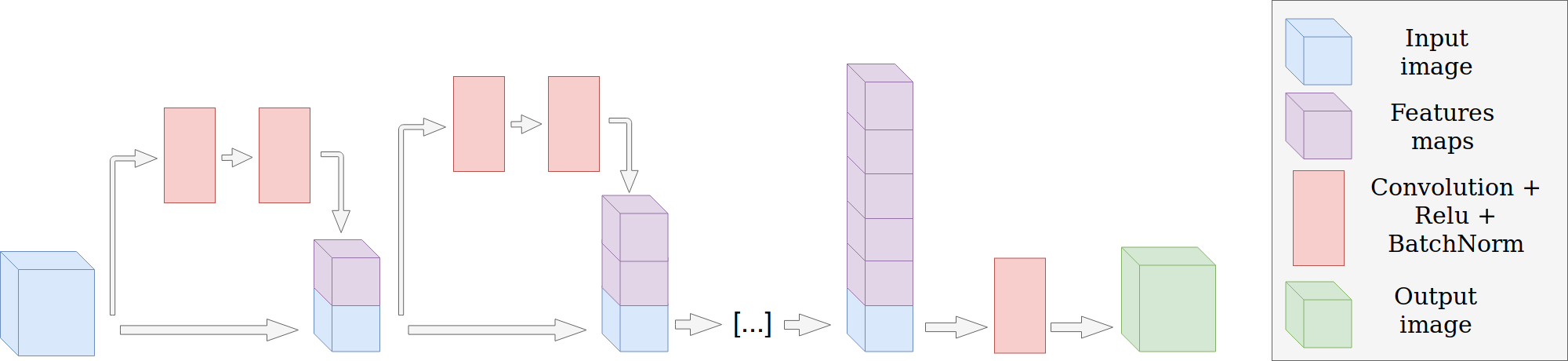}
\caption{Main blocks of our incremental Transformer network. The blue blocks represent the input image data, while the red blocks are a sequence of convolutional + Relu + Batch Normalization layers. The output of the two convolutional blocks are concatenated with the previous inputs, forming a group of image data and features maps that grow along the depth of the network. The number of features increases from 3 (input data) to 256 (final aggregation step), while a last convolutional layer squeezes the features back into 3 channels interpretable as an RGB image (green block).}
\label{fig:adage_transformer}
\centering
\end{figure}

We assume to observe $i=1\ldots S$ source domains with the $i$th domain containing $N_i$ 
labeled instances $\{x_j^i,y_j^i\}_{j=1}^{N_i}$, where $x_j^i$ is the $j$th 
input image and $y_j^i\in\{1\ldots M\}$ is the class label. 
In addition we also have an unlabeled target domain whose data $\{x_j^{t}\}_{j=1}^{N_t}$ might 
(DA) or might not (DG) be provided at training time. All the source and target domains
share the same label space and overall they have the same conditional distribution, but their marginal distribution is different thus inducing a domain shift. 

The goal of \adage is to learn a domain agnostic model by adapting both the images and 
their representation: this is obtained by jointly learning a mapping towards new visual 
and feature spaces where the domains are confused but the relevant semantic information 
of the data is maintained. We realize this goal by combining a \emph{feature extractor} $F$
and a \emph{classifier} $C$ sub-networks to three modules: a new \emph{image transformer} $T$
and two \emph{domain discriminators} $H,D$ respectively for images and features.
An overview of our deep learning architecture is shown in Figure ~\ref{fig:adage_architecture}.

\textbf{Image Transformer} $T$ modifies the input images to remove their domain-specific style. 
We defined a new \emph{incremental} structure for this module that exploits the power of layer 
aggregation ~\cite{shelhamerdeep}: the output of two $3\times 3$ convolutional layers each 
followed by Relu and Batch Normalization are stacked up with the
input and propagated to every subsequent layer (see Figure ~\ref{fig:adage_transformer}).
Specifically, the produced feature build up in size resulting in a growing sequence of 
$\{3,8,16,32,4,64,128\}$ maps, after which a convolution layer brings them down to 3 channels, 
interpretable as RGB images.

\textbf{Image Domain Discriminator} $H$ receives as input the images produced by $T$
and predicts their domain label. More in details, this module is a multi-class classifier 
that learns to distinguish among the $S$ source domains in DG, and $S+1$ in DA 
(including the target), by minimizing a simple cross-entropy loss $\mathcal{L}_{H}$.
The information provided by this module is used in two ways:  to adversarially 
guide the transformer $T$ to produce images with confused domain identity, and
to estimate a similarity measure between the source and the target data when available.
The first task is executed through a gradient reversal layer as in ~\cite{Ganin:DANN:JMLR16}.
The second is obtained as a byproduct of the domain classifier $H$ by collecting the
probability of every source sample in each batch to be recognized as belonging to the target.

\textbf{Feature Domain Discriminator} $D$ is analogous to $H$ but, instead of images,
it takes as input their features, performing domain classification by minimizing the cross-entropy loss $\mathcal{L}_{D}$. Finally, during backpropagation, the inverted gradient regulates the feature extraction process to confuse the domains.

\textbf{Feature Extractor and Classifier} $F$ and $C$ are standard deep learning modules. 
 We built them with the same network structure used in ~\cite{MDAN_ICLRW18} to 
put them on equal footing. In particular, in the DG setting the classifier learns to 
distinguish among the $M$ categories of the sources by minimizing the cross-entropy 
loss $\mathcal{L}_{C}$, while for the DA setting it can also provide the classification 
probability on the target samples $p(x^t)=C(F(T(x_t)))$ that is used to minimize the 
related entropy loss $\mathcal{L}_{e}=p(x^t)log(p(x^t))$. 

If we indicate with $\theta$ the network parameters and we use subscripts to identify the
different network modules, we can formally write the overall loss function optimized 
by \adage as:
\begin{align}
\mathcal{L}(\theta_T,\theta_F,\theta_D,\theta_{H},\theta_C) = 
\sum_{i=1..S,S+1}\sum_{j=1..N^i} & \mathcal{L}_C^{j,i}(\theta_T,\theta_F,\theta_C) + \eta \mathcal{L}_{e}^{j,i=S+1}(\theta_T,\theta_F,\theta_C)\nonumber\\
     & - \lambda \mathcal{L}^{j,i}_D(\theta_T,\theta_F,\theta_D) - \gamma \mathcal{L}^{j,i}_H(\theta_T,\theta_H)~.
\label{eq:adage_main_loss}
\end{align}
We remark that, as specified by its superscripts, $ \mathcal{L}_{e}^{j,i=S+1}$ is only active in the DA setting, while $\mathcal{L}_D$ and $\mathcal{L}_H$ in the DA case deal with an $\{S+1\}$-multiclass task involving also the target together with the source domains.

As can be noted from (~\ref{eq:adage_main_loss}), the number of meta-parameters of our approach is very limited.
For $\lambda$ we use the same rule introduced by ~\cite{Ganin:DANN:JMLR16} that grows the importance of the feature domain discriminator with the training epochs: 
$\lambda_k = \frac{2}{1+exp(-10k)}-1$, where $k = \frac{current\_epoch}{total\_epochs}$. 
We set $\gamma_k=0.1 \lambda_k$ so that only a small portion of the full gradient of the image domain discriminator is backpropagated: in this way we can still get useful similarity measures among the domains while progressively guiding the transformer to make them alike. Finally, the experimental evaluation indicates that \adage is robust to the exact choice of $\eta$, thus we keep it always fixed to 0.5 just for simplicity.

We conclude this section with some remarks on the characteristics of \adage. To our knowledge this is the first method designed to work seamlessly both in the domain generalization and in the unsupervised domain adaptation settings. It is also the first method to introduce an image-level component in a deep learning architecture for domain generalization. Differently from existing GAN-based methods that need a typical alternating training between image adaptation and classification, we train the whole model of \adage with a single optimizer. Nonetheless, we are still performing adversarial training, as the gradient originating from our domain discriminators is inverted before reaching our feature extractor and image transformer.
Moreover, GAN adaptive approaches aim at transferring the source style to the target data and/or vice-versa ~\cite{russo17sbadagan,Bousmalis:Google:CVPR17,liu2016coupled}, while our goal is that of projecting the data of all the available domains to a new agnostic space, where the domain-specific signatures are discarded. Technically we avoid the risk of degenerating all inputs to random noise  by priming the network to correctly perform label classification and by starting to confuse the domains only later (see the $\lambda_k$ update agenda), while still receiving feedback from the classification loss.
Finally we underline that the image transformation proposed for \adage is not meant to be pleasant to the human eye: its purpose is to start to close the domain gap, instead of fully delegating this task to the features at later stages in the learning process.

%%%%%%%%%%%%%%%%%%%%%%%%%%%%%%%
%%%%%%%% EXPERIMENTS %%%%%%%%%%
%%%%%%%%%%%%%%%%%%%%%%%%%%%%%%%
\subsection{Experiments}
\label{sec:adage_experiments}

We tested \adage on the DA and DG scenarios always considering the availability of multiple sources.
Our framework can easily switch between the two cases with a few key differences. For DG the image $H$ 
and the feature $D$ domain discriminators deal with $S$ domains, while for DA they need to distinguish among
$S+1$ domains including the target. Moreover, in DA, the unlabeled target data trigger the classification
block $C$ to activate the entropy loss and to use the source domain weights provided by the image 
domain discriminator $H$. Specifically these weights make sure that our classifier is biased towards
the sources more similar to the target.

\textbf{Datasets and Scenarios} 
We focus on five well known digits datasets: \textbf{MNIST}~\cite{lecun1998gradient}, \textbf{MNIST-M}~\cite{Ganin:DANN:JMLR16}, \textbf{USPS}~\cite{friedman2001elements}, \textbf{SVHN}~\cite{netzer2011reading} and (\textbf{SYNTH Digits}) (dataset details are in section ~\ref{sec:da_datasets}).

To define the multi source experimental scenarios we follow ~\cite{cocktail_CVPR18,MDAN_ICLRW18} and 
reproduce their settings.
A first case from ~\cite{MDAN_ICLRW18} involves \emph{three sources} chosen in \{MNIST, MNIST-M, SYNTH, SVHN\}.
Each dataset with the exception of SYNTH, is cyclically used as target. All the images are resized to
$28\times28$ pixels and subsets of $20$k and $9$k samples are chosen respectively from each source
and from the target.
A second case from ~\cite{cocktail_CVPR18} involves \emph{four sources} by adding USPS to the previous dataset group, and focuses on two possible targets, SVHN and MNIST-M. Even in this case the imagesare resized to $28\times28$ pixels, and $25$/$9$k samples are drawn from each dataset to define the source/target sets\footnote{The authors of ~\cite{cocktail_CVPR18} kindly shared the exact splits used for their experiments. For the three and five sources experiments we considered instead multiple random selections of the samples from the datasets.}. 
A third case from ~\cite{DGautoencoders} involves \emph{five sources} and exploits variants of MNIST
denoted as \{$M_{0},M_{15},M_{30},M_{45},M_{60},M_{75}$\}. We randomly chose 1000 digit images of ten classes from the original MNIST training set to represent the basic view $M_{0}$ with 100 images for each class. 
The other views are then obtained by rotating the images of $15$ degrees in counterclock-wise direction.  

For our experiments all the datasets were normalized and zero-centered. In the DG case, 
the mean and standard deviation of the target for data normalization are calculated 
batch-by-batch during the testing process. 
A standard random crop of $90-100\%$ of the total image size was applied as data augmentation.
The training procedure requires $200$ epochs for DA, while for the DG experiments we found beneficial to increase the number of training epochs to $600$. For DA we used RmsProp ~\cite{rmsprop}
with a lr of $5e^{-4}$, while for \DG we used Adam ~\cite{kingma2014adam} with a lr of $1e-3$. 
In both cases we step down the lr after $80\%$ of the training.

All the experiments are repeated tree times and we report the average on the obtained classification
accuracy results.

\begin{table}
    \centering
    \begin{tabular}{@{}l@{~~}l@{}c@{}c@{}c@{}c}  %\rowfont{\footnotesize}
\hline
%\multicolumn{2}{c}{SVHN MnistM SYNTH $\rightarrow$ Mnist}   \\
   \multicolumn{2}{c}{ \multirow{3}{*}{Sources}}& SVHN & SVHN & MNIST-M&\multirow{4}{*}{Avg.}\\ [0.82ex]
  \multicolumn{2}{c}{ }& MNIST-M & MNIST & SYNTH\\ [0.82ex]
   \multicolumn{2}{c}{ }& SYNTH & SYNTH & MNIST\\ [0.82ex] \cline{2-5}
  \multicolumn{2}{c}{Target} & MNIST & MNIST-M & SVHN\\ \hline
 \multirow{4}{*}{DG}  & combine sources             &  98.7  & 62.6 & 69.5 & 76.9\\ 
  & MLDG ~\cite{MLDG_AAA18}   &  99.1  & 61.2 & 69.7 & 76.7\\\cline{2-6}
  & \adage Residual            &  \textbf{99.2}  & 65.8 & 74.6 & 79.9 \\
 & \adage Incremental         &  99.1  & \textbf{66.3} & \textbf{76.4}  & \textbf{80.3}\\
 \hline \hline
  \multirow{5}{*}{DA} & combine sources &  98.7  & 62.6 & 69.5 & 76.9\\ 
% \rowfont{\footnotesize}  & best single DANN ~\cite{MDAN_ICLRW18}   & 96.7    &  59.1 &  81.8 & 79.2\\
   & combine DANN ~\cite{MDAN_ICLRW18}   &  92.5   & 65.1  &  77.6 & 78.4\\
   & MDAN ~\cite{MDAN_ICLRW18}   &  97.9   &  68.7 &  81.6 & 82.7\\ \cline{2-6}
  & \adage Residual    &  99.2   &  87.6 &  84.1 & 90.3\\
   & \adage Incremental    &  \textbf{99.3}   &  \textbf{88.5} &  \textbf{86.0} & \textbf{91.3}\\
\hline
\end{tabular}
    \caption{Classification accuracy results: experiments with 3 sources.}
    \label{tab:adage_results_a}
\end{table}

\begin{table}
\centering
    \begin{tabular}{@{}l@{~~}l@{}c@{}c@{}c}
\hline
   \multicolumn{2}{c}{ \multirow{4}{*}{Sources}}& SYNTH & SYNTH &\multirow{5}{*}{Avg.}\\
   \multicolumn{2}{c}{ }& MNIST & MNIST &\\
   \multicolumn{2}{c}{ }& MNIST-M & SVHN &\\
   \multicolumn{2}{c}{ }& USPS & USPS &\\ \cline{2-4}
  \multicolumn{2}{c}{Target} & SVHN & MNIST-M &\\ \hline
   \multirow{4}{*}{DG}  & {combine sources}             &  73.2  &  61.9& 67.5\\ 
  & MLDG ~\cite{MLDG_AAA18}     &  68.0  &  65.6 &  66.8 \\ \cline{2-5}
                      & \adage  Residual             &  68.2  &  65.7 & 66.9 \\
                       & \adage Incremental  &  \textbf{75.8}  &  \textbf{67.0} & \textbf{71.4} \\
 \hline \hline
 \multirow{5}{*}{DA} & {combine sources}             &  73.2  &  61.9 & 67.5\\ 
% \rowfont{\footnotesize}  & separate DANN av.~\cite{cocktail_CVPR18}   &  61.4  &   71.1 & 66.3\\
  & combine DANN ~\cite{cocktail_CVPR18}   &  68.9   & 71.6  & 70.3 \\
  & DCTN ~\cite{cocktail_CVPR18}   &  77.5   &   70.9 & 74.2\\\cline{2-5}
  & \adage Residual    & 82.3    &  84.1  & 83.2\\
  & \adage Incremental    & \textbf{85.3}    &  \textbf{85.3}  & \textbf{85.3} \\
 \hline
\end{tabular}
    \caption{Classification accuracy results: experiments with 4 sources.}
    \label{tab:adage_results_b}
\end{table}

\begin{table}
    \centering
    \begin{tabular}{@{~}c@{~}@{~}l@{~}@{~}c@{~}@{~}c@{~}@{~}c@{~}@{~}c@{~}@{~}c@{~}@{~}c@{~}@{~}c@{~}} 
\hline
& Target & $M_{0}$ & $M_{15}$ & $M_{30}$ & $M_{45}$ & $M_{60}$ & $M_{75}$ & Avg.\\ \hline
  \multirow{5}{*}{DG} & D-MTAE ~\cite{DGautoencoders}   & 82.5 & 96.3 & 93.4 & 78.6 & 94.2 & 80.5 & 87.6\\
& CCSA ~\cite{doretto2017}   & 84.6 & 95.6 & 94.6 & 82.9 & 94.8 & 82.1 & 89.1 \\ 
& MMD-AAE ~\cite{Li_2018_CVPR}  & 83.7  & 96.9 & 95.7 & 85.2 & 95.9 & 81.2 & 89.8\\ 
& CROSS-GRAD ~\cite{DG_ICLR18}  & 88.3 & \textbf{98.6} & \textbf{98.0} & 97.7 & \textbf{97.7} & 91.4 & \textbf{95.3}\\ 
\cline{2-9}
& ADAGE Incremental & \textbf{88.8} & 97.6 & 97.5 & \textbf{97.8} & 97.6 & \textbf{91.9} & 95.2\\
\hline 
    \end{tabular}
    \caption{Domain Generalization accuracy results on experiments with 5 MNIST-rotated sources. For compactness we only indicate the considered target.}
    \label{table:rotatedminst}
\end{table}

\begin{table}[ht!]
\centering
\begin{tabu}{lcccc}
\hline
 & \multicolumn{2}{c}{Residual T} & \multicolumn{2}{c}{\quad Incremental T} \\
Mode & DG & DA & \quad DG & DA \\
\hline
T &  \multicolumn{2}{c}{61.73} & \multicolumn{2}{c}{\quad 63.23} \\
T + E & 61.7 & 56.7 & \quad 63.2 & 63.9\\
T + D & 62.1 & 65.4 & \quad 62.2 & 69.9\\
D     & 53.0 & 65.9 & \quad 53.0 & 65.9\\
D + E & 53.0 & 75.1 & \quad 53.0 & 75.1\\
T + H & 58.7  & 61.3  & \quad 61.4  & 60.8 \\
T + D + H & 59.0  & 62.5  & \quad 61.2  &  68.8\\
T + E + H & 58.7 & 61.6 & \quad 61.4 & 63.9\\
T + D + E & 62.2 & 82.9 & \quad 62.2 & 82.4\\
T + D + E + H & 65.8 & 87.6 & \quad 66.3 & 88.5\\
\hline
\end{tabu}
\caption{Ablation analysis on the experiment with three sources and target MNIST-M.
We turn on and off the different parts of the model: \textbf{T}= Transformer, 
\textbf{E}= Entropy, \textbf{D}= Feature Domain Discriminator, \textbf{H}= Image Domain Discriminator}
\label{table:ablation2}
\end{table}

\textbf{Results in Tables ~\ref{tab:adage_results_a} and ~\ref{tab:adage_results_b}} 
%Table ~\ref{tab:results} shows how \adage compares to existing DG and DA methods\footnote{We report in the supplementary material an extended version of this table with more baseline results.}. 
As a main baseline for the three and four sources settings we use the na\"ive \textbf{combine sources} 
strategy that consists in learning a classifier on all the source data combined together. 
For a fair comparison we produced these results by keeping only 
the feature extractor $F$ and the classifier $C$ of our network, while turning off all the adaptive blocks.
For DG we benchmark against the meta-learning method \textbf{MLDG} presented in ~\cite{MLDG_AAA18} using
the code provided by the authors and running the experiments on our settings. 
For DA we report the reference results from previous works. In particular for the three sources
experiments the comparison is with the Multisource Domain Adversarial Network \textbf{MDAN}~\cite{MDAN_ICLRW18}.
Since this method builds over the DANN algorithm ~\cite{Ganin:DANN:JMLR16} the result
obtained with DANN applied on the combination of all the sources (\textbf{combine DANN}) is also reported. 
For the four sources experiments the main comparison is instead with the Deep Cocktail Network 
(\textbf{DCN})~\cite{cocktail_CVPR18}, a recent method able to work even with partial class overlap among the sources. 
We present the accuracy for both the residual and incremental transformer variants of \adage and we see that 
they outperforms all the reference sota baselines in DG and DA, both using three and four sources, 
with margins of up to 6.84\% in DG and of up to 11.07\% in DA.  Interestingly, using four sources slightly 
worsens the performances when SVHN is the target: our interpretation is that adding the USPS dataset slightly
increases the domain shift between the whole training and test domains, making the adaptation somehow more difficult.
Results obtained with the incremental transformer are overall stronger that those obtained with the residual 
version: we think that this is due to the peculiar structure of the incremental transformer that allows to retain
all the expressive capacity of bigger and deeper architectures while keeping the number of parameters low.
Indeed the incremental $T$ has only \(\frac{1}{3}\)
of the parameters with respect to the residual version, thus it is faster in training and allows to 
better avoid overfitting while mapping the source domain images into a compact agnostic space.

\textbf{Results in Table ~\ref{table:rotatedminst}} For the five sources experiments we focus on DG and on the
most efficient incremental version of \adage. We benchmark against two autoencoder-based DG methods \textbf{D-MTAE}
and \textbf{MMD-AAE} respectively presented in ~\cite{DGautoencoders} and ~\cite{Li_2018_CVPR}, as well
as against the metric-learning \textbf{CCSA} method ~\cite{doretto2017} and the very recent \textbf{CROSS-GRAD} ~\cite{DG_ICLR18}.
The results indicate that \adage outperforms three of the four competitors and has results similar to CROSS-GRAD
which proposes an adaptive solution based on data augmentation that could potentially be combined with \adage.

\textbf{Further Results} Besides evaluating \adage on digits images, we tested it also on the 
\textbf{ETH80 object dataset}. We followed ~\cite{DGautoencoders} focusing on the ETH80-p setting with 5 domains 
obtained from 5 pitch-rotated views of 8 objects. The images are subsampled to $28\times 28$ and greyscaled.
By running \adage for DG on this setting, training in turn on four sources and testing on the remaining domain, 
we get an average accuracy of 94.1\%, significantly higher than 87.9\% of D-MTAE.

While \adage is specifically tailored for the multi-source settings, the reader might 
wonder how it would behave in the case of a \textbf{single source DA with access to unlabeled target data}. 
As a proof of concept experiment, we tested \adage using SVHN as source and MNIST as target.
With the same protocol used in our DA experiments, we achieve an $95.7\%$ accuracy, which is 
on par with the very recent ~\cite{haeusser17}
and better than many other competitive methods ~\cite{hoffman2017cycada,russo17sbadagan,CycleGAN2017,saito2017asymmetric,TRUDA-NIPS16_savarese}, 
all scoring an accuracy lower than $91.0\%$. 

\textbf{Ablation Study}
Table ~\ref{table:ablation2} shows the effect of progressively enabling the key components of \adage, 
showcasing the relative importance of each piece. If only the transformer is enabled, but there is no effort 
to align the domains, the final accuracy is not significantly better than the non-adaptive baseline ($62.6\%$), 
which shows that simply using a longer network provides only a minimal advantage. 
A first jump in accuracy appears when the feature domain discriminator $D$ is introduced, but only 
for DA, as the DG accuracy stays low.
The contribution of the image domain discriminator $H$ is negligible by itself and this 
behavior can be explained considering that we backpropagate only a
small part of the $H$ gradient ($\gamma = 0.1\lambda$, see section ~\ref{sec:adage_architecture}).
However its beneficial effect becomes evident in collaboration with the other network modules:
passing from T+D+E to T+D+E+H implies an improvement in accuracy of at least 4\%
which indicates that the adversarial guidance provided by $H$ on $T$ allows for an image adaptation process 
complementary to the feature adaptation one. 
Using the entropy loss helps a lot, with gains in performance of over 
 $15\%$. The presence of multiple sources very likely helps in reducing the risk that the entropy loss might 
 mislead the classifier.
Note that since the image domain discriminator backpropagates only on the transformer, it is not 
possible to test any combination containing $H$ but not $T$.

\begin{figure}[htb!]
    \centering
    \includegraphics[trim={4.5cm 6.5cm 4cm 8cm, width=0.8\textwidth},clip, width=0.95\textwidth]{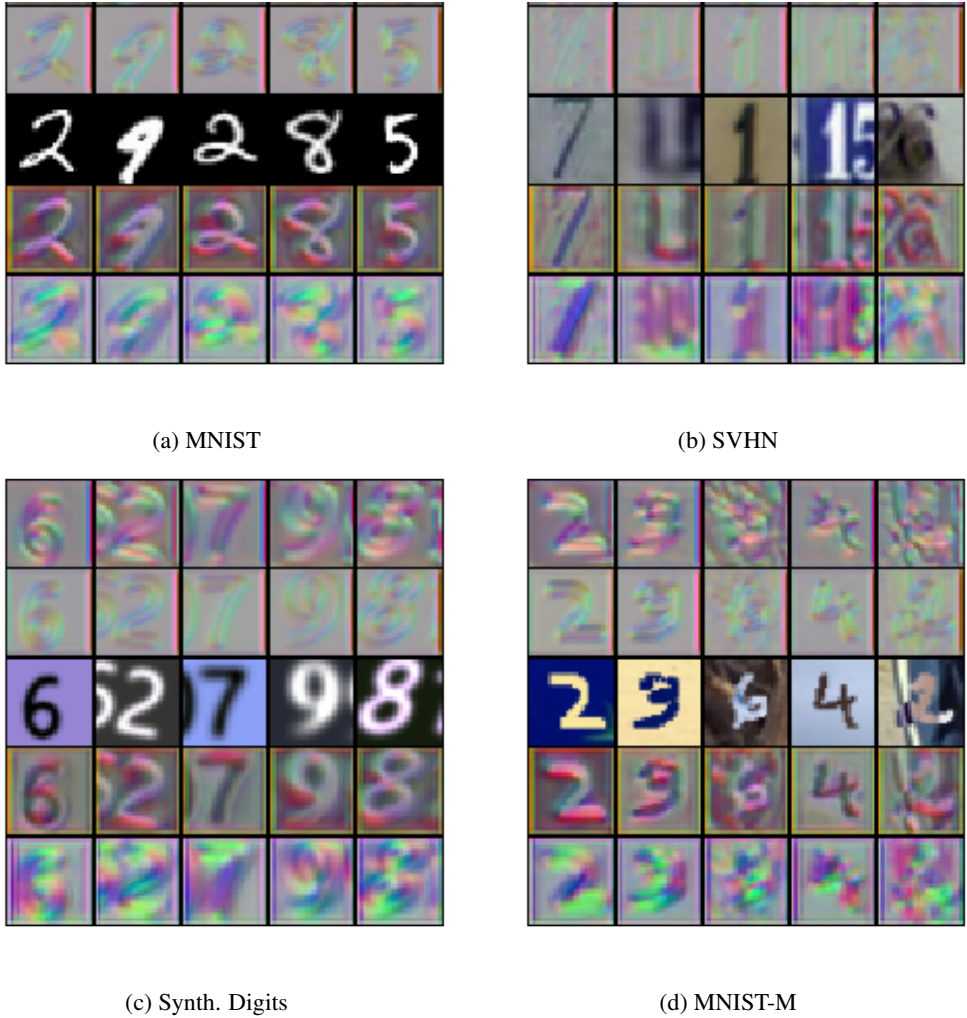}
\caption{
Examples of domain-agnostic digits generated by the transformer block in the experiments with three sources and MNIST-M as target.
The top two rows show images produced in the DG setting by our residual based (line $1$) and incremental based (line $2$) transformers. Line $3$ shows the original images and in the last two rows we display images produced by the residual (line $4$) and incremental transformers in the DA setting. Images transformed with the residual architecture tend to preserve more of the original input. It is worth mentioning that, while the DG images from the target class are more noisy than their equivalent in the DA setting, our transformer does a good job at transforming images that it has never seen before.
    }
    \label{fig:adage_examples_best_mnistm}
\end{figure}

\begin{figure}[hb!]
\centering
\includegraphics[trim={4cm 9.5cm 4.5cm 9.4cm},clip, width=0.95\textwidth]{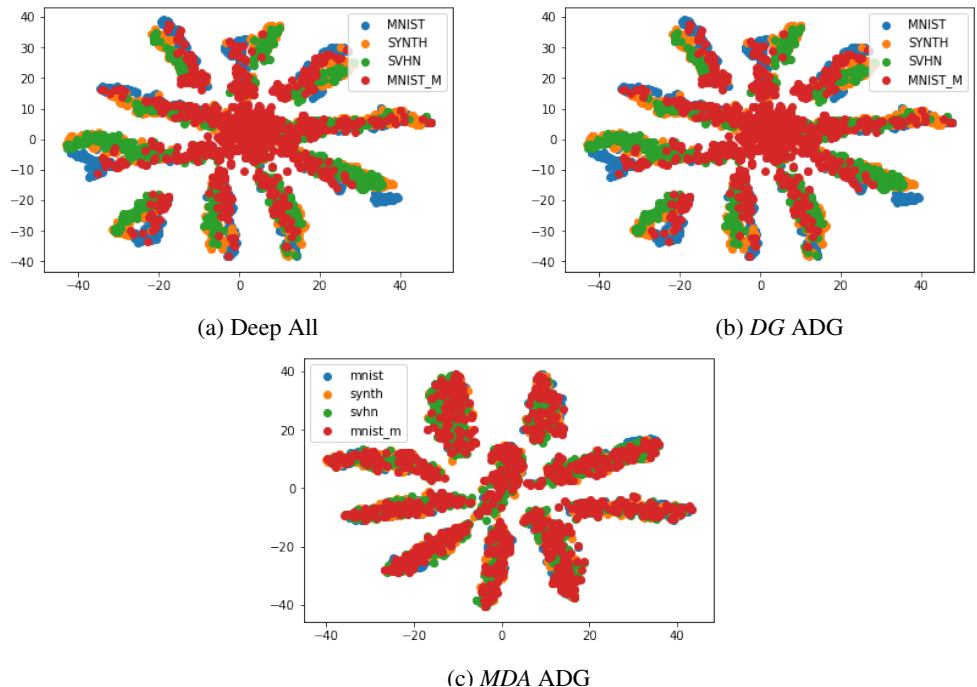}
 
\caption{This figure shows the TSNE visualization of the class distributions for the three source experiment, keeping MNIST-M as target. The features have been extracted from the penultimate layer. It is interesting to note that while Deep All does a reasonable job at aligning the domains, our method completely aligns each class for every domain.}
\label{fig:adage_tsne_mnistm}
\end{figure}

\textbf{Qualitative results}
While strong numerical results suggest that we are indeed closing the distances between domains, we can improve our understanding by looking at both the  images generated for the agnostic domain, and at their embeddings.
Figure ~\ref{fig:adage_examples_best_mnistm} shows the agnostic images generated by the residual and the incremental transformers, in the 
three source experiment with target MNIST-M, both in the DA and DG settings. We see that the main effect of $T$ is that of removing the backgrounds and enhancing the edges. We might still be able to distinguish between domains (more or less, depending on the 
used $T$) by looking at the style of the digits.  We believe this is why we also need domain invariance at the feature level. 
Fig ~\ref{fig:adage_tsne_mnistm} shows the TSNE embedding of features extracted immediately before the final classifier. 
We see that in the DA setting we completely align the feature spaces of the domains, resulting in a clear per class clustering. 
In the DG setting the results are less clean, but the clusters are still tighter than those obtained by the combine source baseline.

\subsection{Conclusions}
This section tackles the problem of domain generalization and adaptation when multiple sources are available, proposing the first deep architecture for these settings which jointly performs image and feature adaptation. This makes it possible to learn how to project images 
into an agnostic visual space, that can be further used for domain alignment in the feature space. Our architecture, \adage, achieves impressive results on several benchmarks, outperforming the current state of the art by a significant margin. Future work will further explore alternative architectural choices for performing domain alignment in the feature space, will expand the experimental evaluation to include object-based domain adaptation benchmarks like Office, and will extend \adage to the open set multi source domain adaptation and generalization scenarios.  

% \section{Domain Generalization}
% \label{sec:da_domain_generalization}
% Jigsaw

\chapter{Learning to see across modalities}
\label{chap:auxiliary}
\textit{This chapter presents methods which can be used even when classical domain adaption approaches are not applicable; the most common requirement for the use of DA methods is that source and target must share the same classes. Another, more implicit, requirement is that both domains are usually assumed to exist in the same modality. This is not true in many settings. As a real life case study, commonly encountered in robotics, we will focus on RGB-D recognition.}

\section{Synthetic Data: DepthNet}
\label{sec:depthnet}

\textit{
This section presents a study on the use of synthetic data as a proxy for real data, specifically in the context of RGB-D perception. The key intuition behind this approach is that it is possible to gather a much smaller amount of labeled $3D$ data and exploit it to generate a much larger dataset of 2D images.
}
\newline

Convolutional Neural Networks (CNNs) trained on large scale RGB databases have become the secret sauce in the majority of recent approaches for object categorization from RGB-D data.
Thanks to colorization techniques, these methods exploit the filters learned from 2D images to extract meaningful representations in $2.5D$. Still, the perceptual signature of these two kind of images is very different, with the first usually strongly  characterized by textures, and the second mostly by silhouettes of objects. Ideally, one would like to have two CNNs, one for RGB and one for depth, each trained on a suitable data collection, able to capture the perceptual properties of each channel for the task at hand. This has not been possible so far, due to the lack of a suitable depth database. This section addresses this issue, proposing to opt for synthetically generated images rather than collecting by hand a $2.5D$ large scale database. While being clearly a proxy for real data, synthetic images allow to trade quality for quantity, making it possible to generate a virtually infinite amount of data. We show that the filters learned from such data collection, using the very same architecture typically used on visual data, learns very different filters, resulting in depth features (a)  able to better characterize the different facets of depth images, and (b) complementary with respect to those derived from CNNs pre-trained on 2D datasets. Experiments on two publicly available databases show the power of our approach

\subsection{Context: RGB-D data}
Deep learning has changed the research landscape in visual object recognition over the last few years. Since their spectacular success in recognizing $1,000$ object categories~\cite{krizhevsky2012imagenet}, convolutional neural networks have become the new off the shelf state of the art in visual classification. The robot vision community has also attempted to take advantage of the deep learning trend, as the ability of robots to understand what they see reliably is critical for their deployment in the wild. A critical issue when trying to transfer results from computer to robot vision is that robot perception is tightly coupled with robot action. Hence, pure RGB visual recognition is not enough.  

The heavy use of $2.5D$ depth sensors on robot platforms has generated a lively research activity on $2.5D$ object recognition from depth maps~\cite{lai2011large,socher2012convolutional,cheng2015convolutional}. Here a strong emerging trend is that of using Convolutional Neural Networks (CNNs) pre-trained over ImageNet~\cite{russakovsky2015imagenet} by colorizing the depth channel~\cite{schwarz2015rgb}.  The approach has proved successful, especially when coupled with fine tuning~\cite{eitel2015multimodal} and/or spatial pooling strategies~\cite{zaki2016convolutional,cheng2014semi,cheng2015semi} (for a review of recent work we refer to section ~\ref{sec:robo_related}). 
These results suggest that the filters learned by CNNs from ImageNet are able to capture information also from depth images, regardless of their perceptual difference. 

Is this the best we can do? What if  one would train from scratch a CNN over a very large scale $2.5D$ object categorization database, wouldn't the filters learned be more suitable for object recognition from depth images? RGB images are perceptually very rich, with generally a strong presence of textured patterns, especially in ImageNet. Features learned from RGB data are most likely focusing on those aspects, while depth images contain more information about the shape and the silhouette of objects.
Unfortunately, as of today a $2.5D$ object categorization database large enough to train a CNN on it does not exist. A likely reason for this is that gathering such data collection is a daunting challenge: capturing the same variability of ImageNet over the same number of object categories would require the coordination of very many laboratories, over an extended period of time.

\begin{figure*}[!htb]
\centering
\includegraphics[trim={2cm 11.5cm 2cm 11.5cm},clip, width=\textwidth]{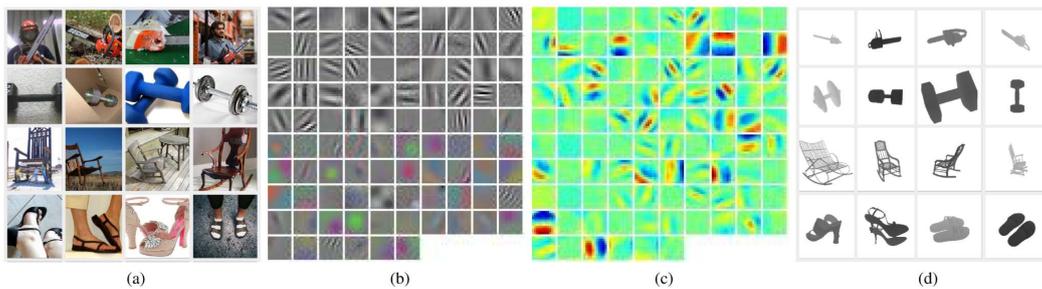}
\caption{Sample images for the classes chainsaw, dumbbell, rocker chair and sandal from ImageNet (a) and VANDAL (d). We show the corresponding filters learned by the very same CNN architecture respectively in (b) and (c) (note that this is colorized for easier viewing). We see that even though the architecture is the same, using $2D$ rather than $2.5D$ images for training leads to learning quite different filters. In (c) some of the features appear to be undefined.}
\label{fig:depthnet_first_layer_weights}
\end{figure*}

In this chapter we show a different approach: rather than acquiring a $2.5D$ object categorization database, we propose to use synthetic data as a proxy for training a deep learning architecture specialized in learning depth specific features. To this end, we constructed the VANDAL database, a collection of $4.5$ million depth images from more than $9,000$ objects, belonging to $319$ categories. The depth images are generated starting from $3D$ CAD models, downloaded from the Web, through a protocol developed to extract the maximum information from the models. VANDAL is used as input to train from scratch a deep learning architecture, obtaining a pre-trained model able to act as a depth specific feature extractor. Visualizations  of the filters learned by the first layer of the architecture show that the filter we obtain are indeed very different from those learned from ImageNet with the very same convolutional neural network (fig. ~\ref{fig:depthnet_first_layer_weights}). As such, they are able to capture different facets of the perceptual information available from real depth images, more suitable for the recognition task in that domain. We call our pre-trained architecture DepthNet. 

Experimental results on two publicly available databases confirm this: when using only depth, our DepthNet features achieve better performance compared to previous methods based on a CNN pre-trained over ImageNet, without using fine tuning or spatial pooling. The combination of the DepthNet features with the descriptors obtained from the CNN pre-trained over ImageNet, on both depth and RGB images, leads to strong results on the Washington database~\cite{lai2011large}, and to results competitive with fine-tuning and/or sophisticated spatial pooling approaches on the JHUIT database~\cite{li2015beyond}. To the best of our knowledge, this is the first work that uses synthetically generated depth data to train a depth-specific convolutional neural network.
All the VANDAL data, the protocol and the software for generating new depth images, as well as the pre-trained DepthNet, is publicly available: \href{https://sites.google.com/site/vandaldepthnet/}{https://sites.google.com/site/vandaldepthnet/}.

The rest of the section is organized as follows. First, we introduce the VANDAL database, describing its generation protocol and showcasing the obtained depth images (subsection ~\ref{depthnet_db}). Subsection ~\ref{depthnet_net} describes the deep architecture used and subsection ~\ref{depthnet_expers} reports our experimental findings. This section concludes with a summary and a discussion on future research.

\subsection{The VANDAL database}
\label{depthnet_db}
Here we present VANDAL and the protocol followed for its creation. With $4.5M$ synthetic images, it is the largest existing depth database for object recognition. Subsection ~\ref{depthnet_db-1} describes the criteria used to select the object categories composing the database and the protocol followed to obtain the $3D$ CAD models from Web resources. Subsection ~\ref{depthnet_db-2} illustrates the procedure used to generate depth images from the $3D$ CAD models. 

\begin{figure}[!htb]
\centering
 \includegraphics[width=0.7\textwidth]{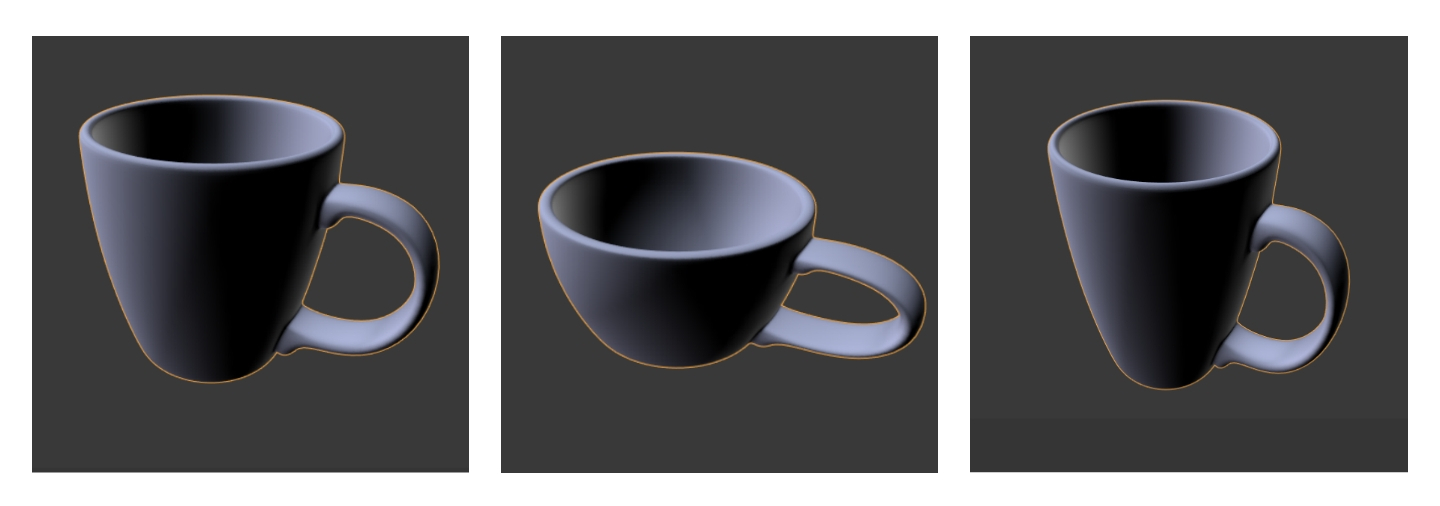}
\caption{Sample morphs (center, right) generated from an instance model for the category coffee cup (left). }
\label{fig:depthnet_morphs}
\end{figure}

\begin{figure*}[!htb]
\centering
\includegraphics[trim={2cm 11cm 2cm 11cm},clip, width=\textwidth]{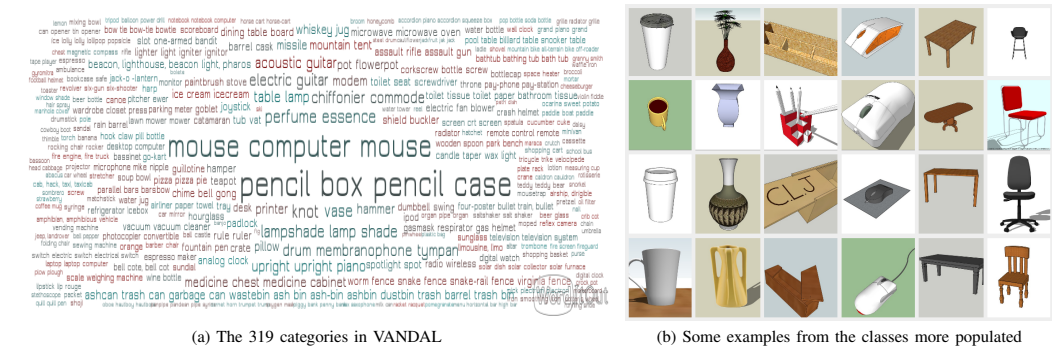}
\caption{The VANDAL database. On the left, we show a word cloud visualization of the classes in it, based on the numbers of $3D$ models in each category. On the right, we show exemplar models for the six categories more populated: coffee cup, vase, pencil case, computer mouse, table and chair.}
\label{fig:depthnet_Vandal}
\end{figure*}

\subsubsection{Selecting and Generating the $3D$ Models} 
\label{depthnet_db-1}
CNNs trained on ImageNet have been shown to generalize well when used on other object centric datasets. Following this reasoning, we defined a list of object categories as a subset of the ILSVRC2014 list~\cite{russakovsky2015imagenet}, removing by hand all scenery classes, as well as objects without a clear default shape such as clothing items or animals. This resulted in a first list of roughly 480 categories, which was used to query public $3D$ CAD model repositories like $3D$ Warehouse, Yeggi, Archive3D, and many others.
 
Five volunteers\footnote{Graduate students from the MARR program at DIAG, Sapienza Rome University.} manually downloaded the models, removing all irrelevant items like floor or other supporting surfaces, people standing next to the object and so forth, and running a script to harmonize the size of all models (some of them were originally over 1GB per file). They were also required to create significantly morphed variations of the original $3D$ CAD models, whenever suitable. 
Fig. ~\ref{fig:depthnet_morphs} shows examples of morphed models for the object category coffee cup. 
Finally, we removed all categories with less than two models, ending up with $319$ object categories with an average of 30 models per category, for a total of $9,383$ CAD object models. Fig. ~\ref{fig:depthnet_Vandal}, left, gives a world cloud visualization of the VANDAL dataset, while on the right it shows examples of $3D$ models for the 6 most populated object categories.

\subsubsection{From $3D$ Models to $2.5$ Depth Images} %Here how we generate the depth images, also 
%here we need a figure.
\label{depthnet_db-2}
All depth renderings were created using Blender\footnote{\texttt{www.blender.org}}, with a python script fully automating the procedure, and then saved as 8bit grayscale .png files, using the convention that black is close and white is far. 

The depth data generation protocol was designed to extract as much information as possible from the available $3D$ CAD models. This concretely means obtaining the greatest possible variability between each rendering. The approach commonly used by real RGB-D datasets consists in fixing the camera at a given angle and then using a turntable to get all possible viewpoints of the object~\cite{lai2011large,li2015beyond}. We tested here a similar approach, but we found out using perceptual hashing\footnote{\texttt{http://www.phash.org/}} that a significant number of object categories had more than 50\% nearly identical images.

\begin{figure}[!b]
\centering
\includegraphics[width=0.43\textwidth,height=0.26\textheight]{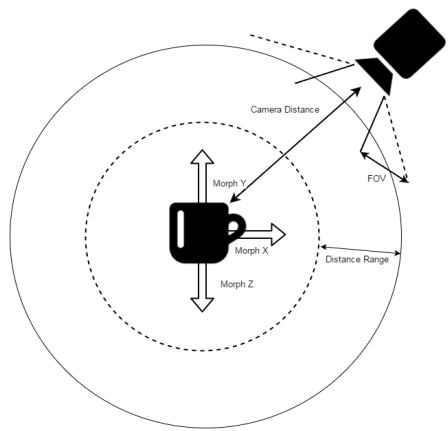}
\caption{Configuration space used for generating renderings in the VANDAL database.}
\label{fig:depthnet_rendering}
\end{figure}

We defined instead a configuration space consisting of: 
(a) object distance from the camera,
(b) focal length of the camera,
(c) camera position on the sphere defined by the distance, and
 (d) slight (\(< 10\%\)) random morphs along the axes of the model.
 Figure ~\ref{fig:depthnet_rendering}  illustrates the described configuration space. This protocol 
ensured that almost none of the resulting images were 
identical. Object distance and focal length were constrained to make sure the object always appears in a recognizable way.
%the same to each other.
We sampled this configuration space with roughly $480$ depth images for each model, obtaining a total of $4.5$ million images. Preliminary experiments showed that increasing the sampling rate in the configuration space did lead to growing percentages of nearly identical images.
%(preliminary experiments showed diminishing returns from this point on). In the end, more than 4 million images were created. 

\begin{figure}[!tb]
\centering
 \includegraphics[width=0.8\textwidth]{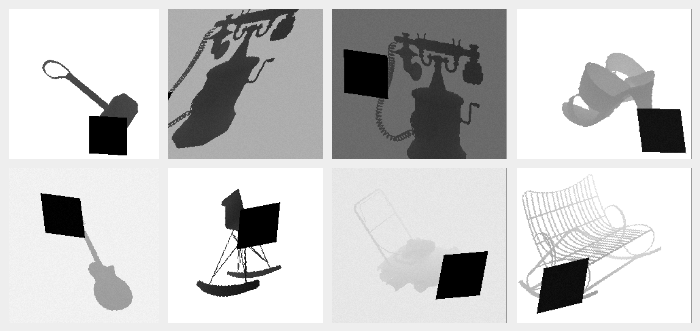}
\caption{Data augmentation samples from various classes (hammer, phone, sandal, guitar, rocker, lawn mower, bench). Note that the contrast brightness variations and noise are hard to visualize on small thumbnails.}
\label{fig:depthnet_data_aug}
\end{figure}

The rendered depth images consist of objects always centered on a white background, as it allows us the maximum freedom to perform various types of data augmentation at training time, which is standard practice when training convolutional neural networks. This is here even more relevant than usual, as synthetically generated data are intrinsically perceptually less informative. 
%As can be seen from sample images in figure ~\ref{fig:data_aug}(a), all rendering are extremely clean and the object is always centered on a white background. This is completely intentional as it allows us the maximum freedom to perform various types of data augmentation at training time. While synthetic data is generally less informative than real data, by also leveraging data augmentation methods we can offset this, up to a certain point.
The data augmentation methods we used are: image cropping, occlusion (1/4 of the image is randomly occluded to simulate gaps in the sensor scan), contrast/brightness variations (which allows us to train our network to deal with both raw and normalized data), in depth views corresponding to scaling the Z axis and shifting the objects along it, background substitution (substituting the white background with one randomly chosen farther away than the object's center of mass),  random uniform noise (as in film grain), and image shearing (a slanting transform). While not all of these data augmentation procedures produce a realistic result, they all contribute as regularizers; we avoided modeling a more complex sensor model noise for efficiency reasons. Figure ~\ref{fig:depthnet_data_aug} shows some examples of data augmentation images obtained with this protocol.\footnote{More in depth information can be found at the project's website: \texttt{https://sites.google.com/site/vandaldepthnet/}}
%\end{itemize}

%Practically speaking, the depth renderings were created in Blender, using a python script to fully automate the procedure, and then saved as grayscale PNGs, with the convention that black is close and white is far.
%The principles behind the procedure were that we:
%\begin{enumerate}
%\item \label{1} wanted to exploit our CAD models as much as possible
%\item \label{2} wanted to be able to easily do advanced data augmentation in the network itself 
%\end{enumerate}
%To do ~\ref{1}), we needed the greatest possible variance between the renderings. The approach commonly used by real RGB-D datasets consists in fixing the camera at a given angles and then using a turntable to get all possible viewpoints of the object; we understand the need for such a procedure for real data, but most of the resulting depth images are identical if the object is symmetric (a bottle, for example).

%We tested a similar approach but found out, using a perceptual hashing technique, that some object categories had more than 50\% nearly identical images.
%Instead, we defined a configuration space consisting of:
%\begin{itemize}

\subsection{Learning Deep Depth Filters}
\label{depthnet_net}
Once the VANDAL database has been generated, it is possible to use it to train any kind of convolutional deep architecture. In order to allow for a fair comparison with previous work, we opted for CaffeNet, a slight variation of AlexNet~\cite{krizhevsky2012imagenet}. Although more modern networks have been proposed in the last years~\cite{simonyan2014very,ioffe2015batch,he2016deep}, it still represents the most popular choice among practitioners, and the most used in robot vision\footnote{Preliminary experiments using the VGG, Inception and Wide Residual networks on the VANDAL database did not give stable results and require further investigation.}.  Its well know architecture consists of 5 convolutional layers, interwoven with pooling, normalization and ReLU layers, plus three fully connected layers. \footnote{CaffeNet differs from AlexNet in the pooling, which is done there before normalization. It usually performs slightly better and has thus gained wide popularity.}

Although the standard choice in robot vision is using the output of the seventh activation layer as feature descriptors, several studies in the vision community show that lower layers, like the sixth and the fifth, tend to have higher generalization properties~\cite{zheng2016good}. We followed this trend, and opted for the fifth layer (by vectorization) as deep depth feature descriptor
(an ablation study supporting this choice is reported in subsection ~\ref{depthnet_expers}). We name in the following as \textbf{DepthNet} the CaffeNet architecture trained on VANDAL using as output feature the fifth layer, and \textbf{Caffe-ImageNet} the same architecture trained over ImageNet.

\begin{figure*}[!t]
\centering
%\subfloat {%
%\includegraphics[width=1.0\textwidth]{depthnet_images/ICRA2017-Overview}
%}

\includegraphics[width=1.0\textwidth]{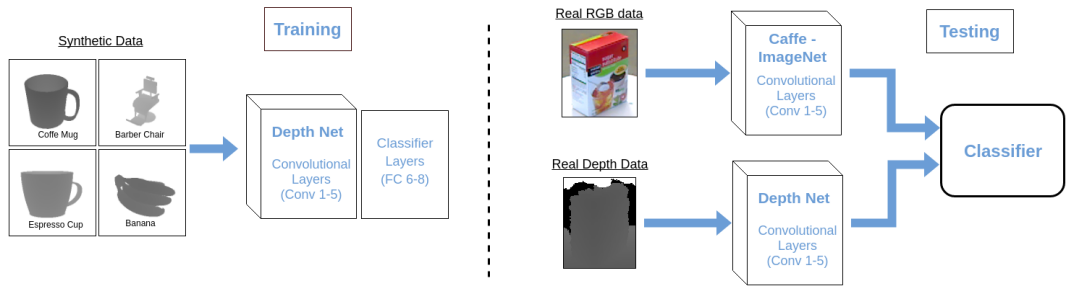}

\caption{DepthNet and our associated RGB-D object classification framework. During training, we learn depth filters from the VANDAL synthetic data (left). During test (right), real RGB and depth data is processed by two distinct CNNs, each specialized over the corresponding modality. The features, derived from the activations of the fifth convolutional layer, are then fed into a cue integration classifier.}
\label{fig:depthnet_architecture}
\end{figure*}

Once DepthNet has been trained, it can be used as any depth feature descriptor, alone or in conjunction with Caffe-ImageNet for classification of RGB images. We explore this last option, proposing a system for RGB-D object categorization that combines the two feature representations through a multi kernel learning classifier~\cite{orabona2010online}. Figure ~\ref{fig:depthnet_architecture} gives an overview of the overall RGB-D classification system. Note that DepthNet can be combined with any other RGB and/or $3D$ point cloud descriptor, and that the integration of the modal representations can be achieved through any other cue integration approach. This underlines the versatility of DepthNet, as opposed to recent work where the depth component was tightly integrated within the proposed overall framework, and as such unusable outside of it~\cite{eitel2015multimodal,zaki2016convolutional,cheng2015convolutional,li2015beyond}. 

\subsection{Experiments}
\label{depthnet_expers}
We assessed the DepthNet, as well as the associated RGB-D framework of fig. ~\ref{fig:depthnet_architecture}, on two publicly available databases. Subsection ~\ref{depthnet_setup} describes our experimental setup and the databases used in our experiments. Subsection ~\ref{depthnet_ablation} reports a set of experiments assessing the performance of DepthNet on depth images, compared to Caffe-ImageNet, while in subsection ~\ref{depthnet_rgb-d} we assess the performance of the whole RGB-D framework with respect to previous approaches.

\subsubsection{Experimental setup}
\label{depthnet_setup}
We conducted experiments on the Washington RGB-D~\cite{lai2011large} and the JHUIT-50~\cite{li2015beyond}  object datasets. The first consists of $41,877$ RGB-D images organized into $300$ instances divided in $51$ classes. Each object instance was positioned on a turntable and captured from three different viewpoints while rotating. Since two consecutive views are extremely similar, only 1 frame out of 5 is used for evaluation purposes. We performed experiments on the object categorization setting, where we followed the evaluation protocol defined in~\cite{lai2011large}. The second is a challenging recent dataset that focuses on the problem of fine-grained recognition. It contains $50$ object instances, often very similar with each other (e.g. 9 different kinds of screwdrivers). As such, it presents different classification challenges compared to the Washington database. 

All experiments, as well as the training of DepthNet, were done using the publicly available Caffe framework~\cite{jia2014caffe}. As described above, we obtained DepthNet by training a CaffeNet over the VANDAL database. The network was trained using Stochastic Gradient Descent for 50 epochs. Learning rate started at 0.01 and gamma at 0.5 (halving the learning rate at each step). We used a variable step down policy, where the first step took 25 epochs, the next 25/2, the third 25/4 epochs and so on. These parameters were chosen to make sure that the test loss on the VANDAL test data had stabilized at each learning rate. Weight decay and momentum were left at their standard values of 0.0005 and 0.9. Training and test data was centered by removing the mean pixel. 
%We also evaluated the performances of the network at each epoch on a validation set of the real datasets.

%the parameters of all layers were updated using a fixed learning rate schedule, i.e. we used an initial learning rate of xxxx, reduced to xxxx after xxxx iterations. Training stopped after xxxx.

To assess the quality of the DepthNet features we performed three set of experiments:
\begin{enumerate}
\item \emph{Object classification using depth only:}  features were extracted with DepthNet and a linear SVM\footnote{Liblinear: http://www.csie.ntu.edu.tw/~cjlin/liblinear/} was trained on it. We also examined how the performance varies when extracting from different layers of the network, comparing against a Caffe-ImageNet used for depth classification, as in~\cite{schwarz2015rgb}.
\item \emph{Object classification using RGB + Depth:}  in this setting we combined our depth features with those extracted from the RGB images using  Caffe-ImageNet. While~\cite{eitel2015multimodal} train a fusion network to do this, we simply use an off the shelf Multi Kernel Learning (MKL) classifier~\cite{orabona2010online}. %\footnote{Obscure Multi Kernel Learning - see DOGMA at http://francesco.orabona.com/software.html}.
%\item Object classification using our Depth + Imagenet pretrained Depth: we propose this experiment to see if the features learned on different data also highlight different qualities. This was also performed with MKL.
\end{enumerate} 
%\subsubsection{Implementation details}
For all experiments we used the training/testing splits originally proposed for each given dataset. For linear SVM, we set $C$ by cross validation. When using MKL, we left the default values of $100$ iterations for online and $300$ for batch and set $p$ and $C$ by cross validation.

Previous works using  Caffe-ImageNet as feature extractor for depth, apply some kind of input preproccessing~\cite{schwarz2015rgb,eitel2015multimodal,zaki2016convolutional}. While we do compare against the published baselines, we also found that by simply normalizing each image (min to 0 and max to 255), one achieves very competitive results.
%already gives the best performances that can be obtained from a network trained on RGB data. 
Also, since our DepthNet is trained on depth data, it does not need any type of preprocessing over the depth images, obtaining strong results over raw data.
%performs very well (better actually) with the raw data.
Because of this, in all experiments reported in the following we only consider raw depth images and normalized depth images.

\subsection{Assessing the performance of the DepthNet architecture}
\label{depthnet_ablation}

\begin{figure}
\centering
\includegraphics[width=0.9\textwidth]{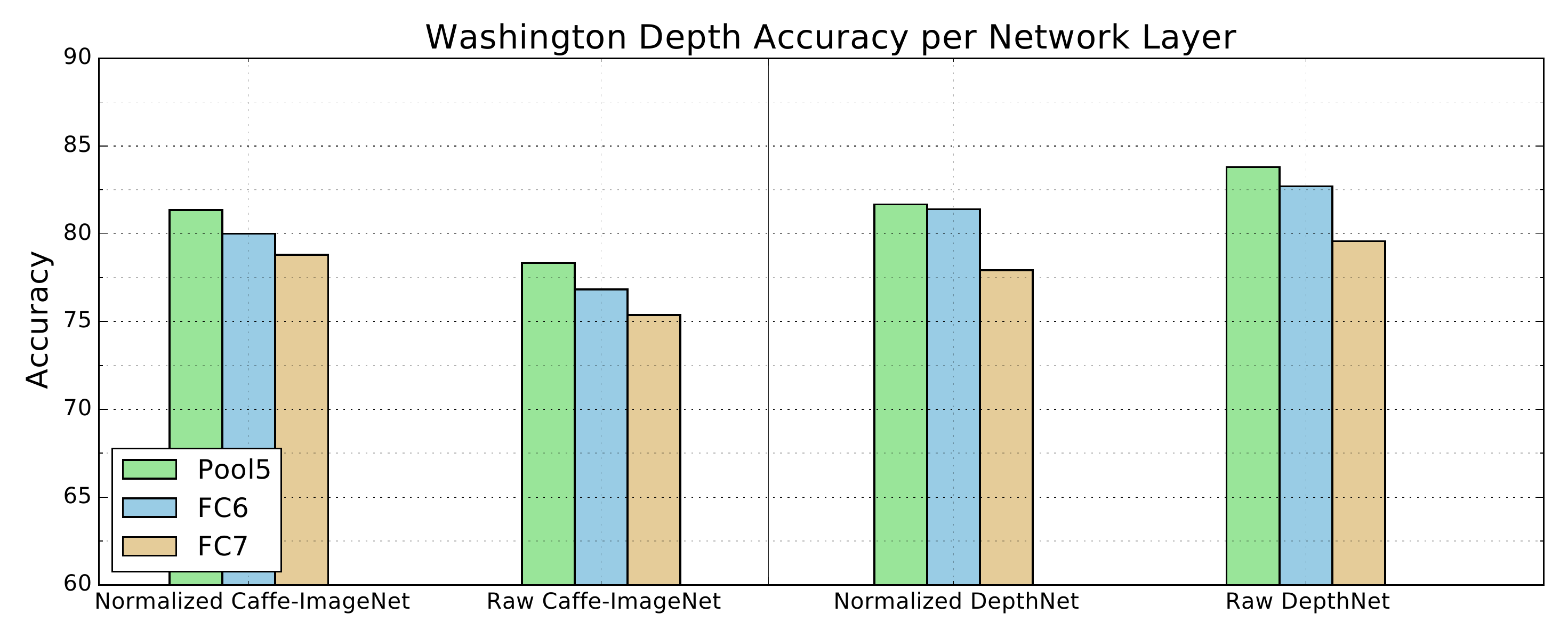}
\caption{Accuracy obtained by DepthNet and Caffe-ImageNet over the Washington database, using as features pool5, FC6 and FC7. Results are reported for raw and normalized depth images.}
\label{fig:depthnet_washington_ablation}
\end{figure}

We present here an ablation study, aiming at understanding the impact of choosing features from the last fully convolutional layer as opposed to the more popular last fully connected layer, and of using normalized depth images instead of raw data. By comparing our results with those obtained by  Caffe-ImageNet, we also aim at illustrating up to which point the features learned from VANDAL are different from those derived from ImageNet.
\begin{figure*}[!tb]
\centering
 \includegraphics[width=1.0\textwidth,height=0.23\textheight]{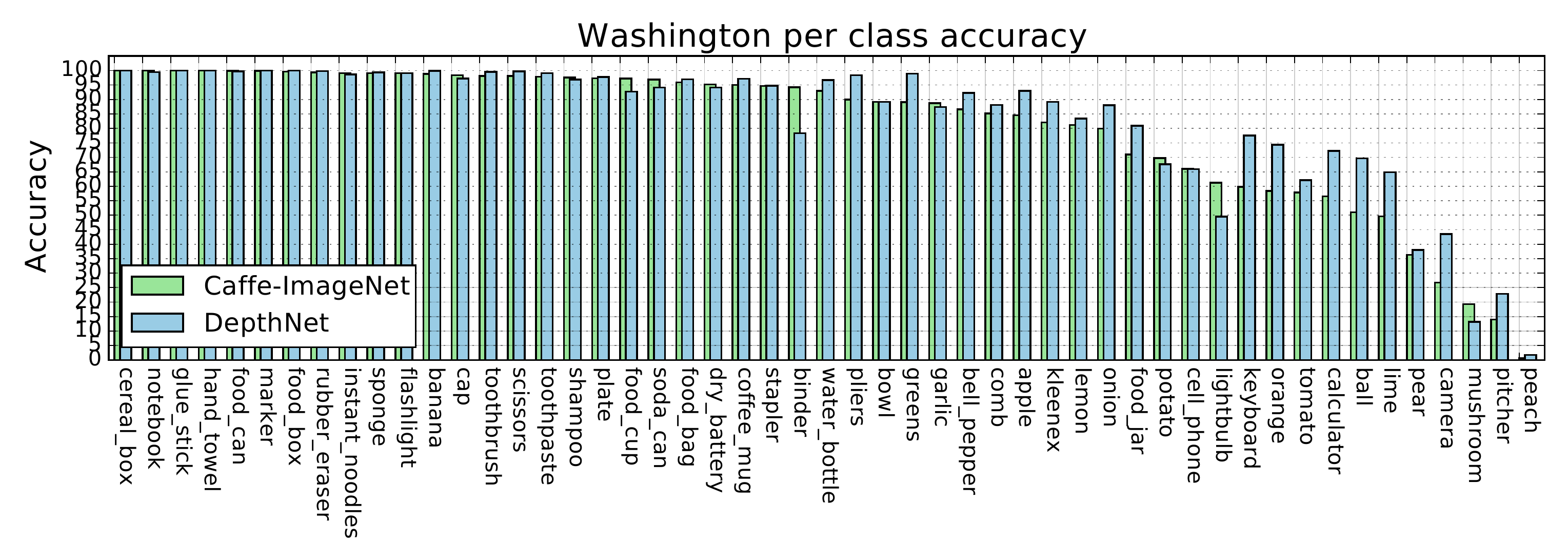}
\caption{Accuracy per class on the Washington dataset, depth images. Classes sorted by the  Caffe-ImageNet accuracies.}
\label{fig:depthnet_washington_acc}
\end{figure*}

Figure ~\ref{fig:depthnet_washington_ablation} shows results obtained on the Washington database, with normalized and raw depth data, using as features the activations of the fifth pooling layer (pool5), of the sixth fully connected layer (FC6), and of the seventh fully connected layer (FC7). Note that this last set of activations is the standard choice in the literature. We see that for all settings, pool5 achieves the best performance, followed by FC6 and FC7. This seems to confirm recent findings on RGB data~\cite{zheng2016good}, indicating that pool5 activations offer stronger generalization capabilities when used as features, compared to the more popular FC7. 

The best performance is obtained by DepthNet, pool5 activations over raw depth data, with a $83.8\%$ accuracy. DepthNet achieves also better results compared to Caffe-ImageNet over normalized data. To get a better feeling of how performance varies when using DepthNet or Caffe-ImageNet, we plotted the per-class accuracies obtained using pool5 and raw depth data. We sorted them in descending order according to the Caffe-ImageNet scores (fig. ~\ref{fig:depthnet_washington_acc}). 

\begin{figure*}[!htb]
\centering
\includegraphics[trim={3.5cm 8cm 3.5cm 7.3cm},clip, width=0.95\textwidth]{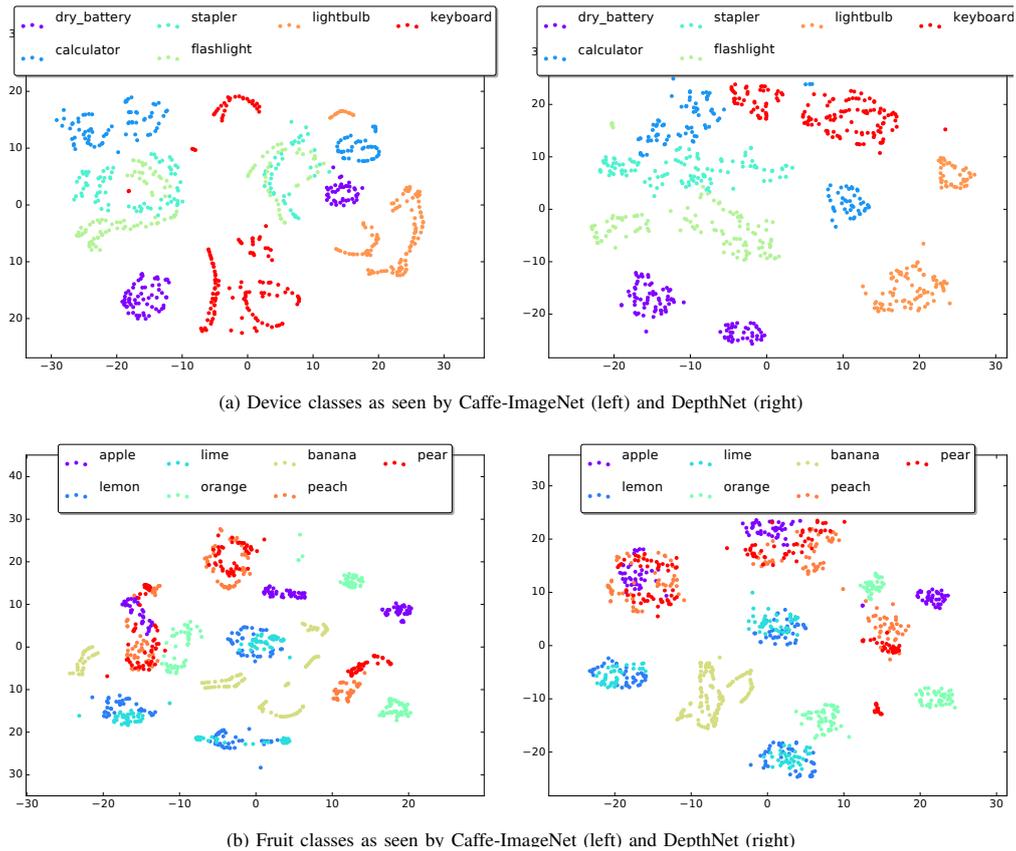}
\caption{t-SNE visualizations for the categories device (top) and fruit (bottom).}
\label{fig:depthnet_features}
\end{figure*}

While there seems to be a bulk of objects where both features perform well (left), DepthNet seems to have an advantage over challenging objects like apple, onion, ball, lime and orange (right), where the round shape tends to be more informative than the specific object texture. This trend is confirmed also when performing a t-SNE visualization~\cite{maaten2008visualizing} of all the Washington classes belonging to the high-level categories 'fruit' and 'device' (fig.~\ref{fig:depthnet_features}). We see that in general the DepthNet features tend to cluster tightly the single categories while at the same time separating them very well. For some classes like dry battery and banana, the difference between the two representations is very marked.
This does not imply that DepthNet features are always better than those computed by Caffe-ImageNet. Fig.~\ref{fig:depthnet_washington_acc} shows that CaffeNet features obtain a significantly better performance compared to DepthNet over the classes binder and mushroom, to name a few. 

The features learned by the two networks seem to focus on different perceptual aspects of the images. This is likely due to the different set of samples used during training, and the consequent different learned filters (fig.~\ref{fig:depthnet_first_layer_weights}).

From these figures we can draw the following conclusions: (a) DepthNet provides the overall stronger descriptor for depth images, regardless of the activation layer chosen and the presence or not of preprocessing on the input depth data; (b) the features derived by the two networks tend to capture different features of the data, and as such are complementary. As we will show in the next subsection, this last point leads to very strong results when combining the two with a principled cue integration algorithm.

\subsubsection{Assessing the performance of the RGB-D architecture}
\label{depthnet_rgb-d}

In this subsection we present experiments on RGB-D data, from both the Washington and JHUIT databases, assessing the performance of our DepthNet-based framework of fig. ~\ref{fig:depthnet_architecture} against previous approaches. Table ~\ref{depthnet_desp:washington} shows in the top row our results, followed by results obtained by Caffe-ImageNet using the pool5 activations as features, as well as results from the recent literature based on convolutional neural networks. 

First, we see that the results in the RGB column stresses once more the strength of the pool5 activations as features: they achieve the best performance without any form of fine tuning, spatial pooling or sophisticated non-linear learning, as done instead in other approaches~\cite{eitel2015multimodal,zaki2016convolutional,cheng2015convolutional}. 
Second, DepthNet on raw depth data achieves the best performance among CNN-based approaches with or without fine tuning like~\cite{schwarz2015rgb,eitel2015multimodal}, but it is surpassed by approaches encoding explicitly spatial information through pooling strategies, and/or by using a more advanced classifier than a linear SVM, as we did. We would like to stress that we did not incorporate any of those strategies in our framework on purpose, to better assess the sheer power of training a given convolutional architecture on perceptually different databases. Still, nothing prevents in future work the merging of DepthNet with the best practices in spatial pooling and non-linear classifiers, with a very probable further increase in performance.

Lastly, we see that in spite of the lack of such powerful tools, our framework achieves the best performance on RGB-D data. This clearly underlines that the representations learned by DepthNet are both powerful and able to extract different nuances from the data than Caffe-ImageNet.  Rather than the actual overall accuracy reported here in the table, we believe this is the breakthrough result we offer to the community in this work.

\begin{table}[!htb]
\begin{adjustbox}{width=1.0\textwidth,center=\textwidth}
$\begin{array}{|l|c|c|c|c|}
\hline
\text{Method:}                                      & \text{RGB}          & \text{Depth Mapping}  & \text{Depth Raw}    & \text{RGB-D}        \\
\hline
\textbf{DepthNet RGB-D Framework}                                           & \mathbf{88.49 \pm 1.8}           & 81.68 \pm 2.2       & \mathbf{83.8 \pm 2.0}  & \mathbf{92.25 \pm 1.3} \\
\hline
\hline
\text{Caffe-ImageNet Pool5}                    & \mathbf{88.49 \pm 1.8} & 81.11 \pm 2  & 78.35 \pm 2.5 & 90.79 \pm 1.2 \\
\text{Caffe-ImageNet FC7 finetuning~\cite{eitel2015multimodal}} & 84.1 \pm 2.7  & 83.8 \pm 2.7 & -            & 91.3 \pm 1.4  \\
\text{Caffe-ImageNet FC7~\cite{schwarz2015rgb}}              & 83.1 \pm 2.0  & -           & -            & 89.4 \pm 1.3 \\
\text{CNN only~\cite{cheng2015convolutional}}                     &     82.7 \pm 1.2 &  78.1 \pm 1.3 & - & 87.5 \pm 1.1 \\
\text{CNN + FisherKernel + SPM~\cite{cheng2015convolutional}}     &     86.8 \pm 2.2 &  \mathbf{85.8 \pm 2.3} & - & 91.2 \pm 1.5 \\
\text{CNN + Hypercube Pyramid + EM~\cite{zaki2016convolutional}}  & 87.6  \pm 2.2 &  85.0 \pm 2.1 & - & 91.4 \pm 1.4 \\
\text{CNN-SPM-RNN+CT~\cite{cheng2015semi}}       &     85.2 \pm 1.2 &  83.6 \pm 2.3 & - & 90.7 \pm 1.1 \\
\text{CNN-RNN+CT~\cite{cheng2014semi}}         &	  81.8 \pm 1.9 &  77.7 \pm 1.4 & - & 87.2 \pm 1.1 \\
\text{CNN-RNN~\cite{socher2012convolutional}}                                  &     80.8 \pm 4.2 &  78.9 \pm 3.8 & - & 86.8 \pm 3.3 \\
\hline
\end{array}$
\end{adjustbox}
\caption{Comparison of our DepthNet framework with previous work on the Washington database. With \textit{depth mapping} we mean all types of depth preprocessing used in the literature.}
\label{depthnet_desp:washington}
\end{table}

Experiments over the JHUIT database confirms the findings obtained over the Washington collection (table ~\ref{table:depthnet_jhuit-results}). Here our RGB-D framework obtains the second best result, with the state of the art achieved by the proposers of the database with a non CNN-based approach. Note that this database focuses over the fine-grained classification problem, as opposed to object categorization as explored in the experiments above. While the results reported in Table ~\ref{table:depthnet_jhuit-results} on Caffe-ImageNet using FC7 seem to indicate that the choice of using pool5 remains valid, the explicit encoding of local information is very important for this kind of tasks~\cite{zhang2012pose,angelova2014benchmarking}. We are inclined to attribute to this the superior performance of~\cite{li2015beyond}; future work incorporating spatial pooling in our framework, as well as further experiments on the object identification task in the Washington database and on other RGB-D data collections will explore this issue. 

\subsection{Conclusions}
\label{depthnet_conclu}
In this section we focused on object classification from depth images using convolutional neural networks. We argued that, as effective as the filters learned from ImageNet are, the perceptual features of $2.5D$ images are different, and that it would be desirable to have deep architectures able to capture them. 

To this purpose, we created VANDAL, the first depth image database synthetically generated, and we showed experimentally that the features derived from such data, using the very same CaffeNet architecture widely used over ImageNet, are stronger while at the same time complementary to them. This result, together with the public release of the database, the trained architecture and the protocol for generating new depth synthetic images, is the contribution of this work.
\begin{table}[!tb]
\centering
$\begin{array}{|l|c|c|c|c|}
\hline
\text{Method:}                                      & \text{RGB}          & \text{Depth Mapp.}  & \text{Depth Raw}    & \text{RGB-D}        \\
\hline
\textbf{DepthNet Pool5}                                           & -            & \mathbf{54.37}       & \mathbf{55.0}  & \mathbf{90.3} \\
\text{Caffe-ImageNet Pool5}                      & 88.05 & 53.6  & 38.9 & 89.6 \\
\text{Caffe-ImageNet FC7~\cite{schwarz2015rgb}}         &  82.08 & 47.87 & 26.11 & 83.6 \\
\hline
\text{CSHOT + Color pooling} & & & & \\
\text{+ MultiScale Filters~\cite{li2015beyond}} &  - & - & - & \mathbf{91.2} \\
\text{HMP~\cite{li2015beyond}}                                          & 81.4 & 41.1 & - & 74.6 \\
\hline
\end{array}$
\caption{Comparison of our DepthNet framework with previous work on the JHUIT database.
As only one split is defined, we do not report std.}
\label{table:depthnet_jhuit-results}
\end{table}

We see this work as the very beginning of a long research thread. By its very nature, DepthNet could be plugged into all previous work using CNNs pre-trained over ImageNet for extracting depth features. It might substitute that module, or it might complement it; the open issue is when this will prove beneficial in terms of spatial pooling approaches, learning methods and classification problems. A second issue we plan to investigate is the impact of the deep architecture over the filters learned from VANDAL. 

While in this work we chose on purpose to not deviate from CaffeNet, it is not clear that this architecture, which was heavily optimized over ImageNet, is able to exploit at best our synthetic depth database. While preliminary investigations with existing architectures have not been satisfactory, we believe that architecture surgery might lead to better results. Furthermore, including the $3D$ models from~\cite{wu20153d} might greatly enhance the generality of the DepthNet.

Finally, we believe that the possibility to use synthetic data as a proxy for real images opens up a wide array of possibilities: for instance, given prior knowledge about the classification task of interest, would it be possible to generate on the fly a task specific synthetic database, containing the object categories of interest under very similar imaging conditions, and train and end-to-end deep network on it? How would performance change compared to the use of network activations as done today? Future work will focus on these issues.

\section{Transfer Learning across modalities: \texorpdfstring{$DE^2CO$}{DE2CO}}
\label{sec:deco}
\textit{
While Domain Adaptation methods perform extremely well, they have certain limitations; the main one being the need for source and the target to share the same categories. There have been some works attempting to~\cite{busto2017open} work around this issue, but they still assumes some kind of semantic overlap. On the other hand, when there are sufficient labeled samples from the target, one can usually perform transfer learning via finetuning~\cite{yosinski2014transferable}.
The limitation to this approach is that we assume that the source and target exist in the same modality (i.e. RGB images).
In this section we will focus on a method which allows us to perform transfer learning across modalities, with no assumption being made on the class labels.}
\newline

The ability to classify objects is fundamental for robots. Besides knowledge about their visual appearance, captured by the RGB channel, robots heavily need also depth information to make sense of the world. While the use of deep networks on RGB robot images has benefited from the plethora of results obtained on databases like ImageNet, using convnets on depth images requires mapping them into three dimensional channels. This transfer learning procedure makes them processable by pre-trained deep architectures. Current mappings are based on heuristic assumptions over pre-processing steps and on what depth properties  should be most preserved, resulting often in cumbersome data visualizations, and in sub-optimal performance in terms of generality and recognition results. 

Here we take an alternative route and we attempt instead to \emph{learn}  an optimal colorization mapping for any given pre-trained architecture, using as training data a reference RGB-D database. We propose a deep network architecture, exploiting the residual  paradigm,  that learns how to map depth data to three channel images. %from a reference database. 
%, to be used as input to  pre-trained deep networks. 
A qualitative analysis of the images obtained with this approach clearly indicates that learning the optimal mapping  preserves the richness of depth information better than
%leads to images  capturing far better the richness of depth information, as opposed to 
current hand-crafted approaches. 
%currently in use. 

Experiments on the Washington, JHUIT-50 and BigBIRD  public benchmark databases, using CaffeNet, VGG-16, GoogleNet, and ResNet50  clearly showcase the power of our approach, with gains in  performance of up to $16\%$ compared to state of the art competitors on the depth channel only, leading to top performances when dealing with RGB-D data.    

\subsection{Motivation}
%%%%%%%%%%%%%%%%%%%%%%%%%%%%%%%%%%%%%%%%%%%%%%%%%
%%%%%%%%%%%%%%%%% FIRST CHAPTER %%%%%%%%
%%%%%%%%%%%%%%%%%%%%%%%%%%%%%%%%%%%%%%%%%%%%%%%%%%
Robots need to recognize what they see around them to be able to act and interact with it. Recognition must be carried out in the RGB domain, capturing  mostly the visual appearance of things related to their reflectance properties, as well as in the depth domain, providing information about the shape and silhouette of objects and supporting both recognition and interaction with items.  The current mainstream state of the art approaches for object recognition are based on Convolutional Neural Networks (CNNs,~\cite{NIPS1989_293-CNN}), which use end-to-end architectures achieving feature learning and classification at the same time. Some notable advantages of these networks are their ability to reach much higher accuracies on basically any visual recognition problem, compared to what would be achievable with heuristic methods; their being domain-independent, and their conceptual simplicity. Despite these advantages, they also present some limitations, such as high computational cost, long training time and the demand for large datasets, among others.

This last issue has so far proved crucial in the attempts to leverage over the spectacular success of CNNs over RGB-based object categorization~\cite{googlenet,krizhevsky2012imagenet} in the depth domain. Being CNNs data-hungry algorithms, the availability of very large scale annotated data collections is crucial for their success, and architectures trained over ImageNet~\cite{deng2009imagenet} are the cornerstone of the vast majority of CNN-based recognition methods. Besides the notable exception of~\cite{carlucci2016deep}, the mainstream approach for using CNNs on depth-based object classification has been through transfer learning, in the form of a mapping able to  make the depth input channel compatible with the data distribution 
expected by RGB architectures.

Following recent efforts in transfer learning~\cite{Transfer1,Transfer2,Transfer3} that made it possible to use depth data with CNN pre-trained on a database of a different modality, several authors proposed hand-crafted mappings to colorize depth data, obtaining impressive improvements in classification % over shallow features 
over the Washington~\cite{washington} database, that has become the golden reference benchmark in this field~\cite{schwarz2015rgb,eitel2015multimodal}.
%used AlexNet~\cite{krizhevsky2012imagenet} pre-trained on ImageNet to perform classification on depth data.

We argue that this strategy is sub-optimal. By hand-crafting the mapping for the depth data colorization, one has to make strong assumptions on what information, and up to which extent, should be preserved in the transfer learning towards the RGB modality. %across the depth and RGB modalities. 
While some choices might be valid for some classes of problems and settings, it is questionable whether the family of algorithms based on this approach can provide results  combining high recognition accuracies with robustness across different settings and databases. Inspired by recent works on colorization of gray-scale photographs~\cite{IizukaSIGGRAPH2016,larsson2016learning,cheng2015deep}, we tackle the problem by exploiting the power of end-to-end convolutional networks, proposing  a deep depth colorization architecture able to learn the optimal  transfer learning from depth to RGB   for any given pre-trained convnet. 

\begin{figure}[t]
\centering
\includegraphics[width=0.9\textwidth]{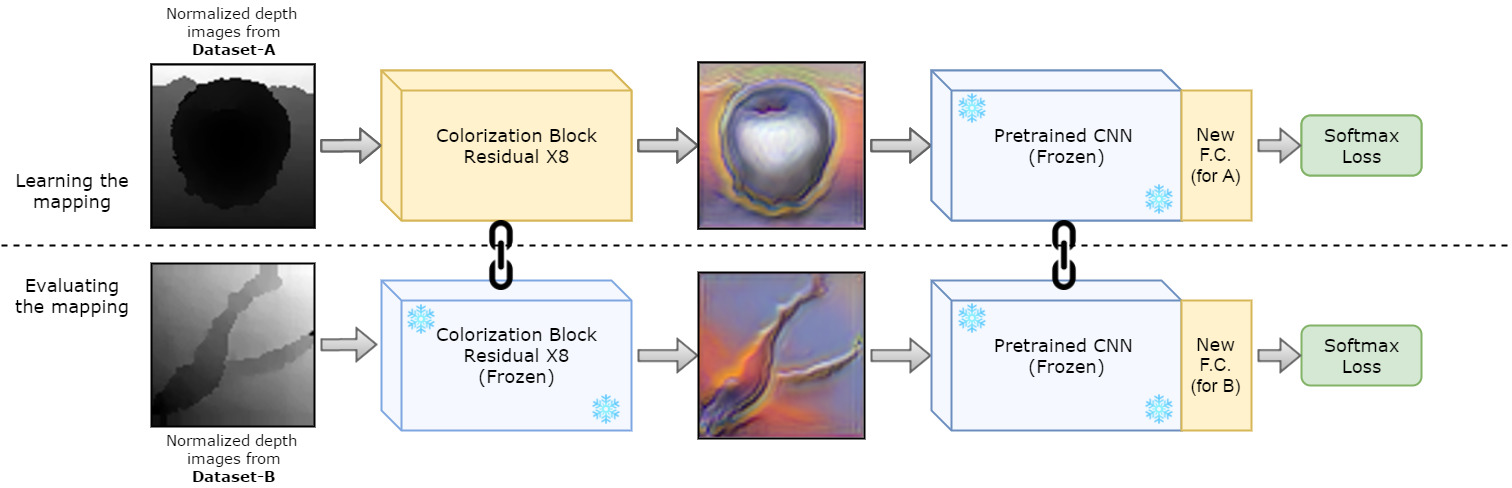}
\caption{The \DECO{} pipeline consists of two phases. First, we learn the mapping, from depth to color, maximizing the discrimination capabilities of a network pre trained on ImageNet. In this step the network is frozen and we are only learning the mapping and the final layer. 
%(which must be relearned, as the classes are different). 
We then evaluate the colorization on a \emph{different} depth dataset: here we also freeze the colorization network and only train a new final layer for the testbed dataset.}
\label{fig:deco_arch}
\end{figure}

Our deep colorization network takes advantage of the residual approach~\cite{ResNet}, learning how to map between the two modalities by leveraging over a reference database (Figure~\ref{fig:deco_arch}, top), for any given architecture. After this training stage, the colorization network can be added on top of its reference pre-trained architecture, for any object classification task  (Figure~\ref{fig:deco_arch}, bottom).
We call our network \DECO: DEep DEpth COlorization. 

We assess the performance of \DECO\ in several ways. A first qualitative analysis, comparing the colorized depth images obtained by \DECO\  and by other state of the art hand-crafted approaches, gives intuitive insights on the advantages brought by learning the mapping as opposed to choosing it, over several databases. We further deepen this analysis with an experimental evaluation of our and other existing transfer learning methods on the depth channel only, using four different deep architectures and three different public databases, with and without fine-tuning. Finally, we tackle the RGB-D object recognition problem, combining  \DECO\  with off-the shelf state of the art RGB deep networks, benchmarking it against the current state of the art in the field. For all these experiments, results clearly support the value of our algorithm. 

All the \DECO\  modules, for all architectures employed, are available at \url{https://github.com/fmcarlucci/de2co}.

% METHOD
\subsection{Colorization of Depth Images}
\label{colorization}
Although depth and RGB are modalities with significant differences,
%two different modalities with significant differences,
they also share enough similarities (edges, gradients, shapes)
to make it plausible that convolutional filters learned from RGB data could be re-used effectively for representing colorized depth images. The approach currently adopted in the literature consists of designing ad-hoc colorization algorithms, as revised in section~\ref{sec:robo_related}. We refer to these kind of approaches as \emph{hand-crafted depth colorization}. Specifically, we choose ColorJet 
~\cite{eitel2015multimodal}, SurfaceNormals~\cite{Bo} and \textit{SurfaceNormals++}  
~\cite{aakerberg2017depth} as baselines against which we will assess our data driven approach because of their popularity and effectiveness.

In the rest of the section we first briefly summarize ColorJet  (subsection~\ref{shallow-depth}), SurfaceNormals and SurfaceNormals++  (subsection~\ref{shallow-depth-surfacenormals}). We then describe our deep approach to depth colorization (subsection~\ref{de2co}). To the best of our knowledge, \DECO\ is the first deep colorization architecture applied successfully to depth images.

\subsubsection{Hand-Crafted Depth Colorization: ColorJet}
\label{shallow-depth}
ColorJet works by assigning different colors to different depth values. The original depth map is firstly normalized between 0-255 values. Then the colorization works by  mapping the lowest value to the blue channel and the highest value to the red channel. The value in the middle is mapped to green and the intermediate  values are arranged accordingly~\cite{eitel2015multimodal}. The resulting image exploits the full RGB spectrum, with the intent of leveraging at best the filters learned by deep networks trained on very large scale RGB datasets like ImageNet. Although simple, the approach gave very strong results when tested on the Washington database, and when deployed on a robot platform. 

Still, ColorJet was not designed to create realistic looking RGB images for the objects depicted in the original depth data (Figure~\ref{fig:deco_multiple_mappings}, bottom row). This  raises the question whether this mapping, although more effective than other methods presented in the literature, might be sub-optimal. In subsection~\ref{de2co}  we will show that by fully embracing the end-to-end philosophy at the core of deep learning, it is indeed possible to achieve significantly higher recognition performances while at the same time producing more realistic colorized images.   

\subsubsection{Hand-Crafted Depth Colorization: SurfaceNormals(++)}
\label{shallow-depth-surfacenormals}
The SurfaceNormals mapping has been often used to convert depth images to RGB~\cite{Bo,wang2016correlated,eitel2015multimodal}. The process is straightforward:  for each pixel in the original image the corresponding surface normal is computed as a normalized $3D$ vector, which is then treated as an RGB color. Due to the inherent noisiness of the depth channel, such a direct conversion results in noisy images in the color space. To address this issue, the mapping we call \textit{SurfaceNormals++} was introduced by Aakerberg~\cite{aakerberg2017depth}: first, a recursive median filter is used to reconstruct missing depth values, subsequently a bilateral filter smooths the image to reduce noise, while preserving edges.
Next, surface normals are computed for each pixel in the depth image. Finally the image is sharpened using the unsharp mask filter, to increase contrast around edges and other
high-frequency components.

\subsubsection{Deep Depth Colorization: \texorpdfstring{(DE)$^2$CO}{DE2CO}}
\label{de2co}
\DECO{}
%, our deep depth colorization method, 
consists of feeding the depth maps, normalized into grayscale images, to a \textit{colorization network} linked to a standard CNN architecture, pre-trained on ImageNet.

\begin{figure}[ht]
\centering
\includegraphics[width=0.4\textwidth]{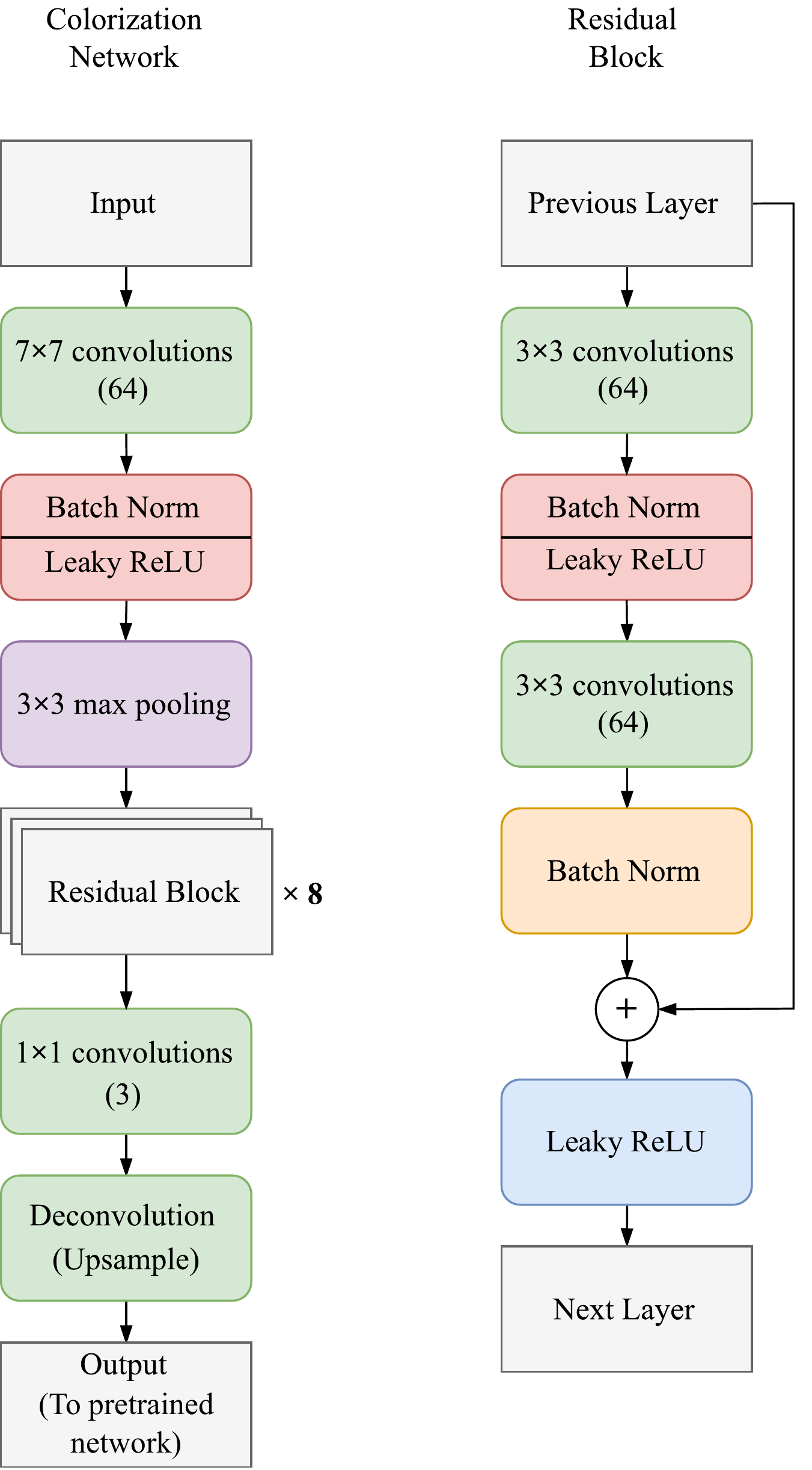}
\caption{Overview of the  \DECO\ colorization network. On the left, we show the overall architecture; on the right, we show details of the residual block.}
\label{fig:deco_rescol}
\end{figure}

Given the success of deep colorization networks from grayscale images, we first tested   existing architectures in this context~\cite{zhang2016colorful}. Extensive experiments showed that while the visual appearance of the colorized images was very good, the recognition performances obtained when combining such network with pre-trained RGB architectures was not competitive. Inspired by the generator network in~\cite{bousmalis2016unsupervised}, we propose here a \textit{residual} convolutional architecture (Figure~\ref{fig:deco_rescol}). 
By design~\cite{ResNet}, this architecture is robust and allows for deeper training. This is helpful here, as \DECO\ requires stacking together two networks, which, even for not very deep architectures, might lead to vanishing gradient issues.
% Fabio: il paragrafo che segue è nuovo
%One good reason for this choice is that, once you stack two networks, you risk (depending on how deep the pretrained is) running into vanishing gradient issues. By design~\cite{ResNet}, this architecture is robust and allows for deeper training. 
Furthermore, residual blocks works at pixel level~\cite{bousmalis2016unsupervised} helping to preserve locality.

Our architecture works as follows: the $1x228x228$ input depth map, reduced to $64x57x57$ size by a conv\&pool layer, passes through a sequence of $8$ residual blocks, composed by two small convolutions with batch normalization layers and leakyRelu~\cite{maas2013rectifier} as non linearities. 
The last residual block output is convolved by a three features convolution to form the 3 channels image output. Its resolution is brought back to 228x228 by a \textit{deconvolution} (upsampling) layer. %We call this new depth colorization architecture (DE)$^2$CO. 
%The colorization network full structure can be seen at figure~\ref{fig:deco_rescol}

Our whole system for object recognition in the depth domain using  deep networks pre-trained over RGB images can be summarized as follows:
%The colorization learning technique can be so summarized: 
the entire network, composed by  \DECO\  and the classification network of choice, is trained on an annotated reference  depth image dataset.  The weights of the chosen classification network are kept frozen in their pre-trained state, as the only layer that needs to be retrained is the last fully connected layer connected to the softmax layer.  Meanwhile, the weights of   \DECO\  are updated until convergence.

\begin{figure}[ht]
\centering
\includegraphics[trim=0 0 0 0,clip,width=0.95\textwidth]{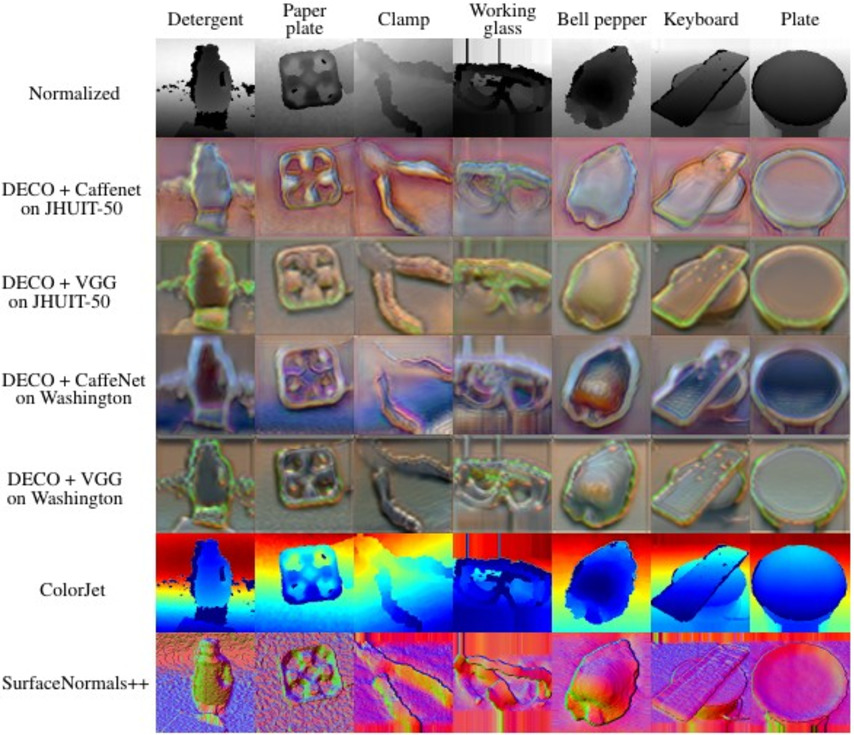}
\caption{ {(DE)$^2$CO} colorizations applied on different objects, taken from~\cite{jhuit,washington,singh2014bigbird}. Top row shows the depth maps mapped to grayscale. From the second to the fourth row, we show the corresponding {(DE)$^2$CO} colorizations learned on different settings. Fifth row shows \textit{ColorJet} views~\cite{eitel2015multimodal}, while the last row shows the surface normals mapping.~\cite{aakerberg2017depth} \textit{SurfaceNormals++}. These images showcase \DECO{}'s ability 
to emphasize the object's shape and to capture its salient features. % with respect to a shallow color mapping. 
}
\label{fig:deco_multiple_mappings}
\end{figure}

After this step, the depth colorization network has learned the mapping that maximizes the classification accuracy on the reference training dataset. 
It can now be used to colorize \emph{any} depth image, from any data collection. 
Figure~\ref{fig:deco_multiple_mappings}, top rows, shows exemplar images colorized with \DECO\ trained over different reference databases, in combination with two different architectures (CaffeNet, an implementation variant of AlexNet, and VGG-16~\cite{VGG}). 

We see that, compared to the images obtained with ColorJet and SurfaceNormal++, our colorization technique emphasizes the objects contours and their salient features while flatting the object background, while the other methods introduce either high frequency noise (SurfaceNormals++) or emphasize background gradient instead of focusing mainly on the objects (ColorJet).    %We see that, compared to the images obtained with ColorJet, our results are more nuanced and expressive, especially when the training is conducted on databases containing  varied and perceptually rich visual samples  as is the case for the Washington database.
In the next subsection we will show how this qualitative advantage translates also into a numerical advantage, i.e.  how learning \DECO\ on one dataset and performing depth-based object recognition on another
leads to a very significant increase in performance on several settings, compared to hand-crafted color mappings.

% -- Figure -- 

%%%%%%%%%%%%%%%%%%%%%%
%%% EXPERIMENTS: %%%%%
%%%%%%%%%%%%%%%%%%%%%%

\subsection{Experiments}
\label{deco_experiments}
% mettere cose comuni qui
\label{deco_setup}
We evaluated our colorization scheme on three main settings: an ablation study of how different depth mappings perform when the network weights are kept frozen (subsection~\ref{deco_ablation}), a comparison of depth performance with network finetuning (subsection~\ref{deco_fine-tuning}) and finally an assessment of \DECO\ when used in RGB-D object recognition tasks (subsection~\ref{deco_rgb-d}). Before reporting on our findings, we illustrate the baselines we used  (subsection~\ref{deco_exper-setup}). The datasets we use (Washington~\cite{washington}, JHUIT-50~\cite{jhuit} and BigBIRD~\cite{singh2014bigbird}) are described in section~\ref{sec:rgbd_datasets}.

\subsubsection{Experimental Setup}
\label{deco_exper-setup}

\textbf{Hand-crafted Mappings}
According to previous works~\cite{eitel2015multimodal,aakerberg2017depth}, the two most effective mappings are \textit{ColorJet}~\cite{eitel2015multimodal} and \textit{SurfaceNormals}~\cite{Bo,aakerberg2017depth}.
For ColorJet we normalized the data between $0$ and $255$ and then applied the mapping using the OpenCV libraries\footnote{"COLORMAP\_JET" from \textit{http://opencv.org/}}. 
%Using this procedure we mapped red to the farthest away point, and blue to the closest. We also experimented with doing the opposite  but did not observe any benefits. 
For the SurfaceNormals mapping we considered two versions: the straightforward conversion of the depthmap to surface normals~\cite{Bo} and the enhanced version \textit{SurfaceNormals++}~\cite{aakerberg2017depth} which uses extensive pre-processing and generally performs better\footnote{The authors graciously gave us their code for our experiments.}.

\subsubsection{Ablation Study}
\label{deco_ablation}
In this setting we compared our \DECO\ method against hand crafted mappings, using pre-trained networks as feature extractors and only retraining the last classification layer.
We did this on the three datasets described in section~\ref{sec:rgbd_datasets},  over four architectures:
 CaffeNet (a slight variant of the AlexNet~\cite{krizhevsky2012imagenet}), VGG16~\cite{VGG} and GoogleNet~\cite{googlenet} were chosen because of their popularity within the robot vision community.  We also tested the recent ResNet50~\cite{ResNet}, which %  and the very efficient SqueezeNet v$1.1$~\cite{SqueezeNet}
although not currently very used in the robotics domain, has some promising properties.
%it presents different advantages in terms of compactness and performance that promise to make it soon popular in this domain.
In all cases we considered models pretrained on ImageNet~\cite{deng2009imagenet}, which we retrieved from Caffe's \textit{Model Zoo}\footnote{\textit{https://github.com/BVLC/caffe/wiki/Model-Zoo}}.

%\noindent
%\textbf{Learning \DECO\ }
Training \DECO\ means minimizing the multinomial logistic loss of a network trained on RGB images. This means that our  network is attached between the depth images and the pre-trained network, of which we freeze the weights of all but the last layer, which are relearned from scratch (see Figure~\ref{fig:deco_arch}).
We trained each network-dataset combination for $50$ epochs using the Nesterov solver~\cite{nesterov1983method} and $0.007$ starting learning rate (which is stepped down after $45\%$). During this phase, we  used the whole  \textit{source} datasets, leaving aside only $10\%$ of the samples for validation purposes.

%\subsubsection{Assessing the colorization strategies (ablation)}
%Once \DECO\ has finished learning on dataset \textit{A}, we tested its mapping on dataset \textit{B} (which it had never seen before), by only training the new final layer (and freezing all the rest) for the classification task with softmax.
When the dataset on which we train the colorizer is different from the test one, we simply retrain the new final layer (freezing all the rest) for the new classes.

Effectively we are using the pre-trained networks as feature extractors, as done in~\cite{schwarz2015rgb,eitel2015multimodal,zaki2016convolutional} and many others; for a performance analysis in the case of network finetuning we refer to paragraph~\ref{deco_fine-tuning}.  In this setting we used the Nesterov (for Washington and JHUIT-50) and ADAM (for BigBIRD) solvers. As we were only training the last fully connected layer, we learned  a small handful of parameters with a very low risk of overfitting.

%Upon acceptance of this paper we will release full code for training and evaluation procedures.

\begin{figure}
\centering
\includegraphics[width=0.95\textwidth]{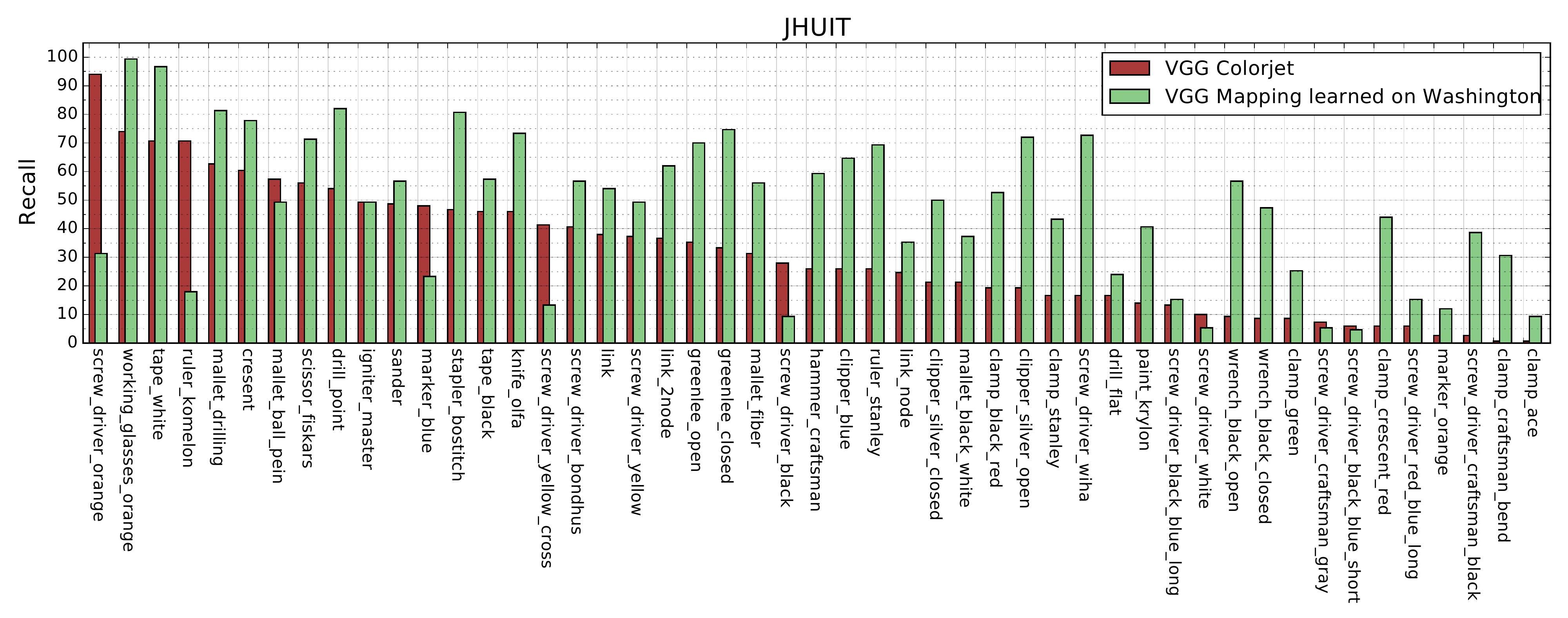}
\caption{%Ablation experiments: 
Per class recall on JHUIT-50, using VGG, with \DECO\ learned from Washington. Recalls per class are sorted in decreasing order, according to ColorJet performance. In this setting, % we see that 
\DECO{}, while generally performing better, seems to focus on different perceptual properties and is thus, compared with the baseline, better at some classes rather than others. }
\label{fig:deco_jhuit_recall}
\end{figure}

Table~\ref{table:deco_comparison} reports the results from the ablation
%obtained over the three different databases, for the five different architectures we selected, while Figure~\ref{fig:washington_recall} and 
while Figure~\ref{fig:deco_jhuit_recall} focuses on the class recall for a specific experiment.
%on a couple of those experiments. 
For every architecture, we report the results obtained using ColorJet, SurfaceNormals (plain and enhanced) and \DECO\ learned on a reference database between Washington or JHUIT-50, and \DECO\ learned on the combination of Washington and JHUIT-50. For the CaffeNet and VGG networks we also present results on simple grayscale images.
We attempted also to learn \DECO\ from BigBIRD alone, and in combination with one (or both) of the other two databases. Results on BigBIRD only were disappointing, and results with/without adding it to the other two databases did not change the overall performance. We interpret this result as caused by the relatively small variability of objects in BigBIRD with respect to depth, and for sake of readability we decided to omit  them in this work.

%For every \DECO\ mapping, we report the recognition accuracy only when the colorization was learned from a database different from the one used as testbed for recognition. This is to ensure a fair comparison with hand mapping methods, as with this protocol all methods use training data from the testbed database only when learning the final classification layer.

We see that, for all architectures and for all reference databases, \DECO\ achieves higher results. The difference goes from $+1.7\%$, obtained with CaffeNet on the Washington database, to the best of $+16.8\%$ for VGG16 on JHUIT-50.  JHUIT-50 is the testbed database where, regardless of the chosen architecture, \DECO\ achieves the strongest gains in performance compared to hand crafted mappings. Washington is, for all architectures, the database where hand crafted mappings perform best, with the combination Washington to CaffeNet being the most favorable to the shallow mapping. 

On average it appears the CaffeNet is the architecture that performs best on this datasets; still, it should be noted that we are using here all architectures as feature extractors rather than as classifiers. On this type of tasks, both ResNet and GoogLeNet-like networks are known to perform worse than CaffeNet~\cite{azizpour2016factors}, hence our results are consistent with what reported in the literature.  
In Table~\ref{table:DECO_ablation} we report a second ablation study performed on the width and depth of \DECO{} architecture. Starting from the standard \DECO{} made of 8 residual blocks with 64 filters for each convolutional layer (which we found to be the best all-around architecture), we perform additional experiments by doubling and halving the number of residual blocks and the number of filters in each convolutional layer.
As it can be seen, the \DECO{} architecture is quite robust but can be potentially finetuned to each target dataset to further increase performance.

In table~\ref{table:DECO_speed} we report runtimes for the considered networks. As the results show, while \DECO{} requires some extra computation time, in real life this is actually offset  by the fact that only $\frac{1}{3}$ of the data is being moved to the GPU.
\begin{table}
\centering
$\begin{array}{|l|c|}
\hline
\text{Network} & \text{Time (ms)} \\
\hline
\text{CaffeNet} & 695 \\
\hline
\text{VGG} &  1335 \\
\hline
\text{GoogleNet} &  1610 \\
\hline
\text{ResNet-50} &  1078\\
\hline
\text{\DECO{} colorizer} & 400 \\
\hline
\end{array}$
$\begin{array}{|l|c|}
\hline
\text{Network} & \text{Time (s)} \\
\hline
\text{CaffeNet} & 1.87 \\
\hline
\text{\DECO{} + CaffeNet} &  1.23 \\
\hline
\text{VGG} &  2.91 \\
\hline
\text{\DECO{} + VGG} &  2.16\\
\hline
\end{array}$
\caption{Left: forward-backward time for 50 iterations, as by \textit{caffe time}. Right:  feature extraction times for 100 images; note that using \DECO{} actually speeds up the procedure. 
We explain this by noting that \DECO{} uses single channel images and thus needs to transfer only $\frac{1}{3}$ of the data from memory to the GPU - clearly the bottleneck here. }
\label{table:DECO_speed}
\end{table}

\begin{table}[!htb]
\begin{adjustbox}{width=1.0\textwidth,center=\textwidth}
$\begin{array}{|l|c|c|c|}
\hline
\text{Method:}                                      & \text{Washington~\cite{washington}}          & \text{JHUIT-50~\cite{jhuit}}    & \text{BigBIRD Reduced~\cite{singh2014bigbird}}        \\
\hline
\text{VGG16 on Grayscale}	& 74.9 & 33.7 & 22 \\
\text{VGG16 on ColorJet}	& 75.2 & 35.3 & 19.9 \\
\text{VGG16 on SurfaceNormals}	& 75.3 & 30.8 & 16.8 \\
\text{VGG16 on SurfaceNormals++}	& 77.3 & 35.8 & 11.5 \\
\text{VGG16 \DECO{} learned on Washington} & \mathbf{79.6} & \mathbf{52.7} & 22.8 \\
\text{VGG16 \DECO{} learned on JHUIT-50} 	& 78.1	& 51.2 & 23.7 \\
%\text{VGG16 \DECO{} learned on Washington + JHUIT-50} 	& 78.8	& 47.2	& \mathbf{24.0}\\
\hline
\text{CaffeNet on Grayscale}	& 76.6 & 44.6 & 22.9 \\
\text{CaffeNet on ColorJet}	& 78.8 & 45.0 & 22.7 \\
\text{CaffeNet on SurfaceNormals}	& 79.3 & 38.3 & 18.9 \\
\text{CaffeNet on SurfaceNormals++}	& 81.4 & 44.8 & 14.0 \\
\text{CaffeNet \DECO{} learned on Washington} & \textcolor{red}{\mathbf{83.1}} & 53.1 & \textcolor{red}{\mathbf{28.6}} \\
\text{CaffeNet \DECO{} learned on JHUIT-50} 	& 79.1	& \textcolor{red}{\mathbf{57.5}} & 25.2 \\
%\text{CaffeNet \DECO{} learned on Washington + JHUIT-50} 	& 81.2	& 55.8	& \textcolor{red}{\mathbf{29.3}}\\
\hline
\hline
\text{GoogleNet on ColorJet}	& 73.5 & 40.0 & 21.8 \\
\text{GoogleNet on SurfaceNormals}	& 72.9 & 36.5 & 18.4 \\
\text{GoogleNet on SurfaceNormals++}	&\mathbf{76.7} & 41.5 & 13.9 \\
\text{GoogleNet \DECO{} learned on Washington} & - & \mathbf{51.9} & \mathbf{25.2} \\
\text{GoogleNet \DECO{} learned on JHUIT-50} 	& 76.6 & - & 24.4 \\
%\text{GoogleNet \DECO{} learned on Washington + JHUIT-50} & - & -	& \mathbf{28.6} \\
\hline
\text{ResNet50 on ColorJet}	& 75.1 & 38.9 & 18.7 \\
\text{ResNet50 on SurfaceNormals}	& 77.4 & 33.2 & 16.5 \\
\text{ResNet50 on SurfaceNormals++}	& \mathbf{79.6} & 45.4 & 13.8 \\
\text{ResNet50 \DECO{} learned on Washington} & - & \mathbf{45.5} & 23.9 \\
\text{ResNet50 \DECO{} learned on JHUIT-50} 	& 76.4 & - & \mathbf{24.7}\\
%\text{ResNet50 \DECO{} learned on Washington + JHUIT-50} & - & -	& 23.7 \\
\hline

\end{array}$
\end{adjustbox}
\caption{Object classification experiments in the depth domain,  comparing \DECO{} and hand crafted mappings, using $5$ pre-trained networks 
%on ImageNet
as feature extractors. Best results for each network-dataset combination are in \textbf{bold}, overall best in \textcolor{red}{\textbf{red bold}}. Extensive experiments were performed on VGG and Caffenet, while GoogleNet and ResNet act as reference.}
\label{table:deco_comparison}
\end{table}

\begin{table}[!htb]
\centering
$\begin{array}{|l|c|c|c|}
\hline
\text{filters/blocks } & \text{4 blocks} & \text{8 blocks}   & \text{
  16 blocks}  \\
\hline
\text{32 filters} & 56.5 & 52.8 & 57.1  \\
\hline
\text{64 filters } & 56.8 & 53.1 & 53.6  \\
\hline
\text{128 filters } & 53.1 & 53.9 & 53.3 \\
\hline
\end{array}$
\caption{\DECO{} ablation study, learned on Washington, tested on JHUIT-50. Grid search optimization over width and depth of generator architecture shows improved results. }
\label{table:DECO_ablation}
\end{table}

\subsubsection{Finetuning}
\label{deco_fine-tuning}
In our finetuning experiments we focused on the best performing network from the ablation, the CaffeNet (which is also used by current competitors~\cite{eitel2015multimodal,aakerberg2017depth}), to see up to which degree the network could exploit a given mapping. 
The protocol was quite simple: all layers were free to move equally, the starting learning rate was $0.001$ (with step down after $45\%$) and the solver was \textit{SGD}. Training went on for $90$ epochs for the Washington and JHUIT-50 datasets and $30$ eps. for BigBIRD (a longer training was detrimental for all settings). 
To ensure a fair comparison with the static mapping methods, the \DECO{} part of the network was kept frozen during finetuning. 

Results are reported in Table~\ref{table:deco_finetuning_comparison}. We see that here the gap between hand-crafted and learned colorization methods is reduced (very likely the network is compensating existing weaknesses). 
\textit{SurfaceNormals++} performs pretty well on Washington, but less so on the other two datasets (it's actually the worse on BigBIRD). Surprisingly, the simple grayscale conversion is the one that performs best on BigBIRD, but lags clearly behind on all other settings.
\DECO{} on the other hand, performs comparably to the best mapping on every single setting \textbf{and} has a $5.9\%$ lead on JHUIT-50; we argue that it is always possible to find a shallow mapping that performs very well on a specific dataset, but there are no guarantees it can generalize.

\subsubsection{RGB-D}
\label{deco_rgb-d}
%using the same protocol used in the previous paragraph
While this section focuses on how to best perform recognition in  the depth modality using convnets, we wanted to provide a reference value for RGB-D object classification using \DECO\ on the depth channel. To classify RGB images we follow~\cite{aakerberg2017depth} and use a pretrained VGG16 which we finetuned on the target dataset (using the previously defined protocol).
RGB-D classification is then performed, without further learning, by computing the weighted average (weight factor $\alpha$ was cross-validated) of the \textit{fc8} layers from the RGB and Depth networks and simply selecting the most likely class (the one with the highest activations). This cue integration scheme can be seen as one of the simplest, off-the-shelf algorithm for doing classifications using two different modalities~\cite{TommasiOC08}. We excluded BigBIRD from this setting, due to lack of competing works to compare with. 

Results are reported in Tables~\ref{table:deco_washington_comparison}-\ref{table:deco_jhuit_comparison}.  We see that \DECO{} produces results on par or slightly superior to the current state of the art, even while using an extremely simple feature fusion method. This is remarkable, as competitors like 
~\cite{aakerberg2017depth,eitel2015multimodal}
use instead sophisticated, deep learning based cue integration methods. Hence, our ability to compete in this setting is all due to the \DECO\ colorization mapping, clearly more powerful than the other baselines.  

It is worth stressing that, in spite of the different cue integration and depth mapping approaches compared in Tables~\ref{table:deco_washington_comparison}-\ref{table:deco_jhuit_comparison}, convnet results on RGB are already very high, hence in this setting the advantage brought by a superior performance on the depth channel tends to be covered. Still, on Washington we achieve the second best result, and on JHUIT-50 we get the new state of the art.

\begin{table}[!htb]
\begin{adjustbox}{width=1.0\textwidth,center=\textwidth}
$\begin{array}{|l|c|c|c|}
\hline
\text{Method:}                                      & \text{Washington~\cite{washington}}          & \text{JHUIT-50~\cite{jhuit}}    & \text{BigBIRD Reduced~\cite{singh2014bigbird}}        \\
% \hline
% \text{VGG16 on ColorJet}	& 75.2 & 35.3 & 19.9 \\
% \text{VGG16 on SurfaceNormals}	& 75.3 & 30.8 & 16.8 \\
% \text{VGG16 on SurfaceNormals++}	& 77.3 & 35.8 & 11.5 \\
% \text{VGG16 mapping learned on Washington} & - & 52.7 & 22.8 \\
% \text{VGG16 mapping learned on JHUIT} 	& 78.1	& - & 23.7 \\
% \text{VGG16 mapping learned on Washington + JHUIT} 	& -	& -	& 24.0\\
\hline
\text{CaffeNet on Grayscale}	& 82.7 \pm 2.1 & 53.7 & \mathbf{29.6} \\
\text{CaffeNet on ColorJet}	& 83.8 \pm 2.7 & 54.1 & 25.4 \\
\text{CaffeNet on SurfaceNormals++}	& \mathbf{84.5} \pm 2.9 & 55.9 & 17.0 \\
%\text{CaffeNet Gupta~\cite{Gupta_2016_CVPR}}	& 75.6 \pm 3.6 & 50.0 & \text{did not converge} \\
\hline
\text{CaffeNet \DECO{} learned on Washington} & 84.0 \pm 2.0 & 60.0 & - \\
\text{CaffeNet \DECO{} learned on JHUIT-50} 	& 82.3 \pm 2.3	& \mathbf{62.0} & - \\
\text{CaffeNet \DECO{} learned on Washington + JHUIT-50} 	& 84.0 \pm 2.3	& 61.8	& 28.0\\
\hline
\end{array}$
\end{adjustbox}
\caption{CaffeNet finetuning using different colorization techniques.}
\label{table:deco_finetuning_comparison}
\end{table}

\begin{table}
\centering
$\begin{array}{|l|c|c|c|}
\hline
\text{Method:}                                      & \text{RGB}          & \text{Depth}    & \text{RGB-D}        \\
\hline
\text{FusionNet~\cite{eitel2015multimodal}}	& 84.1 \pm 2.7 & 83.8 \pm 2.7 & 91.3 \pm 1.4 \\
\text{CNN + Fisher~\cite{li2015hybrid}}	& \mathbf{90.8} \pm 1.6 & 81.8 \pm 2.4 & \mathbf{93.8} \pm 0.9 \\
\text{DepthNet~\cite{carlucci2016deep}} & 88.4 \pm 1.8 & 83.8 \pm 2.0 & 92.2 \pm 1.3 \\
\text{CIMDL~\cite{wang2016correlated}} 	& 87.3 \pm 1.6	& 84.2 \pm 1.7 & 92.4 \pm 1.8 \\
\text{FusionNet enhanced~\cite{aakerberg2017depth}} 	& 89.5 \pm 1.9	& \mathbf{84.5} \pm 2.9	& 93.5 \pm 1.1\\
\hline
\text{\DECO{}} 	& 89.5 \pm 1.6	& 84.0 \pm 2.3	& 93.6 \pm 0.9\\
\hline
\end{array}$
\caption{Selected results on Washington RGB-D}
\label{table:deco_washington_comparison}
\end{table}

\begin{table}[!htb]
\centering
$\begin{array}{|l|c|c|c|}
\hline
\text{Method:}                                      & \text{RGB}          & \text{Depth}    & \text{RGB-D}        \\
\hline
\text{DepthNet~\cite{carlucci2016deep}} & 88.0 & 55.0 & 90.3 \\
\text{Beyond Pooling~\cite{jhuit}} 	& -	& - & 91.2 \\
\text{FusionNet enhanced~\cite{aakerberg2017depth}} 	& \mathbf{94.7}	& 56.0	& 95.3\\
\hline
\text{\DECO{}} 	& \mathbf{94.7}	& \mathbf{61.8}	& \mathbf{95.7}\\
\hline
\end{array}$
\caption{Selected results on JHUIT-50}
\label{table:deco_jhuit_comparison}
\end{table}

\subsection{Conclusions}
This section presented a framework for learning deep colorization mappings. Our architecture follows the residual philosophy, learning how to map depth data to RGB images for a given pre-trained convolutional neural network. By using our \DECO\ algorithm, as opposed to the hand-crafted colorization mappings commonly used in the literature, we obtained a significant jump in performance over three different benchmark databases, using four different popular deep networks pre trained over ImageNet. The visualization of the obtained colorized images further confirms how our algorithm is able to capture the rich informative content and the different facets of depth data. 
All the deep depth mappings presented in this section are available at \url{https://github.com/fmcarlucci/de2co}.

Future work will further investigate the effectiveness and generality of our approach, testing it on other RGB-D classification and detection problems, with various fine-tuning strategies and on several deep networks, pre-trained over different RGB databases, and in combination with RGB convnet with more advanced multimodal fusion approaches.

\chapter{Conclusions}
\label{chap:conclusions}
\textit{This chapter summarizes the main results and contributions presented in this thesis, discusses
the open issues and sketches possible future directions of research.}

\section{Summary}
A quick overview of the current state-of-the-art categorization methods shows us that all deep learning approaches based on a large amount of samples reach impressive results on difficult datasets~\cite{krizhevsky2012imagenet,ResNet}. However, most of them provide very few guarantees when only a small amount of training samples is available, or more in general, if there is a mismatch between the training and the testing distribution~\cite{torralba2011unbiased,perronnin2010large}. %This is, surprisingly, a piece of good news: if deep learning simply worked as a black box on any task, there would be no reason to come up with the interesting approaches that are currently being developed in the field.

The purpose of this thesis has been to understand how to best transfer what we know to a new domain, where data is scarce.
We have shown that there are multiple ways of addressing this, depending on the availability of auxiliary data or pretrained models.
More specifically, we tackled the issue from a conventional point of view, as an unsupervised domain adaptation problem, in chapter \ref{chap:DA}, and proposed multiple solutions: we showed that we can reduce the domain gap by aligning the feature distribution using an ad-hoc layer (\DIAL, \ref{sec:da_feature_align}) or by using GANs to project the images in the style of the other domain (\ref{sec:da_image_align}). In section \ref{sec:da_hybrid_align} we showed a complex approach that, by aligning both features and images, reaches even better performance.

In the second part we presented methods which look at the problem from a different angle: how should we proceed when there is no \textit{source} domain to leverage? In section \ref{sec:depthnet} we use $3D$ CAD models to generate simulated depth views as a proxy for real data. A CNN is trained on this synthetic data and then evaluated on real RGB-D datasets; experiments confirm this to be a viable option.

A general purpose transfer learning method is presented in section \ref{sec:deco}. The intuition being that it is possible to learn a non linear transformation of the target image dataset that maps it as the most discriminative input for a network pretrained in a different modality. The method was successfully evaluated on the recognition task for Depth images, by mapping them to RGB; but its prerequisite-free structure means it could also be applied to other tasks (such as semantic segmentation or detection) and modalities.

In conclusion our work demonstrated that, by properly exploiting pre-existing knowledge, it is possible to train effective deep learning models even when data is scarce. For this purpose we presented a number of alternative solutions, ranging from classical domain adaptation approaches to innovative, across modalities, transfer learning techniques, each suitable for different learning conditions.

\section{Open Issues}
All the proposed algorithmic solutions for transfer learning and domain adaptation have been presented together with an analysis of their properties, discussing which is the best setting to apply them and evaluating their limits. We briefly describe in the following a few aspects of this work that might be somehow improved and remain relevant for future work.

Regarding methods which transform the input images(sections \ref{sec:da_hybrid_align}, \ref{sec:da_image_align}) a strong limitation must be noted: to date, no generative approach has successfully managed to fully capture the nuances and the complexities of real life datasets. Until a breakthrough in this field (perhaps an improvements in GANs) solves the issue, algorithms which depends on it will always be limited to working on strongly structured data, such as faces or digits.
Part of the problem is likely due to the size of these real life datasets: it is extremely hard to learn how to generate an object when you can only see 10 samples of it (\ie Office~\cite{Saenko:2010}). The solution to this might lie, as has been done for object recognition~\cite{Transfer3}, in the use of pretrained models which are finetuned for the specific task. To the best of our knowledge, to date, no one has had success with this idea on a GAN; as we use pretrained networks for classification, we should pretrain our GANs on large scale dataset and then finetune them on the smaller, target, dataset.

All of our proposed DA methods are designed to work in the unsupervised domain adaptation setting, with the assumption that no labels are available for the target. Clearly this hypothesis does not always hold and it is likely that, if target labels are available, a much better alignment between domains can be achieved. Future work should investigate how to best adapt our approaches to this semi-supervised setting.
Likewise, our DA methods could be adapted to work on the recent problem of \textit{open set}~\cite{busto2017open} domain adaptation, a more realistic scenario where only a few categories of interest are shared between source and target .

Our work on synthetic depth images, (section \ref{sec:depthnet}) could be significantly improved by trying to reduce the huge domain gap that exists between our virtual images and the real ones. Firstly, we could take into account  and accurately model sensor noise during the rendering process. Furthermore our objects are simply floating in mid-air, which is unlike anything the robot could encounter in real life. The final domain shift could be eliminated by the use of GANs (similarly to SBADA-GAN) to make the synthetic images even more lifelike.

Our $(DE)^2CO$ (section \ref{sec:deco}) framework for transfer learning has been evaluated on the task of depth based object recognition. There are no specific requirements in the algorithm which limit its applicability to this setting, and it would be very interesting to see how much we can transfer to a different task (detection or semantic segmentation, for example) or a different modality. Multispectral cameras in particular seem a very interesting candidate: they are used for many things, from automated produce inspection, to artwork analysis and mine detection, but all relative datasets are quite small in size. Potentially, much better performances could be obtained if we could leverage the wealth of data we have in the RGB modality.

\appendix
\chapter{Domain Adaptation}
\section{\DIAL}

\subsection{\DALs formulas}
\label{sec:grad}

We rewrite Eq. \ref{eqn:dal} to make sample indexes and domain dependency explicit:
\begin{equation}
  y_i = \texttt{DA}(x_i,d_i;\alpha) =
    \frac{x_i - \mu_{d_i,\alpha}}{\sqrt{\epsilon + \sigma_{d_i,\alpha}^2}}\,.
\end{equation}
Using this notation, the global batch statistics are
\begin{equation}
\begin{aligned}
  \mu &= \frac{1}{\con{n}} \sum_{i=1}^\con{n} x_i\,,
    & \sigma^2 &= \frac{1}{\con{n}} \sum_{i=1}^\con{n} (x_i - \mu)^2\,, \\
\end{aligned}
\end{equation}
while the domain-specific statistics are
\begin{equation}
\begin{aligned}
  \mu_d &= \frac{1}{\con{n}_d} \sum_{i=1}^\con{n} \mathsf{1}_{d=d_i} x_i\,, &
  \sigma_d^2 &= \frac{1}{\con{n}_d} \sum_{i=1}^\con{n} \mathsf{1}_{d=d_i} (x_i - \mu_d)^2\,, \\
\end{aligned}
\end{equation}
and the $\alpha$-mixed statistics are
\begin{equation}
\begin{aligned}
  \mu_{d,\alpha} &= \alpha \mu_d + (1 - \alpha) \mu\,, \\
  \sigma^2_{d,\alpha} &= \frac{\alpha}{\con{n}_d} \sum_{i=1}^\con{n} \mathsf{1}_{d=d_i} (x_i - \mu_{d,\alpha})^2
    + \frac{1 - \alpha}{\con{n}} \sum_{i=1}^\con{n} (x_i - \mu_{d,\alpha})^2 \\
    &= \alpha \sigma^2_d + (1 - \alpha) \sigma^2 + \alpha (1 - \alpha) (\mu - \mu_d)^2\,,
\end{aligned}
\end{equation}
where $\con{n}$ and $\con{n}_d$ are, respectively, the total number of samples in the batch and the number of samples of domain $d$ in the batch.

\subsection{Results on the SVHN -- MNIST benchmark}
\label{sec:svhn}

In this section we report the results we obtain in the SVHN~\cite{netzer2011reading} to MNIST~\cite{lecun1998gradient} transfer benchmark.
We follow the experimental protocol in~\cite{ganin2016domain}, using all SVHN images as the source domain and all MNIST images as the target domain, and compare with the following baselines:  CORAL \cite{sun2016return}; the Deep Adaptation Networks (DAN) \cite{long2015learning}; the Domain-Adversarial Neural Network (DANN) in~\cite{ganin2016domain}; the Deep Reconstruction Classification Network (DRCN) in~\cite{ghifary2016deep}; the Domain Separation Networks (DSN) in~\cite{bousmalis2016domain}; the Asymmetric Tri-training Network (ATN) in~\cite{saito2017asymmetric}.

As in all baselines, we adopt the network architecture in~\cite{ganin2014unsupervised}, adding \DALs after each layer with parameters.
Training is performed from scratch, using the same meta-parameters as for \Alex (see section \ref{subsec:dial_experiments}), with the following exceptions: initial learning rate $l_0=0.01$; 25 epochs; learning rate schedule defined by $l_p = l_0 / (1 + \gamma p)^\beta$, where $\gamma=10$, $\beta=0.75$ and $p$ is the learning progress linearly increasing from $0$ to $1$.

\begin{figure}
  \centering
  \includegraphics[width=0.5\columnwidth]{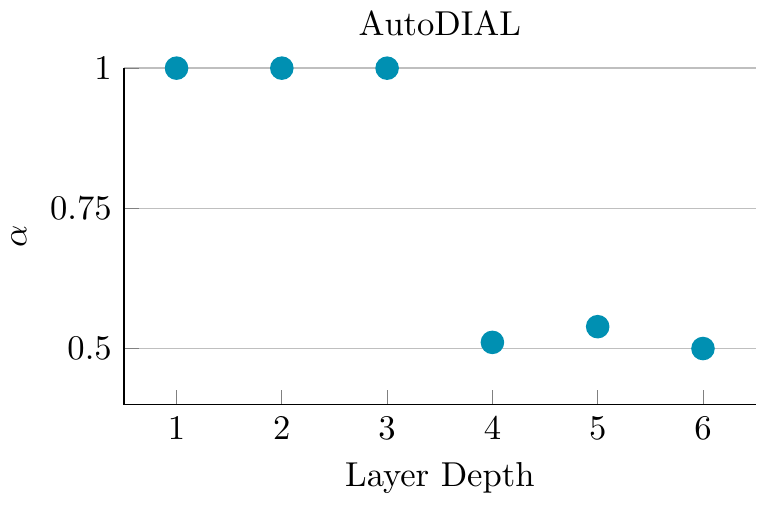}
  \caption{$\alpha$ parameters learned on the SVHN -- MNIST dataset, plotted as a function of layer depth.}
  \label{fig:alpha-svhn}
\end{figure}

As shown in Table~\ref{tab:svhn}, we set the new state of the art on this benchmark.
It is worth of note that \DIAL also outperforms the methods, such as ATN and DSN, which expand the capacity of the original network by adding numerous learnable parameters, while only employing a single extra learnable parameter in each \DAL.
The $\alpha$ parameters learned by \DIAL on this dataset are plotted in Fig.~\ref{fig:alpha-svhn}.
Similarly to the case of \Alex and \Inception on the Office-31 dataset, the network learns higher values of $\alpha$ in the bottom of the network and lower values of $\alpha$ in the top.
In this case, however, we observe a steeper transition from $1$ to $0.5$, which interestingly corresponds with the transition from convolutional to fully-connected layers in the network.

\begin{table}
  \centering
  \begin{tabular}{lc}
    \toprule
    Method & Accuracy \\
    \midrule
        CORAL~\cite{li2016revisiting} & $63.1$ \\
        DAN~\cite{long2015learning} & $71.1$ \\
    DANN~\cite{ganin2016domain} & $73.9$ \\
    DRCN~\cite{ghifary2016deep} & $82.0$ \\
    DSN~\cite{bousmalis2016domain} & $82.7$ \\
    ATN~\cite{saito2017asymmetric} & $86.2$ \\
    \midrule
    \DIAL & $\mathbf{90.3}$ \\
    \bottomrule
  \end{tabular}
  \vspace{0.5em}
  \caption{Results on the SVHN to MNIST benchmark.}
  \label{tab:svhn}
\end{table}

\subsection{Feature distributions}
\label{sec:distr}

\begin{figure}[ht!]
  \includegraphics[trim={1.7cm 11cm 1.7cm 11cm},clip,width=0.9\textwidth] {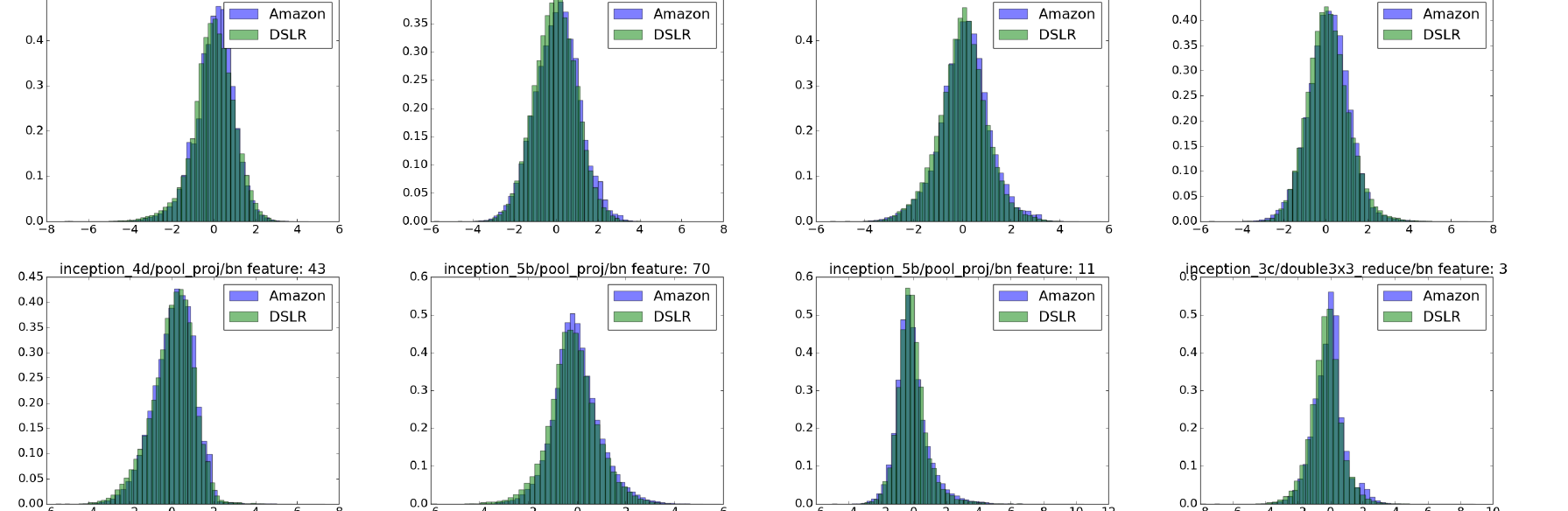}
  \caption{Distributions of randomly sampled source/target features from different layers of \DIALInception{} learned on the Amazon--DSLR task of the Office 31 dataset (best viewed on screen).}
  \label{fig:dial_supp_distr}
\end{figure}

In this section we study the distributions of a set of randomly sampled features from different layers of \DIALInception{}, learned on the Amazon--DSLR task of the Office 31 dataset.
In Fig.~\ref{fig:dial_supp_distr} we compare the histograms of these features, computed on the whole source and target sets and taken \emph{after} the \DALs.
The plots clearly show the aligning effect of our \DALs, as most histograms are very closely matching.
It is also interesting to note how the alignment effect seems to be mostly independent of the particular shape the distributions might take.

%%%%%%%%%%%%%%%%%% ADAGE
\section{ADAGE}

\subsection{Transformers architectures}  

\begin{figure}[tb]
\centering
\includegraphics[width=0.95\textwidth]{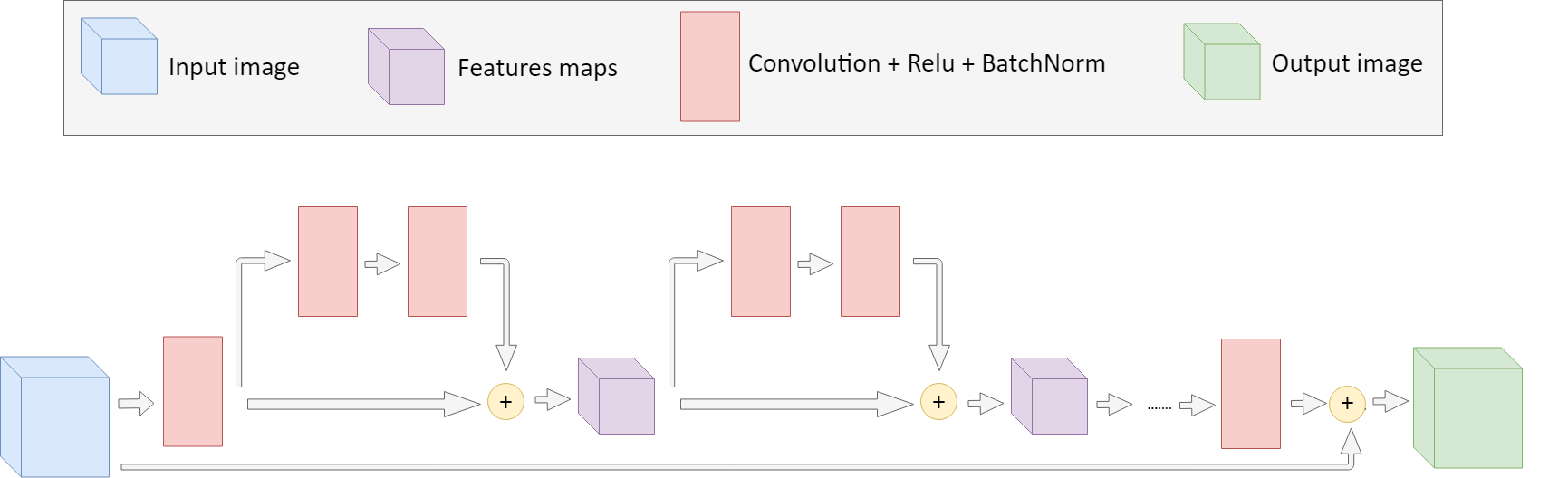}
\caption{Main blocks of our residual Transformer network. The blue block represent the input image data, while the red blocks are a sequence of Convolutional + Relu + Batch Normalization layers. The output of the two convolutional blocks are summed with the previous input. The number of kernels is fixed at 64 for the whole architecture until the last convolutional layer that brings the features back into 3 RGB channels (green block). }
\label{fig:adage_supp_transformer_residual}
\centering
\end{figure}

We have proposed two different types of architecture for the Transformer network: a residual version that follows the trend of
\cite{Bousmalis:Google:CVPR17} \cite{russo17sbadagan} \cite{CycleGAN2017}
 and an incremental version loosely inspired by \cite{shelhamerdeep}.
The residual Transformer \ref{fig:adage_supp_transformer_residual} uses a sequence of residual blocks made of two $3x3$ convolutional layers with 64 filters each to calculate the residual of the input ( an initial convolution brings the input data to 64 channels); this residual is then summed to the residual block input and fed to the next block in a full residual fashion. A total of 4 blocks have been used, after which a final convolutional layer brings the data back to three RGB channels. We finally use a last residual operation by summing up this output to the original input image: we found this operation beneficial to the stability of the algorithm.
The incremental Transformer 2 instead slowly build up the feature maps size by concatenating the output of two 3x3 convolutional layers to its input. The feature maps size grows following this sequence: $3-8-16-32-4-64-128$ after which they are bring down to 3 RGB channels. No residual w.r.t the input image is used for the incremental version. In both residual and incremental architectures, each convolutional layer is always followed up by a Relu non linearity and by a Batch Normalization layer but in the last convolutional layer, where it is beneficial to skip the Batch Normalization layer in order to the output image not being limited by the normalization procedure.

\subsection{Training details}  

During the experiments we found in some cases that the Transformer have been so successful in confusing the domains that the features domain discriminators cannot properly distinguish them, failing into provide a meaningful loss to the architecture. In those cases, the features domain discriminator produce an abnormal high loss and let the full architecture diverge.
We found a simple trick that is able to remove the most cases of instability: whenever the features domain discriminator loss exceed a threshold value, stopping the backpropagation of this loss in this iteration will avoid the training collapse. We empirically found that a threshold value of 3.0 works fine for every case, and we didn't found any performance difference, both accuracy-wise and image quality-wise, between training with and without this trick.

 \begin{figure}[tb]
     \centering

 \includegraphics[trim={4.5cm 20cm 4.5cm 4cm},clip,width=0.9\textwidth]
 {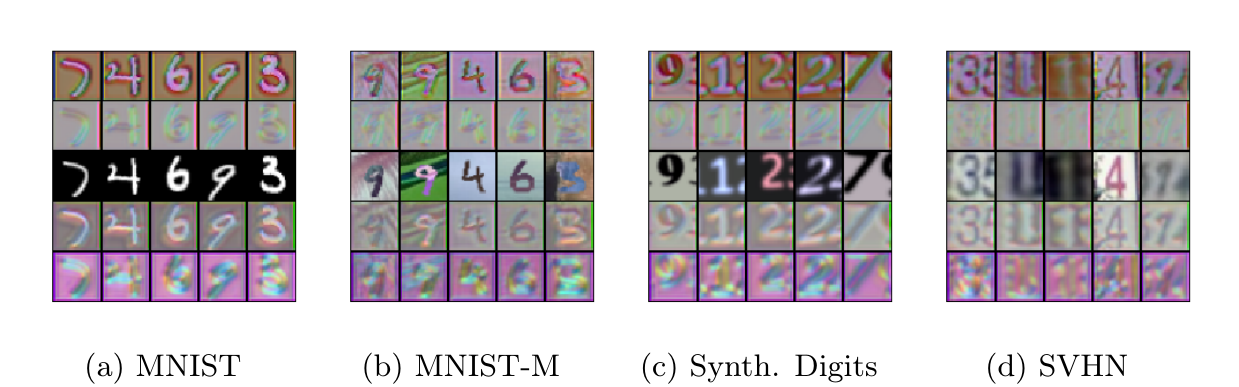}
    
 \caption{Examples of domain-agnostic digits generated by transformer block in the three source experiments with SVHN as target.
Top two rows show images produced in the DG setting by residual based (line $1$) and incremental based (line $2$) transformers. Line $3$ shows the original images and in the last two rows we display images produced by the residual (line $4$) and incremental transformers in the DA setting. As in the case with MNIST-M as target, images transformed with the residual architecture tend to preserve more of the original input. }
     \label{fig:adage_supp_examples_best}
 \end{figure}
 
 \begin{figure}[ht] 
 \includegraphics[trim={4.5cm 20.5cm 5.7cm 4cm},clip,width=0.9\textwidth]
 {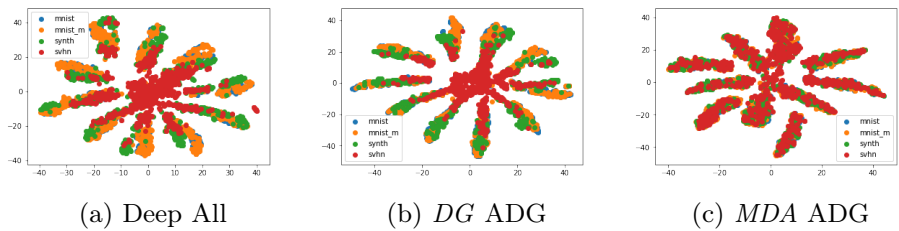} 
 \caption{TSNE visualization of the classification features. Here \textit{SVHN} is the target}
 \label{fig:adage_supp_tsne_svhn}
 \end{figure}

\begin{figure}[tb]
\includegraphics[width=0.9\linewidth]{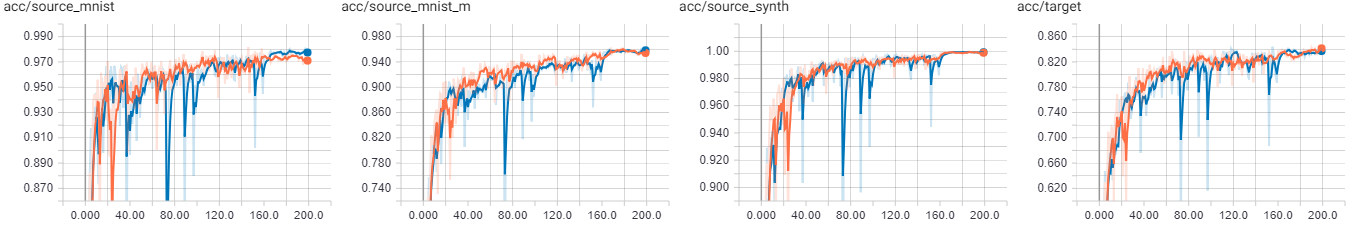}
\caption{Accuracy plots of our \adage residual transformer training on SVHN as target, red and blue are two separate runs. The three plots on the left show the accuracy on the \textbf{sources}, the one on right accuracy on the \textbf{target}. Note how there is a strong correlation between the performance on the source and on the target
%how the domain confusion losses help regularize and avoid the risk of overfitting on the sources
%quello che voglio dire qui è che non stiamo a pescare picchi, il miglior risultato è quello alla fine
\label{fig:adage_supp_accuracies}
%\times
}
\end{figure}

\begin{table}[t]
\centering
 \includegraphics[trim={2cm 10.5cm 3cm 11cm},clip,width=0.9\textwidth] {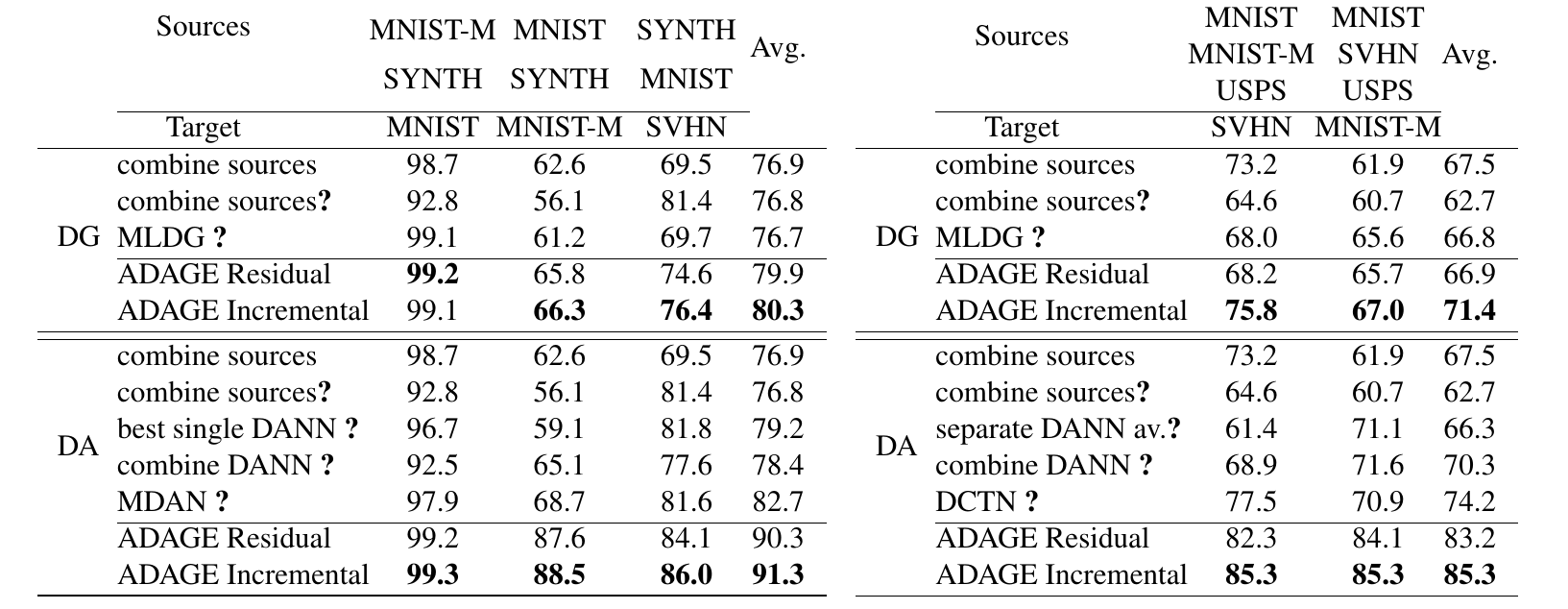}
 \caption{Extended edition of the classification accuracy results. \emph{On the left}: experiments with 3 sources. \emph{On the right}: experiments with 4 sources. Note that we report multiple versions of the \textit{combine source} baseline: we tried our best to replicate the base training protocol and network, but we could not fully replicate the published results with our own implementation. Still, the average results of the two versions of combine source do not differ excessively, especially in the three sources scenario (left).
}
\label{tab:adage_supp_results}
\end{table}

\backmatter
\phantomsection

\bibliographystyle{abbrv}
\bibliography{egbib}

\end{document}